%% file: main.tex
\documentclass{article}
\usepackage[utf8]{inputenc}
\usepackage[letterpaper,vmargin=1in,hmargin=1.25in]{geometry}
\usepackage{amsmath,amssymb,amsthm,mathtools}
\usepackage{graphicx,booktabs,tabularx,array,enumitem,microtype}
\usepackage{float,xcolor}
\floatstyle{ruled}
\newfloat{algorithm}{tbp}{loa}
\floatname{algorithm}{Algorithm}
\usepackage[authoryear,round]{natbib}
\setcitestyle{authoryear,round,citesep={;},aysep={,},yysep={;}}
\usepackage[hidelinks]{hyperref}
\usepackage{url}
\usepackage[nameinlink,capitalise]{cleveref}
\input{macros}

\input{revision-markup}
\input{reading-marks}
\hypersetup{pdfauthor={Andrew Lowy}}
\title{Optimal User-Level Private Stochastic Convex Optimization and Mean Estimation with Heavy Tails}
\author{Andrew Lowy\\
CISPA Helmholtz Center for Information Security\\
\texttt{lowy@cispa.de}}
\date{}
\begin{document}
\maketitle
\input{sections/00_abstract}
\input{sections/01_introduction}
\input{sections/02_preliminaries}

\input{sections/03_mean_estimation}

\input{sections/04_sco}

\input{sections/05_directional}
\input{sections/06_conclusion}
\label{main-text-end}
\input{sections/07_statements}
\bibliographystyle{plainnat}
\bibliography{references}
\clearpage
\appendix
\section*{Appendix}
\input{appendix/00_notation}
\input{sections/01b_related_work}
\input{appendix/03_mean_estimation}

\input{appendix/04_sco}

\input{appendix/05_directional}

\input{appendix/06_runtime}
\end{document}

%% file: macros.tex
\newtheorem{theorem}{Theorem}[section]
\newtheorem{lemma}[theorem]{Lemma}
\newtheorem{proposition}[theorem]{Proposition}
\newtheorem{remark}[theorem]{Remark}
\theoremstyle{definition}

\newtheorem{definition}{Definition}[section]

\newcommand{\E}{\mathbb E}
\newcommand{\Pp}{\mathbb P}
\newcommand{\R}{\mathbb R}
\newcommand{\eps}{\varepsilon}
\newcommand{\1}{\mathbf 1}
\newcommand{\wtG}{\widetilde G}
\newcommand{\cS}{\mathcal S}

\newcommand{\err}{\mathcal E_{n,m}}
\newcommand{\W}{\mathcal W}
\newcommand{\cP}{\mathcal P}

\newcommand{\wt}{\widetilde}

\newcommand{\andy}[1]{\textcolor{red}{[Andy: #1]}}

%% file: revision-markup.tex
\newif\ifshowrevisions
\showrevisionsfalse
\definecolor{revisionred}{RGB}{190,0,0}
\newcommand{\revisionstart}{\ifshowrevisions\color{revisionred}\fi}
\newcommand{\revisionend}{\color{black}}
\renewcommand{\andy}[1]{}
\makeatletter
\@namedef{revisionbib@kamath2022new}{}
\@namedef{revisionbib@kothari2022private}{}
\@namedef{revisionbib@narayanan2023better}{}
\AtBeginDocument{%
  \let\revision@lbibitem\@lbibitem
  \def\@lbibitem[#1]#2{%
    \@ifundefined{revisionbib@#2}{\revisionend}{\revisionstart}%
    \revision@lbibitem[#1]{#2}}%
}
\makeatother

%% file: reading-marks.tex
\renewcommand{\revisionstart}{}
\renewcommand{\revisionend}{}
\newcommand{\readingstart}[1]{\begingroup\color{black}}
\newcommand{\readingend}{\endgroup}

%% file: sections/00_abstract.tex
\begin{abstract}

We study stochastic convex optimization with heavy-tailed stochastic gradients under $(\eps, \delta)$-user-level
differential privacy, which protects each user's entire collection of samples. Instead of assuming a uniformly bounded Lipschitz constant, we only assume bounds on the second- and $k$-th moments of the random sample Lipschitz parameters. This allows for unbounded, heavy-tailed stochastic gradient distributions and can lead to sharper excess risk bounds. We characterize the minimax optimal excess population risk up to logarithmic factors.
Our algorithm achieves the optimal rate with only $\wt{O}(1/\eps)$ users. 
As a byproduct of our algorithmic developments, we obtain tight bounds for user-level private mean
estimation under norm-moment assumptions, improving over the guarantees
of \cite{narayanan2022few,zhao2024huber}.
Finally, for mean estimation under the directional-moment
assumption of \cite{agarwal2024person}, we close their remaining
approximate-DP gap by removing the additional $d/\varepsilon$
user requirement and attaining the optimal sample complexity
up to logarithmic factors.
\end{abstract}

%% file: sections/01_introduction.tex
\section{Introduction}
\label{sec:intro}

\textit{Stochastic convex optimization} (SCO) is a fundamental problem in machine learning. Given samples from an unknown distribution $P$, the goal is to minimize the population loss
\begin{equation}
 \min_{w\in\W} F(w),
 \qquad F(w):=\E_{z\sim P}[f(w,z)],
 \label{eq:SCO}
\end{equation}
where $\W\subset\R^d$ is a compact convex domain of $\ell_2$-diameter $D$ and $f(\cdot,z)$ is convex. We measure the quality of a solution $w$ by its \textit{excess population risk}, $F(w)-F^*$, where $F^*:=\min_{w'\in\W}F(w')$. 

In many applications, the data used to train models contains sensitive information, and each person contributes multiple training examples: messages typed on a phone, repeated measurements from a patient, or interactions with an online service. \textit{Differential privacy} (DP) provides a rigorous framework for protecting sensitive data~\citep{dwork2006calibrating}. In the \textit{user-level} model, this protection extends to each user's entire collection of samples: the algorithm's output distribution must remain stable when all observations of one user are replaced. Motivated by these considerations, user-level private estimation and SCO have received substantial attention~\citep{levy2021user,narayanan2022few,bassily2023user,asi2024user,lowy2024faster,agarwal2024person}.

Much of the user-level SCO literature assumes that the stochastic gradients are \textit{uniformly bounded}~\citep{levy2021user,asi2024user,lowy2024faster}.
The sample Lipschitz constant is
$\sup_{w\in\W}\|\nabla f(w,z)\|$,
and uniform boundedness requires $\sup_z\sup_{w\in\W}\|\nabla f(w,z)\|\le L$. The resulting excess risk guarantees scale with the worst-case Lipschitz parameter $L$. In many problems, however, $L$ can be extremely large or infinite, making these guarantees overly pessimistic or vacuous. 

Consequently, a recent line of work studies \textit{heavy-tailed} DP SCO under a weaker assumption: bounded  \textit{moments} of {the random sample Lipschitz parameter}. Instead of assuming a bounded worst-case Lipschitz parameter, one
assumes that, for public $0<G_2\le G_k$ and some $k\ge2$:
\begin{equation}
 \E_{z\sim P}\left[\sup_{w\in\W}\|\nabla f(w,z)\|^2\right]\le G_2^2,
 \qquad
 \E_{z\sim P}\left[\sup_{w\in\W}\|\nabla f(w,z)\|^k\right]\le G_k^k.
 \label{eq:momentbound}
\end{equation}
\revisionstart Note that $G_2$ and $G_k$ are public upper bounds, not exact moment values; Remark~\ref{rem:moment-inputs} in the Appendix discusses misspecification.\revisionend{}

This assumption allows for unbounded, heavy-tailed stochastic gradients and can yield sharper excess risk bounds when $G_k$ is much smaller than $L$. For instance, for squared-loss regression with $\mathcal{W}$ contained in a ball of radius $R$, we have $\sup_{w\in\W}\|\nabla f(w,(x,y))\|\le R\|x\|^2+|y|\|x\|$. Thus, if moments of the features and responses are bounded, then~\eqref{eq:momentbound} holds even when data is unbounded and $L = \infty$. Heavy-tailed private estimation and SCO have been studied extensively under \textit{item-level privacy}, which protects one data point at a time~\citep{kamath2020heavy,wang2020heavy,kamath2022improved,lowy2023large,asi2024heavy,lowy2026pure}, and {sharp rates are known in several of these settings}. Much less is understood about the user-level setting, however.

There are $n$ users, each holding $m$ observations, and all $nm$ observations are drawn i.i.d. from $P$. Our main contribution is to characterize the minimax optimal excess population risk under second- and $k$th-moment bounds ~\eqref{eq:momentbound} up to logarithmic factors.

\subsection{Our contributions}

Before addressing SCO, we study the simpler problem of mean estimation with \textit{bounded norm-moments}: for a sample $z\sim P$ with unknown mean $\mu$, assume $\E_{z\sim P}\|z-\mu\|^j\le \wtG_j^j$ for $j\in\{2,k\}$. Let $\cP_k(\wtG_2,\wtG_k)$ denote this distribution class. We use $\wtG_j$ for these centered moments and reserve $G_j$ for the raw SCO moments in~\eqref{eq:momentbound}. 
Despite recent progress~\citep{narayanan2022few,zhao2024huber}, existing error bounds for user-level private mean estimation with norm-moment bounds are not tight in every parameter regime. \revisionstart Similarly, naively applying an optimal item-level DP estimator to user averages is suboptimal. \revisionend See Table~\ref{tab:mean} for a summary. Furthermore, mean estimation provides a key building block for our SCO algorithm. This leads to:

\begin{center}
\fbox{
\parbox{0.65\linewidth}{
\textbf{Question 1.}
What is the minimax optimal $\ell_2$-error for user-level private mean estimation under norm-moment bounds?
}}
\end{center}

\paragraph{Contribution 1: Optimal rates for user-level DP mean estimation with norm-moment bounds.} We answer Question 1 by providing novel tight upper and lower bounds under user-level $(\varepsilon,\delta)$-DP (up to logarithms). The minimax optimal $\ell_2$-error rate scales as
\begin{align}
 \err(\wtG_2,\wtG_k)
 &:=\frac{\wtG_2}{\sqrt{nm}}+\frac{\wtG_2\sqrt d}{n\varepsilon\sqrt m}
 \notag\\
 &\quad+\min\Biggl\{
 \wtG_2\left(\frac{\sqrt d}{nm\varepsilon}\right)^{1/2},\;
 \wtG_k\left(\frac{\sqrt d}{nm\varepsilon}\right)^{1-1/k}
 \Biggr\}.
 \label{eq:main-rate}
\end{align}
Our estimator attains the optimal error rate up to logarithmic factors whenever the number of users $n\ge\widetilde\Omega_k(1/\varepsilon)$ --- a user threshold that is information-theoretically necessary (for finite uniform error when $\delta\le\varepsilon^2$) up to logarithmic factors~\citep[Lemma~5.1]{narayanan2022few}.

The three terms in $\err(\wtG_2,\wtG_k)$ reflect non-private sampling error, the user-level privacy cost of typical (subgaussian) fluctuations of user averages, and the privacy cost of rare large observations. The second term is the {optimal privacy term} for bounded-data user-level mean estimation \citep[Corollary 4.2 and Theorem 5.2]{narayanan2022few}, whereas the third term has the form of the optimal item-level DP heavy-tailed error with \(nm\) observations~\citep{barber2014privacy,lowy2023large}, while accounting for both second- and \(k\)th-moment constraints.

These results improve the guarantees of prior user-level mean estimators: see Table~\ref{tab:mean} and Appendix~\ref{app:related}. 
\begin{table}[ht]
\centering
\caption{%
Norm-moment mean estimation with user-level
$(\varepsilon,\delta)$-DP: constant-probability Euclidean errors.
Logarithmic factors are suppressed. Our lower bound assumes~\eqref{eq:lower-delta}.
}
\label{tab:mean}
\small
\begin{tabularx}{\linewidth}{@{}
>{\raggedright\arraybackslash}p{.28\linewidth}
>{\raggedright\arraybackslash}p{.22\linewidth}
>{\raggedright\arraybackslash}X@{}}
\toprule
Result & Error bound & Restrictions\\
\midrule
\citet{narayanan2022few} upper bound
& $\displaystyle\frac{\wtG_2\sqrt d}{n\varepsilon\sqrt m}$
& $1/\varepsilon\lesssim n\lesssim\sqrt d/\varepsilon$\\
\midrule

\citet{zhao2024huber} upper bound
& $\err(\wtG_k,\wtG_k)$
& $n\gtrsim d/\varepsilon$;
known $R_{\mathrm{public}}\ge\|\mu\|$\\
\midrule

\revisionstart
Naive item-level reduction
(Appendix~\ref{app:item-reduction})
\revisionend
& \revisionstart
$\begin{aligned}
&\frac{G}{\sqrt{nm}}\\
&{}+\frac{G}{\sqrt m}
\left(\frac{\sqrt d}{n\varepsilon}\right)^{1-1/k}
\end{aligned}$
\revisionend
& \revisionstart
$\wtG_2=\wtG_k=G$;
$n\gtrsim\sqrt d/\varepsilon$
\revisionend\\
\midrule

\textbf{Our matching bounds}
& $\err(\wtG_2,\wtG_k)$
& $n\gtrsim1/\varepsilon$\\
\midrule

\textbf{Too few users}~\cite[Lemma~5.1]{narayanan2022few}
& No finite uniform error
& $n\le\log(1/\delta)/(3\varepsilon)$; $\delta\le\varepsilon^2$\\
\bottomrule
\end{tabularx}
\end{table}
Using our mean estimation results and techniques, we next tackle stochastic convex optimization. 

\begin{center}
\fbox{
\parbox{0.67\linewidth}{
\textbf{Question 2.}
What is the minimax optimal excess population risk for user-level DP SCO under second- and $k$th-moment bounds ~\eqref{eq:momentbound}?
}}
\end{center}

\paragraph{Contribution 2: Optimal rates for user-level DP SCO with bounded moments.}
We answer Question 2 by providing tight upper and lower bounds (up to logarithms). With $n\ge\widetilde\Omega_k(1/\varepsilon)$ users, the optimal rate under~\eqref{eq:momentbound} is
\begin{equation}
 \widetilde\Theta_k\!\left(
 D\min\{G_2,\err(G_2,G_k)\}
 \right),
 \label{eq:intro-sco}
\end{equation}
where $\err(G_2,G_k)$ is the expression in~\eqref{eq:main-rate} evaluated at the raw SCO moment bounds. 
Table~\ref{tab:main-sco} compares our result with prior SCO guarantees. %

\begin{table}[t]
\centering
\caption{%
Summary of SCO rates: excess population risk. Logarithmic factors are suppressed.
The prior rows assume a bound $\sup_{w,z}\|\nabla f(w,z)\| \le L$; our row
assumes~\eqref{eq:momentbound}, with~\eqref{eq:lower-delta} for the lower bound.
}
\label{tab:main-sco}
\small
\begin{tabularx}{\linewidth}{@{}
>{\raggedright\arraybackslash}p{.29\linewidth}
>{\raggedright\arraybackslash}X
>{\raggedright\arraybackslash}p{.28\linewidth}@{}}
\toprule
Result & Excess population risk & Restrictions\\
\midrule

\citet{levy2021user}
& $LD\min\!\left\{1,\,
\frac{1}{\sqrt{nm}}+\frac{d}{n\varepsilon\sqrt m}\right\}$
& $n\gtrsim\sqrt d/\varepsilon$;
additional smoothness restriction\\
\midrule

\citet{narayanan2022few}
& $LD\min\!\left\{1,\,
\frac{1}{\sqrt{nm}}+\frac{d}{n\varepsilon\sqrt m}\right\}$
& $n\gtrsim1/\varepsilon$;
additional smoothness restriction\\
\midrule

\citet{bassily2023user}
& $LD\min\!\left\{1,\,
\frac{1}{\sqrt{nm}}+\frac{\sqrt d}{n\varepsilon\sqrt m}\right\}$
& $n\gtrsim\sqrt d/\varepsilon$;
restricted $m$, $\delta$, and smoothness parameter\\
\midrule

\citet{asi2024user,lowy2024faster}
& $LD\min\!\left\{1,\,
\frac{1}{\sqrt{nm}}+\frac{\sqrt d}{n\varepsilon\sqrt m}\right\}$
& $n\gtrsim1/\varepsilon$\\
\midrule

\textbf{Our matching bounds}
& $D\min\{G_2,\err(G_2,G_k)\}$
& $n\gtrsim1/\varepsilon$\\

\bottomrule
\end{tabularx}
\end{table}
Finally, we revisit mean estimation under an alternative bounded \textit{directional moment} condition that was studied by~\cite{agarwal2024person}: \begin{equation}
 \sup_{\|v\|=1}\E|\langle z-\mu,v\rangle|^k\le1.
 \label{eq:intro-directional-model}
\end{equation}

Under the small-$\delta$ condition in Theorem~\ref{thm:main-directional}, achieving $\ell_2$-error $\alpha>0$ requires the following number of users (Appendix~\ref{app:directional}, using lower bounds of \citealp{agarwal2024person,narayanan2022few}):
\begin{equation}
\widetilde \Omega_k\!\left(
 \frac d{m\alpha^2}
 +\frac d{\alpha\varepsilon\sqrt m}
 +\frac d{m\varepsilon\alpha^{k/(k-1)}}
 +\frac 1\varepsilon
 \right)
 \label{eq:intro-directional1}
\end{equation}
By contrast, the upper bound of \citet{agarwal2024person} has
$d/\varepsilon$ in place of the $1/\varepsilon$ term
in~\eqref{eq:intro-directional1}. The following question remained open:

\begin{center}
\fbox{
\parbox{0.65\linewidth}{
\textbf{Question 3.}
Does there exist a user-level DP algorithm for mean estimation under~\eqref{eq:intro-directional-model} matching the lower bound~\eqref{eq:intro-directional1}? 
}}
\end{center}

\paragraph{Contribution 3: Optimal directional-moment mean estimation.} We give a positive answer to Question 3, providing an optimal estimator whose user complexity matches~\eqref{eq:intro-directional1} up to logarithms. 
 
Complete proofs are deferred to the appendices.

\subsection{Challenges and techniques}
\label{sec:techniques}
We provide an overview of the key technical challenges and tools that we develop.

\subsubsection{Mean estimation with bounded norm-moments.}
It is unclear how to achieve the optimal rate throughout the $n \ge \wt{\Omega}(1/\eps)$ regime by directly modifying the Huber-loss algorithm of \cite{zhao2024huber} or the private ball-finding method of \cite{narayanan2022few}. Therefore, we depart from these prior approaches for heavy-tailed estimation and instead build on FriendlyCore, which is optimal for \textit{bounded-data} user-level DP estimation with $n \ge \wt{\Omega}(1/\eps)$ users~\citep{tsfadia2022friendlycore,asi2024user}. \revisionstart However, it is suboptimal in the heavy-tailed setting because FriendlyCore can discard rare user averages whose contribution to the mean is significant. To overcome this obstacle,\revisionend{} we develop a novel \textbf{two-stage algorithm}: first,
we use FriendlyCore to provide a coarse estimate of the mean
using half the users; we then use the coarse estimate as a
clipping center and do a clip-average-and-noise step on the
rest of the users\revisionstart, with the clipping radius chosen to balance bias and privacy noise\revisionend.

The clipping center is only coarsely accurate and it is challenging to bound the clipping bias. \revisionstart To overcome this, we argue that an accurate coarse center ensures that clipping leaves user averages near the population mean unchanged, so the distribution's tails control the resulting bias. Our analysis shows that even a coarse, noisy clipping center permits sharp bias bounds by exploiting the concentration of user averages via the Fuk--Nagaev inequality~\citep{mollenhauer2025fuk}. This separates typical fluctuations from rare large observations, retaining the sharp dependence on $m$ that a generic item-level moment bound loses (Appendix~\ref{app:item-reduction}).\revisionend

\paragraph{Lower bound.}
For the heavy-tailed lower bound term, we build on the rare-outlier approach
of~\cite{barber2014privacy,lowy2023large}, which gave a loose
approximate DP lower bound and a tight zCDP lower bound,
respectively.
The challenge is obtaining the sharp dependence on
dimension under approximate DP. \revisionstart These\revisionend{} prior constructions were impeded by using a fixed outlier vector. To overcome this, we draw
outliers randomly\revisionstart, as in~\cite{kamath2022new,narayanan2023better}\revisionend; observing an outlier no longer reveals the mean exactly. We can then apply the fingerprinting lemma
of~\cite{klsu19} to get our tight bound.

An important technical contribution is the sharp joint dependence on both moment scales in order to get tight lower bounds: we scale rare-outlier instances to satisfy both constraints and combine them with a Gaussian lower bound for whole user blocks. Appendix~\ref{app:related} discusses these connections.

\subsubsection{SCO.}
Our SCO algorithm builds on the population-level localization framework of \cite{asi2024heavy}, reducing heavy-tailed SCO to a sequence of regularized ERM problems. We privately solve each regularized ERM problem by running a user-level DP variant of (stochastic) gradient descent. The key question is \textit{how to estimate the gradient of the empirical loss function in each iteration when user-level DP is required for $T$ adaptive iterates}.

\revisionstart A natural approach is to iterate our two-stage mean estimator within this ERM solver.\revisionend{} The central challenge is that increasing the privacy noise in the FriendlyCore stage of our algorithm to account for $T$ releases would destroy the accuracy of clipping centers, leading to suboptimal excess risk. \revisionstart Center noise can increase clipping bias; enlarging the clipping radius instead increases gradient noise. To avoid separately noising the clipping centers,\revisionend{} we develop a novel \textbf{fused adaptive variant} of \cref{alg:main-me}: each time we run the FriendlyCore step, we keep the coarse center that it outputs internal and don't add any noise to it; we only release the noisy clipped gradient. We combine this technique with \textit{AboveThreshold}. \revisionstart ``Fused'' means that coarse-center construction and clipping form one private gradient release. Proving privacy requires controlling how changing one user affects both the internal center and every clipped gradient; our sensitivity analysis controls both effects (Proposition~\ref{prop:private-erm-solver}).\revisionend

Another challenge is that \textit{reusing users makes the gradient queries adaptive}.
To overcome this, we leverage the fact that privacy limits how strongly the iterates depend on any individual user. Rather than standard worst-case bias analysis, we control the \textbf{average squared empirical clipping bias} via DP stability, avoiding a spurious $\text{poly}(T)$ factor.

\revisionstart\paragraph{Lower bound.}
Two obstacles prevent directly transferring our mean-estimation lower bound to SCO: the gradients must satisfy raw moment bounds, and, for linear losses, optimization only requires learning the mean's direction. We adapt the Gaussian fingerprinting argument of~\cite{klsu19} to bound how well a private optimizer can align with that direction. We then use a rare-outlier reduction~\citep{barber2014privacy,lowy2023large,narayanan2023better}. The main technical work is controlling this alignment under user-level privacy and scaling the Gaussian instances and their rare-outlier mixtures to satisfy both raw-moment constraints while retaining the sharp dependence on $G_2$ and $G_k$ (Appendix~\ref{app:sco-lower}).\revisionend

\subsubsection{Mean estimation with bounded directional moments.}
The key challenge is \textit{removing the $d/\eps$ user
requirement} of~\cite{agarwal2024person} in the intermediate-accuracy regime.
We \textbf{modify the refinement step} of their algorithm by relaxing their per-round accuracy requirement and proving that the error still decreases toward
the requested accuracy. \revisionstart We choose the clipping radius using a public bound on the current center error and prove that these bounds contract toward the target accuracy (Appendix~\ref{app:directional}).\revisionend

%% file: sections/02_preliminaries.tex
\section{Preliminaries}
\label{sec:prelim}

Write $B(u,r)=\{x:\|x-u\|\le r\}$ and $\Pi_K$ for Euclidean projection onto a closed convex set. User $i$ holds $Z_i=(z_{i,1},\ldots,z_{i,m})$, and $\mathsf Z=(Z_1,\ldots,Z_n)\sim P^{nm}$. All norms are Euclidean; $I_d$ is the identity matrix. Assume $0<\varepsilon\le1$, $0<\delta<1/2$, fixed \revisionstart $k\ge2$\revisionend, and failure probability $\beta\in(0,1/4)$. $\widetilde O_k$ hides logarithms and constants depending on $k$, and $C_{(\cdot)}$ and $c_{(\cdot)}$ denote constants (possibly depending on $k$). Throughout, $\Lambda:=\log(e nmd/(\varepsilon\delta\beta))$.

\begin{definition}[User-level differential privacy]
An algorithm $A$ is $(\varepsilon,\delta)$-user-level DP if, for all datasets $\mathsf Z,\mathsf Z'$ differing in one entire user block and all measurable output events $S$,
\begin{equation}
 \Pr(A(\mathsf Z)\in S)\le e^\varepsilon\Pr(A(\mathsf Z')\in S)+\delta.
 \label{eq:main-dp}
\end{equation}
\end{definition}
\revisionstart
User-level privacy holds for arbitrary datasets, including correlated or heterogeneous user blocks. The i.i.d. and moment assumptions are used only for utility.
\revisionend

\paragraph{SCO assumptions.}
In addition to the SCO assumptions stated in the introduction, sample losses are differentiable on a neighborhood of $\W$. For differentiable $h$, \textit{convexity} means $h(v)\ge h(w)+\langle\nabla h(w),v-w\rangle$; $\lambda$-\textit{strong convexity} adds $\lambda\|v-w\|^2/2$ to the right-hand side. A function is $L$-\textit{Lipschitz} if $|h(v)-h(w)|\le L\|v-w\|$. By~\eqref{eq:momentbound}, $F$ is $G_2$-Lipschitz, so $F(w) - F^* \le D G_2$~$\forall w \in \W$. 

Assume without loss of generality that $\wtG_2\le\wtG_k$; the SCO parameters already satisfy $G_2\le G_k$.

%% file: sections/03_mean_estimation.tex
\section{Mean estimation under norm-moment bounds}
\label{sec:main-me}
In this section, we provide tight upper and lower bounds for mean estimation over $P\in\cP_k(\wtG_2,\wtG_k)$. %
\subsection{Upper bound: FriendlyCore Coarse Estimation, Clip-Average-Noise}

For user averages $Y_i=m^{-1}\sum_a z_{i,a}$,
our Algorithm~\ref{alg:main-me} uses half the users to obtain
a coarse estimate $c$ of $\mu$ via FriendlyCore (Algorithm~\ref{alg:friendlycore}) \citep{asi2024user,tsfadia2022friendlycore}.
It then clips the remaining users' averages around $c$,
averages them, and adds Gaussian noise.
The clipping radius $R$ balances clipping bias and privacy noise.

FriendlyCore first tests whether the data set is {sufficiently} concentrated --- i.e. whether most user averages have many nearby neighbors. If not, the algorithm aborts; if yes, it then filters out outlier users (i.e. those with few neighbors). Finally, it adds Gaussian noise $u$ to the average $Q$ of the surviving inlier user averages $\mathcal{S}$ and outputs this vector $\wt{Q} = Q + u$ as the coarse estimate of $\mu$.

\begin{algorithm}[H]
\caption{Norm-moment mean estimation}
\label{alg:main-me}
\textbf{Input:} User averages $Y_1,\ldots,Y_n$ and public
$\wtG_2,\wtG_k,m,d,\varepsilon,\delta,\beta$.
\begin{enumerate}[leftmargin=*,itemsep=2pt,topsep=3pt]
\item Split users into two disjoint groups $S_1,S_2$ of sizes $n_1=\lfloor n/2\rfloor$ and $n_2=n-n_1$.
\item Compute the coarse center using Algorithm~\ref{alg:friendlycore}:
\[
 c\leftarrow\mathsf{FriendlyCore}\!\left(
 (Y_i)_{i\in S_1};\,
 2\wtG_2\sqrt{40/m},\,\varepsilon,\,\delta
 \right),
 \qquad n_1=|S_1|.
\]
\item If $\sqrt d/(n\varepsilon)>1$, return $\hat\mu:=c$.
\item Set $a=\sqrt{d\log(e/\delta)}/(n\varepsilon)$ and choose
$j\in\arg\min_{\ell\in\{2,k\}}\wtG_\ell(a/m)^{1-1/\ell}$.
\item Set $R=\max\{C_k\Lambda^2\wtG_2/\sqrt m,\,
\wtG_jm^{-(1-1/j)}a^{-1/j}\}$.
\item Return
$\hat\mu:=n_2^{-1}\sum_{i\in S_2}\Pi_{B(c,R)}(Y_i)+\xi$,
where $\xi\sim\mathcal N(0,\sigma_2^2I_d)$ 
, $\sigma_2=8R\sqrt{\log(e/\delta)}/(n_2\varepsilon)$.
\end{enumerate}
\end{algorithm}

\begin{theorem}[Norm-moment estimation upper bound]
\label[theorem]{thm:main-me}
For every $P\in\cP_k(\wtG_2,\wtG_k)$ and $n\ge C_k\Lambda^4/\varepsilon$, there exist algorithmic parameters such that Algorithm~\ref{alg:main-me} is user-level $(\varepsilon,\delta)$-DP and satisfies
$\|\widehat\mu-\mu\|\le\widetilde O_k(\err(\wtG_2,\wtG_k))$
with probability at least $1-\beta$.
\end{theorem}

\paragraph{Privacy proof sketch.}
FriendlyCore is $(\eps,\delta)$-DP by
\citet[Lemma~3.6]{asi2024user}. \revisionstart Lemma~\ref{lem:fc-privacy} makes the calibration for our constants explicit.\revisionend{} For fixed $c$, the second-stage
average has sensitivity $2R/n_2$, so the Gaussian mechanism
is also $(\eps,\delta)$-DP. Since the stages use disjoint
users, parallel composition gives the claim.
\paragraph{Utility proof sketch.} At a high level, the argument goes as follows: We first show that the coarse center provided by FriendlyCore is accurate with high probability (Lemmas~\ref{lem:core} and~\ref{lem:coarse}). We then argue that conditional on this ``good center'' event, the clipping bias in the second stage of the algorithm is sufficiently small if the clip threshold $R$ is sufficiently large (~\cref{lem:main-bias}). 

\paragraph{Step 1: Accuracy of FriendlyCore.}
With bounded or subexponential data, no users get filtered out with high probability and accuracy follows. 
With heavy tails, some user averages can be far from $\mu$ and hence not every user survives the FriendlyCore filter.
However, a large majority of user averages are near $\mu$ and hence survive the filter: 

\begin{lemma}[Large majority core]
\label[lemma]{lem:core}
Define
$\mathrm{GOOD}:=\{i\in S_1:\|Y_i-\mu\|\le\tau/2\}$ and $\tau=2\wtG_2\sqrt{40/m}.$ With probability at least \(1-e^{-n_1/800}\), we have
$
  |\mathrm{GOOD}|\ge0.95n_1.
$

\end{lemma}
Chebyshev and Hoeffding prove Lemma~\ref{lem:core}.
The large majority core does not give optimal accuracy from FriendlyCore
alone, but it does provide a fairly accurate coarse estimate of $\mu$:

\begin{lemma}[Accuracy of the coarse center]
\label[lemma]{lem:coarse}
Suppose $n_1\ge \frac{1000}{\eps}\log\frac{16}{\delta\beta}.$
Conditional on $|\mathrm{GOOD}|\ge0.95n_1$, %
with probability at least $1-2\beta$, the output $\wt{Q}$ of \cref{alg:friendlycore} satisfies
\begin{equation}
  \|\widetilde Q-\mu\|
  \le \widetilde{O}\!\left(
  \frac{\wtG_2}{\sqrt m}
  +\frac{\wtG_2\sqrt{d+\log(1/\beta)}}{n_1\eps\sqrt m}
  \right).
  \label{eq:coarse-bound}
\end{equation}
\end{lemma}
\begin{proof}[Proof idea.]
With high probability, every GOOD user survives (i.e. $\text{GOOD} \subseteq \cS$),
and every surviving user has a GOOD neighbor within distance $2\tau$. Thus the average $Q$ of the surviving user-averages is $O(\tau)$-close to $\mu$; $\tau = 2\wtG_2 \sqrt{40/m}$. Gaussian noise contributes the second term
in~\eqref{eq:coarse-bound}.
\end{proof}

If $\sqrt d/(n\varepsilon)>1$, returning
$c=\widetilde Q$ already attains~\eqref{eq:main-rate},
up to logarithms, by Lemma~\ref{lem:coarse}.
Otherwise, stage two refines the estimate.

\paragraph{Step 2: Bounding clipping bias on ``good center'' event.}
On the ``good clipping center'' event $\|c - \mu\| \le R/2$, clipping around $c = \widetilde{Q}$ is almost as good as clipping around $\mu$ because clipping only affects user averages who are at least $R/2$-far from $\mu$.
Only these outlier users contribute to the bias. Thus the bias can be controlled by the tail of the random variable $\frac{1}{m}\sum_{j=1}^m {z_j} - \mu$, which can be tightly bounded via Fuk-Nagaev inequality, yielding:

\begin{lemma}[Clipping bias around a good center]
\label[lemma]{lem:main-bias}
Let $Y$ average $m$ independent observations from $P\in\cP_k(\wtG_2,\wtG_k)$. For a fixed center $c$, if $\|c-\mu\|\le R/2$ and $R\gtrsim_k \wtG_2/\sqrt m$, then
\begin{align}
 \|\E\Pi_{B(c,R)}Y-\mu\|
 &\le C \wtG_2^2/(mR),\notag\\
 \|\E\Pi_{B(c,R)}Y-\mu\|
 &\le C_k\left[\frac{\wtG_2}{\sqrt m}e^{-c_kmR^2/\wtG_2^2}
       +\frac{\wtG_k^k}{m^{k-1}R^{k-1}}\right].
 \label{eq:main-bias}
\end{align}
\end{lemma}

Finally, to obtain the error bound in \cref{thm:main-me}, we choose the clipping threshold $R$ optimally to balance the bias term (\cref{lem:main-bias}) and the Gaussian noise term $\sigma_2 \sqrt{d}$, subject to the constraint that the good clipping center event occurs with high probability. By Lemma~\ref{lem:coarse}, this constraint can be satisfied by choosing $R$ at least twice the right-hand side of~\eqref{eq:coarse-bound}.

\revisionstart \paragraph{Runtime.}
Norm-moment mean estimation takes $O(nmd+n^2d)$ arithmetic operations, including FriendlyCore's pairwise distances; Appendix~\ref{app:runtime} gives the accounting.\revisionend{}

\subsection{Matching lower bound}
The following tight lower bound establishes the optimality of ~\cref{alg:main-me} up to logarithmic factors in the regime $n \ge \wt{\Omega}_k(1/\eps)$. 

\begin{theorem}[Norm-moment estimation lower bound]
\label[theorem]{thm:main-me-lower}
Let $\delta \le c\min\{\frac{1}{nm}, \frac{\eps^2}{d}\}$ for sufficiently small $c>0$. For every user-level $(\varepsilon,\delta)$-DP estimator, there exists a distribution $P\in\cP_k(\wtG_2,\wtG_k)$ such that
$\Pr(\|\widehat\mu-\mu\|\ge c_k\err(\wtG_2,\wtG_k))\ge1/3$.
\end{theorem}

Separately, \citet[Lemma~5.1]{narayanan2022few} show that if $\delta\le\eps^2$ and $n\le\log(1/\delta)/(3\eps)$ then no user-level DP algorithm can achieve a finite uniform error bound for unrestricted mean location. Thus, the user count  $n \ge \wt{\Omega}_k(1/\eps)$ of our algorithm is tight up to logarithms.

\paragraph{Proof overview.}
Recall the three terms in~\eqref{eq:main-rate}. The nonprivate term is classical~\cite{duchi2023notes}. For the Gaussian privacy term, we adapt the fingerprinting argument of \citet[Appendix~D]{agarwal2024person}, scaling the covariance to satisfy both norm-moment bounds; translation lets the mean range be larger without changing the centered moments. For the heavy-tailed privacy term, we modify the rare-outlier construction of~\citet{barber2014privacy,lowy2023large}\revisionstart , also drawing on~\citet{kamath2022new,narayanan2023better}\revisionend{}: we calibrate the magnitude of the outliers to satisfy both moment bounds, and randomly sample its direction from a product distribution on $\{-1,1\}^d$. This randomization permits the approximate-DP fingerprinting lemma of~\citet[Lemma~53]{klsu19}; grouping the $nm$ observations into $n$ users then gives the user-level bound via~\cite{levy2021user}.

%% file: sections/04_sco.tex
\section{Stochastic convex optimization under norm-moment bounds}
\label{sec:main-sco}

In this section, we address the SCO problem~\eqref{eq:SCO} with bounded moments~\eqref{eq:momentbound} under user-level DP.

\subsection{Upper bound: Localized SGD with Fused Adaptive Private Gradient Estimator}
Our SCO algorithm builds on the population-level localization framework of \cite{asi2024heavy}, reducing heavy-tailed SCO to a sequence of regularized ERM problems: see Algorithm~\ref{alg:poploc}. We privately solve each regularized ERM problem by running a user-level DP variant of gradient descent.

\subsubsection{Population localization reduces SCO to regularized ERM}
For a fresh dataset $Z$ of $s$ users with $m$ observations each, write $\widehat F_Z$ for the empirical loss. For a regularization center $w_0\in\W$ and $\lambda>0$, define
\begin{equation}
 \widehat F_{\lambda,Z}(w)=\frac1{sm}\sum_{i,a}f(w,z_{i,a})+\frac\lambda2\|w-w_0\|^2,
 \qquad \widehat w_{\lambda,Z}=\arg\min_{w\in\W}\widehat F_{\lambda,Z}(w).
 \label{eq:main-erm}
\end{equation}
The population-localization framework of \citet{asi2024heavy}, as abstracted in \citet[Theorem~2.1]{lowy2026pure}, reduces the problem to a private solver $A_{\rm ERM}$ for~\eqref{eq:main-erm} satisfying
\begin{equation}
 \Pr\bigl(\|A_{\rm ERM}(Z;w_0,\lambda)-\widehat w_{\lambda,Z}\|
 \le C_k\Lambda^4\mathcal E_{s,m}(G_2,G_k)/\lambda\bigr)\ge0.8.
 \label{eq:main-erm-target}
\end{equation}
Here $\mathcal E_{s,m}(G_2,G_k)$ denotes the rate
in~\eqref{eq:main-rate} with $n$ replaced by $s$ and
moment bounds $G_2,G_k$.

The population localization meta-Algorithm~\ref{alg:poploc}
partitions the $n$ users into disjoint blocks. At phase $i$,
it runs $A_{\rm ERM}$ on fresh blocks centered at the preceding
output, aggregates the estimates, and increases the
regularization parameter.

\begin{theorem}[Population localization~\citep{asi2024heavy,lowy2026pure}]
\label[theorem]{thm:main-poploc-meta}
Let $A_{\rm ERM}$ be a $(\varepsilon,\delta)$-user-level DP solver with outputs in $\W$, and suppose it satisfies~\eqref{eq:main-erm-target} for every $w_0\in\W$ and every regularization parameter used by Algorithm~\ref{alg:poploc} on each fresh $s$-user block. If $n\ge C_k\Lambda^6/\varepsilon$, then Algorithm~\ref{alg:poploc} is $(\varepsilon,\delta)$-user-level DP and, with probability at least $1-\beta$,
\begin{equation}
 F(\widehat w)-F^*\le D\min\{G_2,C_k\revisionstart \Lambda^6\revisionend{}\err(G_2,G_k)\}.
 \label{eq:main-poploc-meta}
\end{equation}
\end{theorem}

Thus, it only remains to construct a user-level DP regularized ERM solver $A_{\rm ERM}$ satisfying~\eqref{eq:main-erm-target}. 

\subsubsection{SGD with Noisy Biased Gradients}
We will instantiate Algorithm~\ref{alg:poploc} with a user-level DP variant of SGD for solving \eqref{eq:main-erm}. Our SGD method will involve biased, noisy estimates of the gradient $\nabla \widehat{F}_Z(w_t)$. 
Unlike the worst-case-bias analyses of
\citet{asi2024heavy,lowy2023large}, the following lemma
controls distance to the empirical minimizer using the
\textit{average squared bias}. This avoids requiring
every gradient estimate to be accurate.

\begin{lemma}[Excess empirical risk of biased noisy SGD]
\label[lemma]{lem:main-biased-gd}
Let $\mathcal F_t$ be a filtration such that $Z,w_0,w_t$
are $\mathcal F_t$-measurable. For gradient estimates $g_t$ and updates 
\[
 w_{t+1}=\Pi_\W\!\left(
 w_t-\eta_t[g_t+\lambda(w_t-w_0)]
 \right),
\]
define the bias $b_t:=\E[g_t\mid\mathcal F_t]-\nabla\widehat F_Z(w_t)$ and noise $v_t:=g_t-\E[g_t\mid\mathcal F_t].$
Assume $\E\|\nabla\widehat F_Z(w_t)\|^2\le G_2^2$
and $\E[\|v_t\|^2\mid\mathcal F_t]\le\Sigma^2$
for every $t$.
Let $B^2=\frac1T\sum_{t=1}^T\E\|b_t\|^2$,
$\eta_t=4/(\lambda(t+1))$, and
$\bar w := 2\sum_{t=1}^Ttw_t/[T(T+1)]$. Then
with probability at least $0.9$,
\begin{equation}
 \|\bar w-\widehat w_{\lambda,Z}\|
 \lesssim
 \frac1\lambda\left(
 B+\frac{G_2+\Sigma+\lambda D}{\sqrt T}
 \right).
 \label{eq:main-gd-distance}
\end{equation}
\end{lemma}

The main remaining task is to provide 
a private estimator $g_t$ 
of $\nabla \widehat{F}_Z(w_t)$ at every iteration with bias $B$ and noise $\Sigma$ such that~\eqref{eq:main-gd-distance} implies~\eqref{eq:main-erm-target} for suitable $T$.

\subsubsection{Solving Regularized ERM: SGD with a Fused DP Gradient Estimator}

To extend our mean estimation procedure to the adaptive setting---where
$w_t$ depends on the private data---without incurring excessive clipping
bias or privacy noise, we develop a novel \textit{fused adaptive} variant
of \cref{alg:main-me}.

Fix a block $S$ of $s$ users and an initial $w_0$. At iteration $t$,
form
$\nabla\widehat F_{Z_i}(w_t):=
m^{-1}\sum_a\nabla f(w_t,z_{i,a})$
for each user $i\in S$. An AboveThreshold test checks whether users
are $\tau$-concentrated with $\tau\approx G_2/\sqrt m$.
If accepted, apply the FriendlyCore filter, let $\mathcal S_t$ be its
selected set of users, and set
$Q_t=|\mathcal S_t|^{-1}
\sum_{i\in\mathcal S_t}\nabla\widehat F_{Z_i}(w_t)$
when $\mathcal S_t\ne\varnothing$.
Keep $Q_t$ noiseless and internal; release only
\begin{equation}
 g_t:=A_t+v_t,\qquad
 A_t:=\frac1s\sum_{i=1}^s
 \Pi_{B(Q_t,R)}\nabla\widehat F_{Z_i}(w_t),
 \qquad
 v_t\sim\mathcal N(0,\sigma^2I_d).
 \label{eq:main-fused}
\end{equation}
Then use the update, step sizes, and returned weighted average
of Lemma~\ref{lem:main-biased-gd}.
\revisionstart If the AboveThreshold test is rejected or the user core $\mathcal S_t$
is empty, the algorithm halts and returns $w_0$.\revisionend{}
Algorithm~\ref{alg:fused} in Appendix~\ref{app:fused-algorithm} gives
the full user-level DP SGD algorithm that we use as our regularized
ERM solver $A_{\rm ERM}$.

\begin{proposition}[Private regularized ERM solver---informal]
\label[proposition]{prop:main-erm}
For $C_k\Lambda/\varepsilon\le s\le n$, $0<\lambda D\le32G_2\sqrt{nm}$, and $w_0\in\W$ independent of the fresh block, there are choices of parameters
in Algorithm~\ref{alg:fused} for which it is
$(\varepsilon,\delta)$-user-level DP and
satisfies~\eqref{eq:main-erm-target}.
\end{proposition}

We now discuss the main challenges and ideas of the proof
of \cref{prop:main-erm}.

\paragraph{Privacy.}
The main challenge is bounding the sensitivity of the fused gradient
estimator
$s^{-1}\sum_{i=1}^s
\Pi_{B(Q_t,R)}\nabla\widehat F_{Z_i}(w_t)$:
replacing one user changes the clipping center $Q_t$---which can in
turn change every clipped gradient---while also shifting the final
average of clipped gradients.

The FriendlyCore coupling of \citet{asi2024user} controls the movement
of $Q_t$: with high probability (accounted for in $\delta$), its movement
is bounded by $\widetilde O(\tau/s)$.
Further, moving $Q_t$ by $h$ changes each clipped gradient by at most $2h$,
so the contribution of the changed clipping center to the average of the
clipped user-gradients is $\widetilde O(\tau/s)$.
Combining this with the changed user's $O(R/s)$ contribution
yields a $\widetilde O((\tau+R)/s)$ sensitivity bound for the fused estimator. 

\paragraph{Utility.}
The key challenge is bounding the empirical \textit{bias}
\begin{equation}
 \label{eq:b_t}
 b_t=\frac1s\sum_{i=1}^s
 \left[
 \Pi_{B(Q_t,R)}\left(\nabla\widehat F_{Z_i}(w_t)\right)
 -\nabla\widehat F_{Z_i}(w_t)
 \right]
 =A_t-\nabla\widehat F_Z(w_t)
\end{equation}
of adaptive iterates \revisionstart (before early termination)\revisionend{}:
we reuse the same users at every iteration, so $w_t$ depends on their
data. Conditioning on $w_t$ therefore breaks independence and we cannot
directly apply the fixed-query analysis from Section~\ref{sec:main-me}.

\begin{lemma}[Adaptive empirical clipping bias -- informal]
\label[lemma]{lem:main-adaptive-bias}
Recall the definition of $b_t$ in~\eqref{eq:b_t}. Let
$\psi_2(R):=\frac{CG_2^2}{mR}$ and\revisionstart , for $k>2$,\revisionend{}
$\psi_k(R):=C_k\left[
 \left(R+\frac{G_2^2}{mR}\right)e^{-c_kmR^2/G_2^2}
 +\frac{G_k^k}{m^{k-1}R^{k-1}}
 \right]$.
There exist algorithmic parameters such that for either fixed choice
of $j\in\{2,k\}$, with probability at least $0.97$, \revisionstart the algorithm
completes all $T$ iterations and\revisionend{}
\begin{equation}
 \frac1T\sum_{t=1}^T\|b_t\|^2
 \le C_k\left[\psi_j(R)^2+\frac{G_2^2}{sm}\right].
 \label{eq:main-adaptive}
\end{equation}
\end{lemma}

\paragraph{Proof intuition.}
With an accurate coarse center, clipping leaves user gradients near the
population gradient unchanged, so only the tails contribute to the bias.
The difficulty is that the iterates depend on the same users whose tails
we need to control. Privacy limits this dependence: it lets us compare
each user's contribution at an adaptive iterate with its contribution
when that user is replaced by independent data.

The key is to control the \emph{average squared bias}, rather than demand
a small bias at every iteration. Under heavy tails, requiring all $T$
bias bounds to hold simultaneously can incur a polynomial loss in $T$.
Our SGD bound only needs the average squared bias (c.f. \cref{lem:main-biased-gd}), and DP stability
controls this average without that loss.

\paragraph{Completing the ERM guarantee (\cref{prop:main-erm}).}
We choose the clipping radius $R$ and iteration count $T$ so that
the SGD distance bound~\eqref{eq:main-gd-distance} meets the target
in~\eqref{eq:main-erm-target}.
Increasing $R$ reduces clipping bias but requires more privacy noise.
Lemma~\ref{lem:main-adaptive-bias} controls the bias as a function of $R$.
For the noise, privacy across $T$ iterations requires a per-coordinate
standard deviation $\sigma=\widetilde O(R\sqrt T/(s\varepsilon))$.
Its contribution to the SGD distance bound is therefore
\[
\frac{\sqrt d\,\sigma}{\lambda\sqrt T}
=\widetilde O\!\left(\frac{R\sqrt d}{\lambda s\varepsilon}\right).
\]
We choose $R$ to balance this noise contribution against the clipping-bias
contribution, and then take $T$ large enough to control the remaining
optimization error. Together, these choices yield the required
distance-to-ERM guarantee~\eqref{eq:main-erm-target}.

\subsubsection{The final upper bound}
Combining Theorem~\ref{thm:main-poploc-meta} with Proposition~\ref{prop:main-erm} yields our SCO upper bound.

\begin{theorem}[Norm-moment SCO upper bound]
\label[theorem]{thm:main-sco}
Under~\eqref{eq:momentbound}, if $n\ge C_k\Lambda^{6}/\varepsilon$, Algorithm~\ref{alg:poploc} instantiated with \cref{alg:fused} as the private ERM solver is $(\varepsilon,\delta)$-user-level-DP and, with probability at least $1-\beta$, has output satisfying
\begin{equation}
 F(\widehat w)-F^*\le D\min\{G_2,C_k\revisionstart \Lambda^6\revisionend{}\err(G_2,G_k)\}.
 \label{eq:main-sco-upper}
\end{equation}
\end{theorem}

\revisionstart \paragraph{Runtime.}
SCO uses $\widetilde O_k(n^2m^2)$ sample-gradient evaluations, $\widetilde O_k(nm)$ projections onto $\W$, and $\widetilde O_k(dn^2m(n+m))$ additional arithmetic operations (Appendix~\ref{app:runtime}).\revisionend{}

\subsection{Lower bound}
\label{sec:sco-lower}
\begin{theorem}[SCO lower bound]
\label[theorem]{thm:main-sco-lower}
Under~\eqref{eq:lower-delta}, on a ball of diameter $D$, every user-level $(\varepsilon,\delta)$-DP algorithm $\mathcal{A}$ has a distribution of linear losses satisfying~\eqref{eq:momentbound} such that
\begin{align}
 \E[F(\mathcal{A}(\mathsf Z))-F^*]\ge c_kD\min\{G_2,\err(G_2,G_k)\}.
 \label{eq:main-sco-lower}
\end{align}
\end{theorem}
Take $\beta=1/(8nm)$ in Theorem~\ref{thm:main-sco}. Since $F(w)-F^*\le DG_2$ for every $w\in\W$, the failure event contributes at most $\beta DG_2\le DG_2/\sqrt{nm}$ to the expected excess risk. Thus the upper and lower bounds match in expectation up to logarithmic factors.  

\paragraph{Proof overview.}

The nonprivate term is classical
\citep{nemirovski2009robust}. For the private terms,
use linear losses: a restricted, rescaled Gaussian family
satisfies the gradient moment bounds and yields the user-level
term by fingerprinting. For the heavy-tailed term, we make nonzero Gaussian gradients rare and increase their magnitude, reducing the number of informative observations.

%% file: sections/05_directional.tex
\section{Closing the directional-moment mean-estimation gap}
\label{sec:main-directional}
\revisionstart Assume $k\ge2$ and~\eqref{eq:intro-directional-model}, and let $\alpha>0$ denote target Euclidean accuracy.\revisionend{}

\begin{theorem}[Optimal directional-moment estimation]
\label[theorem]{thm:main-directional}
There is a computationally efficient user-level $(\varepsilon,\delta)$-DP estimator with constant-probability $\ell_2$-error at most $\alpha$ using
\begin{equation}
 n=\widetilde O_k\left[\frac d{m\alpha^2}
 +\frac{d\sqrt{\log(1/\delta)}}{\alpha\varepsilon\sqrt m}
 +\frac{d\sqrt{\log(1/\delta)}}{m\varepsilon\alpha^{k/(k-1)}}
 +\frac{\log(1/\delta)}\varepsilon\right].
 \label{eq:main-directional-final}
\end{equation}
This is tight up to logarithms if $\delta\le c\min\{1/(nm),\varepsilon^2/d,(\varepsilon\min\{1,\alpha\}/d)^{C_k}\}$ for sufficiently large $C_k$.
\end{theorem}

The upper bound in~\cref{thm:main-directional} improves over \cite{agarwal2024person}, which required $n \gtrsim d/\eps$ users in addition to the user requirement in~\eqref{eq:main-directional-final}. The extra $d/\eps$ user requirement of~\citet{agarwal2024person}
comes from refining a coarse estimate: each round of their algorithm clips fresh user
averages around the current estimate, averages them, and adds noise.
Their refinement requires error at most $\widetilde O(R_t/\sqrt d)$
for clipping radius $R_t$. Since the Gaussian noise has magnitude
$\widetilde O(R_t\sqrt d/(n\eps))$, this requirement forces
$n\ge\widetilde\Omega(d/\eps)$.

In the intermediate accuracy regime $1/\sqrt m<\alpha<\sqrt{d/m}$, we modify the refinement phase of their algorithm to obtain optimal sample complexity. Our key observation is that the next estimate only needs to improve the
current error toward the target $\alpha$; it need not satisfy the
stronger $R_t/\sqrt d$ error requirement. We choose the clipping radius to
balance bias and noise using a public upper bound on the current error.
As the estimate improves, this bound decreases and permits a smaller
radius, hence less privacy noise. We prove that these error bounds
reach the target in log-log rounds.  

In the high and low accuracy regimes, we use different algorithms to obtain the optimal rate. 
For $k>2$ and $\alpha\le1/\sqrt m$, the prior estimator is already
optimal because the necessary term $d/(\alpha\eps\sqrt m)$ already
dominates $d/\eps$. For $\alpha\ge\sqrt{d/m}$, our norm-moment
estimator with $\wtG_2=\wtG_k=\sqrt d$ suffices, avoiding the cost
of initialization and refinement. At $k=2$, a direct covariance
bound provides the clipping-bias estimate, so our refinement also
covers small target errors. Appendix~\ref{app:directional} gives
the algorithms and proofs.

\revisionstart \paragraph{Runtime.}
Our estimator uses $\widetilde O_k(nmd+n^2d)$ arithmetic operations; see Appendix~\ref{app:runtime}.\revisionend{}

%% file: sections/06_conclusion.tex
\section{Conclusion}
We characterized the minimax optimal rates (up to logarithms) for three fundamental problems in $(\eps, \delta)$-user-level DP learning with heavy tails: stochastic convex optimization, mean estimation under bounded norm-moments, and mean estimation under bounded directional moments. Our bounds are tight throughout the $n \ge \wt{\Omega}(1/\eps)$ regime, and our algorithms are efficient. Our work paves the way for several interesting future directions: What are the optimal rates under pure user-level DP? Can the runtime of our algorithms be improved? Can we develop practical algorithms for training large-scale ML models with user-level DP?

\revisionstart Other directions for future work include: Empirical evaluation, tight logarithmic factors, heterogeneous users, and parameter-free algorithms.\revisionend{}

%% file: sections/07_statements.tex
\subsection*{AI use statement}
Generative AI tools were used to assist with literature searches and comparisons, develop and revise mathematical proof arguments, and identify potential gaps. They also assisted with LaTeX organization, equation formatting, and compilation checks. This assistance included some mathematical suggestions, not only copyediting. The author is responsible for the final text, mathematical claims, and citations.

\subsection*{Ethics statement}
{This paper is theoretical and uses no human-subject experiments or private empirical datasets. Its guarantees apply under the stated adjacency relation and statistical assumptions.} %

\subsection*{Reproducibility statement}
Appendices~\ref{app:mean}--\ref{app:directional} provide algorithmic definitions, parameter choices, theorem statements, and proofs, including the small-$\delta$ and user-count conditions. The source package includes all LaTeX and bibliography sources needed to reproduce this document. No experimental results are reported.

%% file: appendix/00_notation.tex
 After the additional comparisons in Appendix~\ref{app:related}, the
proofs follow the main text: Appendix~\ref{app:mean} treats norm-moment
mean estimation, Appendix~\ref{app:sco} treats SCO, and
Appendix~\ref{app:directional} treats directional-moment mean estimation.

%% file: sections/01b_related_work.tex
\section{Additional comparisons}
\label{app:related}

\paragraph{Bounded-gradient SCO.}  There is considerable work on the user-level DP SCO problem with uniformly Lipschitz loss functions (i.e.  uniformly bounded stochastic gradients).  In particular, the optimal rate for this problem is known.  \cite{levy2021user} initiated this line of work and  
\citet[Theorem~3.1]{bassily2023user} later obtained the optimal rate under smoothness assumptions when $n>\sqrt d/\varepsilon$. Their theorem also assumes
\[
 m<\max\{\sqrt d,n\varepsilon^2/\sqrt d\},\qquad
 \delta<\frac{d}{4n^{5/2}\sqrt m\,\varepsilon^2},
\]
and a parameter-dependent upper bound on the smoothness constant.  The works \citet{asi2024user,lowy2024faster} improved over \citet[Theorem~3.1]{bassily2023user}, achieving the optimal rate with a nearly dimension-independent user requirement and without any smoothness assumption.  \textit{None of these works consider unbounded or heavy tailed stochastic gradients}.

\paragraph{Comparison with \citet{narayanan2022few}.}
For $n\ge\widetilde\Omega(\sqrt d/\varepsilon)$, grouping
users into $\widetilde\Theta(\sqrt d/\varepsilon)$ blocks
and applying their Theorem~4.1 yields error
$\widetilde O(\wtG_2\sqrt{\sqrt d/(nm\varepsilon)})$.
Privacy is preserved because changing one original user
changes only one block. This bound is worse than our near-optimal bound: namely, it matches our second-moment
branch, whereas our higher-moment branch can improve
the rate when $\wtG_k$ is sufficiently small.

\paragraph{Huber-loss norm-moment estimation.}
\citet{zhao2024huber} study user-level mean estimation through Huber-loss minimization, including imbalanced user sample sizes. Their Theorem~3 gives a normalized mean-squared-error rate whose square root yields {the rate $\err(\wtG_k,\wtG_k)$ in Table~\ref{tab:mean}}, up to logarithms, after restoring one public moment scale. The theorem assumes a public bound on the mean location. Its sufficient user condition is expressed through a smooth-sensitivity parameter of order $\varepsilon/(d+\log(1/\delta))$, hence the dimension-dependent count in Table~\ref{tab:mean}.

\readingstart{item-level-comparison}\paragraph{\revisionstart Item-level DP on user averages.\revisionend{}}
\label{app:item-reduction}
Applying an item-level DP estimator to the $n$ user averages
preserves user-level privacy, but its generic moment-based
guarantee can be suboptimal. To see this, suppose
$\wtG_2=\wtG_k=G$ and let $a=\sqrt d/(n\varepsilon)<1$.
Each user average $Y$ satisfies
$(\E\|Y-\mu\|^k)^{1/k}\lesssim_k G/\sqrt m$.
Using this moment bound in an optimal item-level estimator
gives privacy error
 \[
 \frac{G}{\sqrt m}a^{1-1/k},
 \qquad\text{whereas our bound is}\qquad
 \frac{G}{\sqrt m}a+G(a/m)^{1-1/k},
\]
up to logarithmic factors and constants depending on $k$.
Both bounds also include the sampling term $G/\sqrt{nm}$.
For $k>2$, the generic reduction loses the sharper dependence
on $m$ in our heavy-tail term because it uses only a moment
bound on $Y$, rather than the concentration structure of
an average of independent observations.\readingend

%% file: appendix/03_mean_estimation.tex
 \section{Proofs for Section~\ref*{sec:main-me}: Mean estimation under norm-moment bounds}
\label{app:mean}
\input{appendix/03a_mean_upper}
\input{appendix/03b_mean_lower}

%% file: appendix/03a_mean_upper.tex
 \subsection{Upper bound}
\label{app:mean-upper}
\begin{theorem}[Restatement of Theorem~\ref{thm:main-me}]
\label[theorem]{thm:mean}\label{eq:mean-final}
For every $P\in\cP_k(\wtG_2,\wtG_k)$ and $n\ge C_k\Lambda^4/\varepsilon$, there exist algorithmic parameters such that Algorithm~\ref{alg:main-me} is user-level $(\varepsilon,\delta)$-DP and satisfies
$\|\widehat\mu-\mu\|\le\widetilde O_k(\err(\wtG_2,\wtG_k))$
with probability at least $1-\beta$.
\end{theorem}

 We give the FriendlyCore subroutine used in Algorithm~\ref{alg:main-me},
then prove the theorem using the coarse-center and clipping-bias bounds
from Section~\ref{sec:main-me}.  Their proofs follow in the same order.
 \subsubsection{FriendlyCore subroutine}

\begin{algorithm}[H]
\caption{Coarse FriendlyCore estimator \citep{tsfadia2022friendlycore,asi2024user}}
\label{alg:friendlycore}\label{alg:coarse}
\textbf{Procedure:}
$\mathsf{FriendlyCore}((Y_i)_{i=1}^{n_1};\,\tau,\varepsilon,\delta)$;
$n_1$ is the number of supplied user averages and $\tau$ is the radius.
\begin{enumerate}[leftmargin=*,itemsep=2pt,topsep=3pt]
\item \textit{Public size check.} Set $C_{\rm FC}=1000$ and $L_{\rm FC}=\log(16/\delta)$.
If $n_1<C_{\rm FC}L_{\rm FC}/\varepsilon$, return $0$.

\item \textit{Concentration test.} Compute
$s_{\rm conc}=n_1^{-1}\sum_{i,j=1}^{n_1}
\1\{\|Y_i-Y_j\|\le\tau\}$.
Add $\operatorname{Lap}(20/\varepsilon)$ noise;
if the noisy score is below $4n_1/5$, return $0$.

\item \textit{Outlier removal.} For each $i \in [n_1]$, set
$h_i=\sum_{j=1}^{n_1}\1\{\|Y_i-Y_j\|\le2\tau\}$ and
\[
p_i=
\begin{cases}
0, & h_i<n_1/2,\\
(h_i-n_1/2)/(n_1/6), & n_1/2\le h_i<2n_1/3,\\
1, & h_i\ge2n_1/3.
\end{cases}
\]

\item Starting with $\cS=\varnothing$, independently add each
$i$ to $\cS$ with probability $p_i$.

\item If $\cS=\varnothing$, return $0$.
Otherwise let $Q=|\cS|^{-1}\sum_{i\in\cS}Y_i$.

\item Return $\widetilde Q=Q+u$, where
$u\sim\mathcal N(0,\sigma_1^2I_d)$ is independent and
\[
\sigma_1=\frac{C_{\rm FC}\tau L_{\rm FC}^{3/2}}
{n_1\varepsilon}.
\]
\end{enumerate}
\end{algorithm}

 \readingstart{privacy-calibration}\begin{lemma}[{{Privacy of Algorithm~\ref{alg:friendlycore}; adapted from~\citealp[Lemma~3.6]{asi2024user}}}]
\label[lemma]{lem:fc-privacy}
Algorithm~\ref{alg:friendlycore}, with $n_1$ input user averages, is $(\varepsilon,\delta)$-user-level DP on every dataset.
\end{lemma}
\begin{proof}
Write $N=n_1$ and $L=\log(16/\delta)$, and assume the public size check
passes. For neighboring inputs with concentration scores $q,q'$, we have
$|q-q'|\le2$, so the test is $(\eps/10,0)$-DP.

Suppose $\min\{q,q'\}\ge7N/10$. We use the shared-uniform coupling
of the filtering decisions from~\citet[Lemma~3.5 and proof of Lemma~3.6]{asi2024user}.
Both selected sets have size at least $7N/10$, and
$\sum_i|p_i-p'_i|\le7$: each unchanged user's probability changes by at
most $6/N$. Thus $D_{\rm sel}=|\cS\triangle\cS'|$ satisfies
$\Pr\{D_{\rm sel}>7(e-1)+L\}\le e^{-L}=\delta/16$.
An anchor with $7N/10$ neighbors within $\tau$ has more than $N/2$
unchanged neighbors. Every selected point on either input has more than
$N/2$ neighbors within $2\tau$, and hence lies within $3\tau$ of this
anchor. Subtracting the anchor in the two averages therefore gives,
outside the coupling's exceptional event,
\[
\|Q-Q'\|\le\frac{6\tau(D_{\rm sel}+1)}{\min\{|\cS|,|\cS'|\}}
\le\frac{64\tau L}{N}=:\Delta_Q.
\]
The Gaussian mechanism with parameters $(\eps/2,\delta/16)$ requires
 \[
\frac{2\Delta_Q\sqrt{2\log(20/\delta)}}{\eps}
\le\frac{1000\tau L^{3/2}}{N\eps}=\sigma_1.
\]
Integrating over the coupling gives $(\eps/2,\delta/8)$-privacy for
the noisy center; composing with the test gives $(3\eps/5,\delta/8)$.

If $\min\{q,q'\}<7N/10$, both scores are below $7N/10+2$, so either
input passes the test with probability at most
$\frac12\exp[-\eps(N/10-2)/20]\le\delta/16$.
For this fixed neighboring pair, suppressing the release after acceptance
in this case gives an auxiliary mechanism that is
$(3\eps/5,\delta/8)$-DP and differs from the original on either input
with probability at most $\delta/16$.
Consequently, the original mechanism is $(\eps,\delta)$-DP, since
$\delta/8+(1+e^\eps)\delta/16\le\delta$.
\end{proof}\readingend

\paragraph{{The two-stage estimator.}}
For user $i$, let $Y_i=m^{-1}\sum_a z_{i,a}$, so $\E Y_i=\mu$ and $\E\|Y_i-\mu\|^2\le\wtG_2^2/m$. Split the users into $S_1,S_2$ of sizes $n_1=\lfloor n/2\rfloor$, $n_2=n-n_1$. Set $\tau=2\wtG_2\sqrt{40/m}$. Run Algorithm~\ref{alg:friendlycore} on $S_1$ and call its output $c=\widetilde Q$. If $\sqrt d/(n\varepsilon)>1$, return $c$. Otherwise return
\begin{equation}
 \widehat\mu=\frac1{n_2}\sum_{i\in S_2}\Pi_{B(c,R)}Y_i+\xi,
 \quad \xi\sim\mathcal N(0,\sigma_2^2I_d),\quad
 \sigma_2=\frac{8R\sqrt{\log(e/\delta)}}{n_2\varepsilon}.
 \label{eq:final-mean-estimator}
\end{equation}

 \subsubsection{Proof of Theorem~\ref*{thm:main-me}}
\begin{proof}[Proof of Theorem~\ref{thm:mean}]
 Allocate constant fractions of the target failure probability
$\beta$ to the events below; replacing $\beta$ by a constant
multiple only changes logarithmic factors.

 By Lemmas~\ref{lem:core-appendix}
and~\ref{lem:coarse-appendix}, with the required probability
the first-stage output $c$ satisfies
 \[
  \|c-\mu\|
  \le \widetilde O\left(
  \frac{\wtG_2}{\sqrt m}
  +\frac{\wtG_2\sqrt{d+\log(1/\beta)}}
  {n\eps\sqrt m}
  \right).
\]
 The assumption $n\ge C_k\Lambda^4/\eps$ also ensures the
sample-size requirements of FriendlyCore and makes the
majority-core failure probability negligible.

 If $\sqrt d/(n\eps)>1$, Algorithm~\ref{alg:main-me}
returns $c$.  In this regime the second term in
$\err(\wtG_2,\wtG_k)$ is at least
$\wtG_2/\sqrt m$, so the preceding coarse-center bound already
gives
 \[
  \|c-\mu\|
  \le \widetilde O_k(\err(\wtG_2,\wtG_k)).
\]
 Hence suppose $\sqrt d/(n\eps)\le1$ and consider the
second stage.

Define
 \[
  a:=\frac{\sqrt{d\log(e/\delta)}}{n\eps},
\]
 choose
 \[
  j\in\arg\min_{\ell\in\{2,k\}}
  \wtG_\ell\left(\frac{a}{m}\right)^{1-1/\ell},
\]
 and set
 \begin{equation}
  R=\max\left\{
  C_k\Lambda^2\frac{\wtG_2}{\sqrt m},
  \frac{\wtG_j}{m^{1-1/j}}a^{-1/j}
  \right\}.
  \label{eq:R-mean}
\end{equation}
 For sufficiently large $C_k$, the coarse-center bound and
$\sqrt d/(n\eps)\le1$ imply $\|c-\mu\|\le R/2$.

 Condition on such a center $c$.  Since $c$ depends only on
$S_1$, the user averages $(Y_i)_{i\in S_2}$ remain independent.
Let
 \[
  X_i:=\Pi_{B(c,R)}Y_i .
\]
 Suppressing the conditioning on $c$ in the notation,
 \[
  \widehat\mu-\mu
  =
  \left(
  \frac1{n_2}\sum_{i\in S_2}X_i-\E X_i
  \right)
  +(\E X_i-\mu)+\xi.
\]
 Because $\mu\in B(c,R)$, nonexpansiveness of projection gives
 \[
  \E\|X_i-\mu\|^2
  \le \E\|Y_i-\mu\|^2
  \le \frac{\wtG_2^2}{m}.
\]
 Moreover, both $X_i$ and $\E X_i$ lie in $B(c,R)$, so
$\|X_i-\E X_i\|\le2R$.  Hence the Hilbert-space Bernstein
inequality of~\citet{pinelis1994optimum} gives, with probability
at least $1-\beta$,
 \[
  \left\|
  \frac1{n_2}\sum_{i\in S_2}X_i-\E X_i
  \right\|
  \lesssim
  \wtG_2\sqrt{\frac{\log(1/\beta)}{mn_2}}
  +\frac{R\log(1/\beta)}{n_2}.
\]
 Also, the standard Gaussian norm bound and the definition of
$\sigma_2$ in~\eqref{eq:final-mean-estimator} give, with
probability at least $1-\beta$,
 \[
  \|\xi\|
  \lesssim
  \frac{
  R\sqrt{\log(e/\delta)}
  \bigl(\sqrt d+\sqrt{\log(1/\beta)}\bigr)}
  {n_2\eps}.
\]

 It remains to bound $\|\E X_i-\mu\|$.
 If $j=2$, use the first bound of
Lemma~\ref{lem:clipping-bias-appendix}; if $j=k>2$, use its
second bound.  The bound
$R\ge C_k\Lambda^2\wtG_2/\sqrt m$ makes the exponential term
in the latter bound negligible compared with
$\wtG_2/\sqrt{nm}$.  Thus, in either case,
 \[
  \|\E X_i-\mu\|
  \le
  \widetilde O_k\left(
  \frac{\wtG_2}{\sqrt{nm}}
  +\frac{\wtG_j^j}{m^{j-1}R^{j-1}}
  \right).
\]
 Combining the preceding three bounds, using $n_2\asymp n$,
and observing that
$R\log(1/\beta)/n=\widetilde O(aR)$, gives
 \begin{equation}
  \|\widehat\mu-\mu\|
  \le
  \widetilde O_k\left[
  \frac{\wtG_2}{\sqrt{nm}}
  +aR
  +\frac{\wtG_j^j}{m^{j-1}R^{j-1}}
  \right].
  \label{eq:mean-final-intermediate}
\end{equation}

 We now substitute~\eqref{eq:R-mean}.  Since
 \[
  a\left(
  C_k\Lambda^2\frac{\wtG_2}{\sqrt m}
  \right)
  =
  \widetilde O\left(
  \frac{\wtG_2\sqrt d}{n\eps\sqrt m}
  \right),
\]
and
 \[
  a\left(
  \frac{\wtG_j}{m^{1-1/j}}a^{-1/j}
  \right)
  =
  \wtG_j\left(\frac{a}{m}\right)^{1-1/j},
\]
 while $R$ being at least the second quantity in
\eqref{eq:R-mean} implies
 \[
  \frac{\wtG_j^j}{m^{j-1}R^{j-1}}
  \le
  \wtG_j\left(\frac{a}{m}\right)^{1-1/j},
\]
 equation~\eqref{eq:mean-final-intermediate} becomes
 \[
  \|\widehat\mu-\mu\|
  \le
  \widetilde O_k\left[
  \frac{\wtG_2}{\sqrt{nm}}
  +\frac{\wtG_2\sqrt d}{n\eps\sqrt m}
  +\wtG_j
  \left(\frac{\sqrt d}{nm\eps}\right)^{1-1/j}
  \right],
\]
 where the $\log(1/\delta)$ factor in $a$ is absorbed by
$\widetilde O_k$.  Since $j$ minimizes $\wtG_\ell(a/m)^{1-1/\ell}$, the final term is at most a logarithmic factor times the smaller of the two terms in
\eqref{eq:main-rate}.  This proves the claimed utility bound.

 Finally, Algorithm~\ref{alg:friendlycore} is
$(\eps,\delta)$-user-level DP by
Lemma~\ref{lem:fc-privacy}.  Conditional on $c$, changing one
user in $S_2$ changes the clipped average by at most
$2R/n_2$, so the Gaussian mechanism with the stated
$\sigma_2$ is also $(\eps,\delta)$-DP.
 Since the two stages use disjoint sets of users, adaptive
parallel composition proves that Algorithm~\ref{alg:main-me}
is $(\eps,\delta)$-user-level DP.
\end{proof}
 \subsubsection{Coarse localization}
{We need a large majority of user averages near the mean, not a bound on every user average. The following elementary consequence of Chebyshev's and Hoeffding's inequalities supplies that majority.}

Define
\[
  \mathrm{GOOD}:=\{i\in S_1:\|Y_i-\mu\|\le\tau/2\}.
\]

\begin{lemma}[Restatement of Lemma~\ref{lem:core}]
\label[lemma]{lem:core-appendix}
With probability at least \(1-e^{-n_1/800}\),
\[
  |\mathrm{GOOD}|\ge0.95n_1.
\]
\end{lemma}

\begin{proof}
Chebyshev gives \(\Pp(\|Y_i-\mu\|>\tau/2)\le1/40\).  Apply Hoeffding to the independent Bernoulli indicators \(\1\{i\notin\mathrm{GOOD}\}\).
\end{proof}

{Once that majority exists, every selected point is near a good point, even though a few user averages may be arbitrarily far away. This gives the coarse-center guarantee needed for the second stage.}
\begin{lemma}[Restatement of Lemma~\ref{lem:coarse}]
\label[lemma]{lem:coarse-appendix}
Suppose
\[
  n_1\ge \frac{1000}{\eps}\log\frac{16}{\delta\beta}.
\]
Conditional on $|\mathrm{GOOD}|\ge0.95n_1$, %
with probability at least $1-2\beta$, {the output $\widetilde Q$ of Algorithm~\ref{alg:friendlycore} satisfies}
\begin{equation}
  \|\widetilde Q-\mu\|
  \le \widetilde{O}\!\left(
  \frac{\wtG_2}{\sqrt m}
  +\frac{\wtG_2\sqrt{d+\log(1/\beta)}}{n_1\eps\sqrt m}
  \right).
  \label{eq:coarse-bound-appendix}
\end{equation}
\end{lemma}

\begin{proof}
Every pair of good users is within distance \(\tau\), so the concentration score is at least
\[
  \frac{|\mathrm{GOOD}|^2}{n_1}>0.9n_1.
\]
The Laplace tail bound therefore implies that the test accepts with probability at least \(1-\beta\).

Every good user has \(h_i\ge|\mathrm{GOOD}|>2n_1/3\), hence every good user is selected.  Conversely, a selected user has \(h_i\ge n_1/2\).  Since at most \(0.05n_1\) users are bad, every selected user has a good neighbor within distance \(2\tau\).  Thus, for every \(i\in\cS\),
\[
  \|Y_i-\mu\|\le5\tau/2.
\]
Hence \(\|Q-\mu\|\le5\tau/2\).  The stated result follows from the standard Gaussian norm bound for the final FriendlyCore noise.
\end{proof}

 \subsubsection{Clipping around a good center}
 \readingstart{fuk-nagaev}\revisionstart The following standard inequality separates typical fluctuations of an average from the effect of a rare large observation. 
 \begin{lemma}[{{Hilbert-space Fuk--Nagaev inequality~\citep[Corollary~2.3]{mollenhauer2025fuk}}}]
\label[lemma]{lem:fuk-nagaev}
 Let $X_1,\ldots,X_m$ be independent, mean-zero random vectors in $\R^d$, and let $k>2$.  Suppose
$v^2=\sum_a\E\|X_a\|^2<\infty$ and $M_k=\sum_a\E\|X_a\|^k<\infty$.
 There is a constant $C_k$ depending only on $k$ such that, for every $t>0$,
 \[
\Pr\!\left(\left\|\sum_{a=1}^m X_a\right\|>t\right)
\le2\exp\!\left(-\frac{t^2}{8v^2}\right)+\frac{C_kM_k}{t^k}.
\]
 If $v=0$, the sum is zero almost surely.  In particular, for an average $Y$ of $m$ i.i.d.  samples from $\cP_k(\wtG_2,\wtG_k)$,
 \[
\Pr(\|Y-\mu\|>t)
\le2e^{-mt^2/(8\wtG_2^2)}
+\frac{C_k\wtG_k^k}{m^{k-1}t^k}.
\]
 For $k=2$ we instead use Chebyshev's inequality,
$\Pr(\|Y-\mu\|>t)\le\wtG_2^2/(mt^2)$.
\end{lemma}
 The first bound is the cited inequality specialized to a Hilbert space; the average bound follows by applying it to $X_a=z_a-\mu$ at threshold $mt$.\revisionend\readingend{}

\begin{lemma}[Restatement of Lemma~\ref{lem:main-bias}]
\label[lemma]{lem:clipping-bias-appendix}
{Let $Y$ average $m$ independent observations from $P\in\cP_k(\wtG_2,\wtG_k)$.} For a fixed center $c$, if $\|c-\mu\|\le R/2$ and $R\ge C_k\wtG_2/\sqrt m$, then
\[
 \|\E\Pi_{B(c,R)}Y-\mu\|\le C\wtG_2^2/(mR)
\]
and
\[
 \|\E\Pi_{B(c,R)}Y-\mu\|\le C_k\left[\frac{\wtG_2}{\sqrt m}e^{-c_kmR^2/\wtG_2^2}
 +\frac{\wtG_k^k}{m^{k-1}R^{k-1}}\right].
\]
\end{lemma}
\begin{proof}
 Since $\E Y=\mu$,
 \[
\|\E\Pi_{B(c,R)}Y-\mu\|
\le
\E\|\Pi_{B(c,R)}Y-Y\|.
\]
 For projection onto a Euclidean ball,
 \[
\|\Pi_{B(c,R)}Y-Y\|
=(\|Y-c\|-R)_+
\le
(\|Y-\mu\|-R/2)_+
\le
\|Y-\mu\|\mathbf 1\{\|Y-\mu\|>R/2\},
\]
 where we used $\|c-\mu\|\le R/2$. Therefore
 \begin{equation}
\|\E\Pi_{B(c,R)}Y-\mu\|
\le
\E\!\left[
\|Y-\mu\|\mathbf 1\{\|Y-\mu\|>R/2\}
\right].
\label{eq:mean-bias-tail}
\end{equation}

 For $V:=\|Y-\mu\|$, the second-moment bound gives
 \[
 \E[V\mathbf1\{V>R/2\}]
 \le\frac{2\E V^2}{R}
 \le\frac{2\wtG_2^2}{mR}.
\]
 Together with~\eqref{eq:mean-bias-tail}, this proves the first claim.
 At $k=2$, it also implies the second claim because
$\wtG_k\ge\wtG_2$.  Hence assume $k>2$ for the rest of the proof.

 For any nonnegative
random variable $V$ and $a>0$,
 \begin{equation}
\E[V\mathbf 1\{V>a\}]
=
a\Pr(V>a)+\int_a^\infty \Pr(V>t)\,dt.
\label{eq:tail-integration}
\end{equation}
 Take again $V:=\|Y-\mu\|$ and set $a:=R/2$.

 \readingstart{fuk-nagaev}By Lemma~\ref{lem:fuk-nagaev} (the Fuk--Nagaev inequality)\readingend
\citep[Corollary~2.3]{mollenhauer2025fuk}, applied to the
sum of the $m$ independent centered observations,
 \begin{equation}
\Pr(V>t)
\le
C_k\left[
e^{-c_kmt^2/\wtG_2^2}
+
\frac{\wtG_k^k}{m^{k-1}t^k}
\right].
\label{eq:mean-fuk-nagaev}
\end{equation}
 Substituting~\eqref{eq:mean-fuk-nagaev}
into~\eqref{eq:tail-integration} gives
 \[
\begin{aligned}
\E[V\mathbf 1\{V>a\}]
\le C_k\Bigg[
& a e^{-c_kma^2/\wtG_2^2}
 +\int_a^\infty e^{-c_kmt^2/\wtG_2^2}\,dt\\
&+\frac{\wtG_k^k}{m^{k-1}}
\left(
\frac{1}{a^{k-1}}
+\int_a^\infty\frac{1}{t^k}\,dt
\right)
\Bigg].
\end{aligned}
\]
 For the polynomial part,
 \[
\frac{\wtG_k^k}{m^{k-1}}
\left(
\frac{1}{a^{k-1}}
+\int_a^\infty\frac{1}{t^k}\,dt
\right)
=
\frac{k}{k-1}
\frac{\wtG_k^k}{m^{k-1}a^{k-1}}
\lesssim_k
\frac{\wtG_k^k}{m^{k-1}R^{k-1}},
\]
 where we used $a=R/2$.  For the exponential part,
 \[
\begin{aligned}
a e^{-c_kma^2/\wtG_2^2}
+\int_a^\infty e^{-c_kmt^2/\wtG_2^2}\,dt
&\lesssim_k
\left(
a+\frac{\wtG_2^2}{ma}
\right)
e^{-c_kma^2/\wtG_2^2}\\
&\lesssim_k
\left(
R+\frac{\wtG_2^2}{mR}
\right)
e^{-c_kmR^2/\wtG_2^2},
\end{aligned}
\]
 where constants from $a=R/2$ are absorbed into
$c_k,C_k$. Thus
 \[
\E[V\mathbf 1\{V>R/2\}]
\le
C_k\left[
\left(
R+\frac{\wtG_2^2}{mR}
\right)e^{-c_kmR^2/\wtG_2^2}
+
\frac{\wtG_k^k}{m^{k-1}R^{k-1}}
\right].
\]
 Combining this with~\eqref{eq:mean-bias-tail} yields
 \begin{equation}
\|\E\Pi_{B(c,R)}Y-\mu\|
\le
C_k\left[
\left(
R+\frac{\wtG_2^2}{mR}
\right)e^{-c_kmR^2/\wtG_2^2}
+
\frac{\wtG_k^k}{m^{k-1}R^{k-1}}
\right].
\label{eq:psi-mean}
\end{equation}

 Finally, since $R\ge C_k\wtG_2/\sqrt m$, writing
$x=R\sqrt m/\wtG_2\ge C_k$ and absorbing/re-labeling constants gives
 \[
\left(
R+\frac{\wtG_2^2}{mR}
\right)e^{-c_kmR^2/\wtG_2^2}
=
\frac{\wtG_2}{\sqrt m}
\left(x+\frac1x\right)e^{-c_kx^2}
\le
C_k\frac{\wtG_2}{\sqrt m}e^{-c_k'x^2}.
\]
 Absorbing $c_k'$ into $c_k$ gives the stated second bound.
\end{proof}

%% file: appendix/03b_mean_lower.tex
\subsection{Matching lower bound}
\label{app:lower}
\label{sec:norm-lower-bounds}

 We start with notation.  Recall that  $n$ is the number of users, $m$ is the number
of samples per user, and $N:=nm$ is the total sample size.  Write $z$
for a single sample and $\mathsf Z$ for a user dataset; item-level
datasets are defined locally.  We use the centered norm-moment class
$\mathcal P_k(\wtG_2,\wtG_k)$ from the introduction, with $k\ge2$ and
$0<\wtG_2\le\wtG_k$, and assume $0<\eps\le1$.  We impose
\begin{equation}
 \delta\le c\min\left\{\frac1{nm},\frac{\eps^2}{d}\right\}
 \label{eq:lower-delta}
\end{equation}
for a sufficiently small absolute constant $c>0$.  This condition
suffices for all bounds in this section to hold: the sparse-mixture argument
requires $\delta\lesssim1/N$ and the Gaussian argument uses the condition
in~\eqref{eq:gaussian-delta}.

Recall the rate in~\eqref{eq:main-rate}:
 \begin{equation}
 \begin{aligned}
 \err(\wtG_2,\wtG_k)
 &=\frac{\wtG_2}{\sqrt{nm}}
   +\frac{\wtG_2\sqrt d}{n\eps\sqrt m}\\
 &\quad+\min\Biggl\{
   \wtG_2\left(\frac{\sqrt d}{nm\eps}\right)^{1/2},
   \wtG_k\left(\frac{\sqrt d}{nm\eps}\right)^{1-1/k}
   \Biggr\}.
 \end{aligned}
 \label{eq:norm-mean-rmse-lower}
\end{equation}

 We restate the lower bound before proving it.
 \begin{theorem}[Restatement of Theorem~\ref{thm:main-me-lower}]
\label[theorem]{thm:norm-mean-lower}
 Under~\eqref{eq:lower-delta}, for every user-level $(\eps,\delta)$-DP
estimator, there exists $P\in\mathcal P_k(\wtG_2,\wtG_k)$ such that
 \begin{equation}
 \Pr\bigl(\|\widehat\mu-\mu_P\|\ge
 c_k\err(\wtG_2,\wtG_k)\bigr)\ge\frac13,
 \label{eq:norm-mean-prob-lower}
\end{equation}
 where $c_k>0$ depends only on $k$.
\end{theorem}
 In particular, the worst-case root-mean-square error is at least
$c'_k\err(\wtG_2,\wtG_k)$.  This matches Theorem~\ref{thm:mean} up to
logarithmic factors for every ratio $\wtG_k/\wtG_2$ in that theorem's
user-count range.  The lower bound requires no minimum number of users.

 For sufficiently few users, no finite error bound holds uniformly over
the class.  By \citet[Lemma~5.1]{narayanan2022few}, if $\delta\le\eps^2$
and $n\le\log(1/\delta)/(3\eps)$, then for every finite $R>0$ there is
a point-mass distribution $P$ such that
 \begin{equation}
 \Pr(\|\widehat\mu-\mu_P\|>R)>\frac13.
 \label{eq:unbounded-location-impossibility}
\end{equation}

 We prove Theorem~\ref{thm:norm-mean-lower} by bounding its three terms
separately.  Lemma~\ref{lem:norm-nonprivate-lower} gives the sampling
term, and Lemma~\ref{lem:user-gaussian-lower} gives the Gaussian privacy
term.  For the heavy-tail term, Lemma~\ref{lem:sparse-norm-lower} uses
rare observations from the bounded hard family of~\citet{klsu19};
Lemma~\ref{lem:blocking-lower-bound} transfers this item-level bound
to users.

 \subsubsection{Lower-bound ingredients}

 We first record the standard reduction from item-level to user-level
lower bounds (see, e.g.,~\citealp{levy2021user}).

\begin{lemma}[Grouping item-level samples into users]
\label[lemma]{lem:blocking-lower-bound}
Let $\mathcal A$ be a user-level $(\eps,\delta)$-DP algorithm for $n$
users with $m$ samples per user.  Given $N=nm$ items
$Z=(z_1,\ldots,z_N)$, partition them into $n$ consecutive blocks of
size $m$ and run $\mathcal A$ on these blocks. The resulting algorithm
is item-level $(\eps,\delta)$-DP.  Consequently, every item-level lower
bound for $N$ i.i.d.  samples also applies to user-level estimation with
$n$ users and $m$ i.i.d.  samples per user, after substituting $N=nm$.
\end{lemma}

\begin{proof}
 Changing one item changes at most one block, so user-level privacy
implies item-level privacy for the constructed algorithm. If the items
are i.i.d.  from $P$, the blocks have the distribution of $n$ users,
each holding $m$ i.i.d. samples from $P$.
\end{proof}

 The sparse-mixture construction needs a constant-probability lower
bound for a bounded product family.  The next lemma converts the
mean-squared-error bound of~\citet{klsu19} to this form.

\begin{lemma}[{Consequence of \citet[Lemma~53]{klsu19}}]
\label[lemma]{lem:bounded-private-mean-lower}
Suppose $0<\eps\le1$,
 \[
 s\le\frac{\sqrt d}{10^4\eps}
 \qquad\text{and}\qquad
 \delta\le\frac1{10^5s}.
\]
For every item-level $(\eps,\delta)$-DP estimator
$\mathcal B:(\R^d)^s\to\R^d$, there is a product distribution $Q$
supported on $\{\pm1\}^d$, with $\mu_Q\in[-1/3,1/3]^d$, such that
 \begin{equation}
 \Pr_{Z\sim Q^{\otimes s},\,\mathcal B}
 \left(\|\mathcal B(Z)-\mu_Q\|_2\ge\frac{\sqrt d}{100}\right)
 \ge\frac25.
 \label{eq:bounded-private-mean-prob}
\end{equation}
\end{lemma}

\begin{proof}
 Lemma~53 of~\citet{klsu19} states that an $(\eps,3/(64S))$-DP
estimator with mean-squared error at most $\alpha^2\le d/54$ over this
family requires $S\ge d/(72\alpha\eps)$ samples.

 Suppose, toward a contradiction, that $\mathcal B$ has error less than
$\rho:=\sqrt d/100$ with probability greater than $3/5$ for every $Q$
in the family.  Take $S=1000s$ fresh samples and run $\mathcal B$
independently on $1000$ disjoint blocks.  Project each output onto
$[-1/3,1/3]^d$; this preserves privacy and cannot increase its distance
to $\mu_Q$.  Hoeffding's inequality implies that more than half of the
outputs are within $\rho$ of $\mu_Q$, except with probability $e^{-20}$.

 Use the geometric majority rule of~\citet[Fact~2]{asi2024heavy}:
return the first output with more than $500$ outputs within distance
$2\rho$, or zero if none qualifies.  When a majority is within $\rho$
of $\mu_Q$, a qualifying output exists.  Its $2\rho$-neighborhood
intersects that majority, so its error is at most $3\rho$.  The selected
output $\widehat\mu_S$ always lies in the cube, whose diameter is
$2\sqrt d/3$. Thus
 \[
 \E\|\widehat\mu_S-\mu_Q\|_2^2
 \le9\rho^2+\frac{4d}{9}e^{-20}<\frac d{1000}.
\]
 Parallel composition and post-processing preserve $(\eps,\delta)$-DP,
and $\delta\le1/(10^5s)<3/(64S)$.  Applying the cited lower bound with
$\alpha^2=d/1000$ gives
 \[
 S\ge\frac{\sqrt{1000}}{72}\frac{\sqrt d}{\eps}
 >\frac1{10}\frac{\sqrt d}{\eps}\ge1000s=S,
\]
a contradiction.
\end{proof}

 Next, we obtain the heavy-tailed term in the lower bound -- i.e.  the third term in~\eqref{eq:norm-mean-rmse-lower}.  The following lemma adapts existing private mean-estimation lower
bounds to simultaneous second and $k$-th centered norm-moment
constraints.  \citet[Proposition~4]{barber2014privacy} already treat
norm moments under approximate DP, but their privacy term involves
$d\wedge\log(1/\delta)$, rather than the tight $\sqrt d$ factor obtained below.

 Our construction follows the rare-mixture approach
of~\citet{barber2014privacy,lowy2023large}, with the nonzero
component drawn from the bounded hard family
of~\citet[Lemma~53]{klsu19}, as formulated in
Lemma~\ref{lem:bounded-private-mean-lower}.
 To transfer this family's lower bound to the mixture, we adapt
the sampling argument in the proof of~\citet[Lemma~5.3 of the
full version]{narayanan2023better}.  Although stated there for
a Gaussian base distribution, this argument applies equally
to our bounded family.  We choose the scale of the nonzero
observations to satisfy both moment constraints.

\begin{lemma}[{Sparse-mixture norm-moment item-level DP lower bound}]
\label[lemma]{lem:sparse-norm-lower}
 Fix $k\ge2$ and $0<\wtG_2\le\wtG_k$, and let
\[
 q_N:=\min\left\{1,\frac{\sqrt d}{N\eps}\right\}.
\]
 Suppose $0<\eps\le1$, $\delta\le1/(10^5N)$, and
$\sqrt d/\eps\ge C$ for a sufficiently large universal constant $C$.
 For every item-level $(\eps,\delta)$-DP estimator
$\mathcal A:(\R^d)^N\to\R^d$, there exists
$P\in\mathcal P_k(\wtG_2,\wtG_k)$ such that
 \begin{equation}
 \Pr_{Z\sim P^{\otimes N},\,\mathcal A}
 \left[
 \|\mathcal A(Z)-\mu_P\|_2
 \ge c_k\min\left\{
 \wtG_2q_N^{1/2},\wtG_kq_N^{1-1/k}
 \right\}
 \right]\ge\frac13,
 \label{eq:sparse-norm-prob-lower}
\end{equation}
 where $c_k>0$ depends only on $k$.
\end{lemma}

\begin{proof}
 We construct a mixture distribution whose nonzero observations come from the
bounded hard family in Lemma~\ref{lem:bounded-private-mean-lower}.
 We choose their probability so that $N$ mixture samples contain too
few observations from this bounded family for accurate private estimation.
 We then make the nonzero observations as large as the two moment
constraints permit.

 Fix $0<c_0\le1/(2\cdot10^6)$ and set
 \begin{equation}
 p:=c_0q_N,
 \qquad
 a:=\frac1{2\sqrt d}
 \min\left\{\wtG_2p^{-1/2},\wtG_kp^{-1/k}\right\}.
 \label{eq:sparse-mixture-scale}
\end{equation}
 For each distribution $Q$ in
Lemma~\ref{lem:bounded-private-mean-lower}, let $P_Q$ be the law of
$Z=aUX$, where $U\sim\operatorname{Bernoulli}(p)$ and $X\sim Q$
are independent. Thus
\begin{equation}
 \mu_{P_Q}=pa\mu_Q.
 \label{eq:sparse-mixture-mean}
\end{equation}
 Since $\|X\|_2=\sqrt d$ almost surely, Minkowski's and Jensen's
inequalities give, for $j\in\{2,k\}$,
 \[
 \left(\E\|Z-\mu_{P_Q}\|_2^j\right)^{1/j}
 \le 2\left(\E\|Z\|_2^j\right)^{1/j}
 =2a\sqrt d\,p^{1/j}
 \le\wtG_j.
\]
 Hence every $P_Q$ belongs to $\mathcal P_k(\wtG_2,\wtG_k)$.

 We adapt the sampling argument from the proof
of~\citet[Lemma~5.3 of the full version]{narayanan2023better}.
 Set $s:=\lceil100Np\rceil\le N$.  On input
$X=(X_1,\ldots,X_s)$, draw independent
$U_1,\ldots,U_N\sim\operatorname{Bernoulli}(p)$ and define
 \[
 J_i:=\sum_{\ell=1}^i U_\ell,
 \qquad K:=J_N.
\]
 When $K\le s$, form
 \[
 \widetilde Z_i:=
 \begin{cases}
  aX_{J_i}, & U_i=1,\\
  0, & U_i=0,
 \end{cases}
 \qquad i=1,\ldots,N.
\]
 That is, at the first index $i$ for which $U_i=1$, we set
$\widetilde Z_i=aX_1$; at the second such index, set
$\widetilde Z_i=aX_2$; and so on.  We set $\widetilde Z_i=0$
at every index for which $U_i=0$.  Define $\widetilde Z = (\widetilde Z_1, \ldots, \widetilde Z_N)$ and 
 \[
 \mathcal B(X):=
 \begin{cases}
  \mathcal A(\widetilde Z)/(pa), & K\le s,\\
  0, & K>s.
 \end{cases}
\]
 For fixed indicators with $K\le s$, changing one $X_j$ changes
at most one element of $\widetilde Z$.  When $K>s$, the output of $\mathcal{B}$
does not depend on $X$.  Thus, privacy of $\mathcal A$ and post-processing imply
that $\mathcal B$ is item-level $(\eps,\delta)$-DP.

 To compare the errors of $\mathcal A$ and $\mathcal B$, fix $Q$
and draw $X_1,\ldots,X_s$ independently from $Q$.  For the analysis
only, draw additional independent observations
$X_{s+1},\ldots,X_N\sim Q$, all independent of the indicators.
Define
 \[
 Z_i^\star:=
 \begin{cases}
  aX_{J_i}, & U_i=1,\\
  0, & U_i=0,
 \end{cases}
 \qquad i=1,\ldots,N.
\]
 Unlike $\widetilde Z$, this dataset is defined even when $K>s$.
 Conditional on the indicators, its nonzero entries are independent
draws from the distribution of $aX$ with $X\sim Q$.  Therefore
$Z^\star\sim P_Q^{\otimes N}$.

 On $K\le s$, we have $\widetilde Z=Z^\star$.  Using the same
internal randomness for $\mathcal A$ in both expressions,
\eqref{eq:sparse-mixture-mean} gives
 \[
 \|\mathcal B(X)-\mu_Q\|_2
 =\frac1{pa}\|\mathcal A(Z^\star)-\mu_{P_Q}\|_2
 \qquad\text{on }K\le s.
\]
 Consequently, for every $t>0$,
 \[
 \begin{aligned}
 \{\|\mathcal B(X)-\mu_Q\|_2\ge t\}
 &\subseteq \{K>s\}\\
 &\quad{}\cup
 \{\|\mathcal A(Z^\star)-\mu_{P_Q}\|_2\ge pa\,t\}.
 \end{aligned}
\]
 Taking probabilities and using
 \[
 \Pr(K>s)\le\frac{\E K}{s}
 =\frac{Np}{s}\le\frac1{100}
\]
 proves
 \begin{equation}
 \begin{aligned}
 &\Pr_{X\sim Q^{\otimes s},\,\mathcal B}
 \left(\|\mathcal B(X)-\mu_Q\|_2\ge t\right)\\
 &\quad\le
 \Pr_{Z\sim P_Q^{\otimes N},\,\mathcal A}
 \left(\|\mathcal A(Z)-\mu_{P_Q}\|_2\ge pa\,t\right)
 +\frac1{100}.
 \end{aligned}
 \label{eq:sparse-mixture-transfer}
\end{equation}

 Our choice of $p$ makes
$Np=c_0\min\{N,\sqrt d/\eps\}$.
 Taking $C\ge2\cdot10^4$ ensures
 \[
 s\le N,\qquad
 s\le100c_0\frac{\sqrt d}{\eps}+1
   \le\frac{\sqrt d}{10^4\eps},
 \qquad
 \delta\le\frac1{10^5N}\le\frac1{10^5s}.
\]
 Thus, the conditions of  Lemma~\ref{lem:bounded-private-mean-lower} are satisfied and there exists $Q$
for which $\mathcal B$ has error at least $\sqrt d/100$ with
probability at least $2/5$.  Applying
\eqref{eq:sparse-mixture-transfer} with $t=\sqrt d/100$ yields
 \[
 \Pr_{Z\sim P_Q^{\otimes N},\,\mathcal A}
 \left(
 \|\mathcal A(Z)-\mu_{P_Q}\|_2
 \ge\frac{pa\sqrt d}{100}
 \right)
 \ge\frac25-\frac1{100}>\frac13.
\]
 Finally, $p=c_0q_N$, $k\ge2$, and $0<c_0<1$ imply
 \begin{align*}
 \frac{pa\sqrt d}{100}
 &=\frac1{200}
 \min\left\{\wtG_2p^{1/2},\wtG_kp^{1-1/k}\right\}\\
 &\ge\frac{c_0^{1-1/k}}{200}
 \min\left\{
 \wtG_2q_N^{1/2},\wtG_kq_N^{1-1/k}
 \right\}.
\end{align*}
 Taking $c_k=c_0^{1-1/k}/200$ proves
\eqref{eq:sparse-norm-prob-lower}.
\end{proof}

 Next, we obtain the nonprivate (first) term in
\eqref{eq:norm-mean-rmse-lower}.  The following lemma follows by
scaling the Bernoulli construction in~\citet[Example~8.3.1]
{duchi2023notes}. 

 \begin{lemma}[{Nonprivate mean-estimation lower bound}]
\label[lemma]{lem:norm-nonprivate-lower}
 Fix $k\ge2$ and $0<\wtG_2\le\wtG_k$. For every estimator
$\mathcal A$ based on $N$ i.i.d.  samples, there is a distribution
$P\in\mathcal P_k(\wtG_2,\wtG_k)$ such that
 \[
 \Pr_{Z\sim P^{\otimes N},\,\mathcal A}
 \left(\|\mathcal A(Z)-\mu_P\|_2
       \ge c\frac{\wtG_2}{\sqrt N}\right)
 \ge\frac13,
\]
 where $c>0$ is a universal constant. Consequently,
 \[
 \sup_{P\in\mathcal P_k(\wtG_2,\wtG_k)}
 \left(
 \E_{Z\sim P^{\otimes N},\,\mathcal A}
 \bigl[\|\mathcal A(Z)-\mu_P\|_2^2\bigr]
 \right)^{1/2}
 \ge \frac{c\wtG_2}{\sqrt{3N}}.
\]
 The probability and expectation include the randomness of the
sample and of $\mathcal A$.
\end{lemma}

\begin{proof}
 Let $r:=\wtG_2/2$, $\beta:=1/\sqrt{24N}$, and let $e_1$
be the first coordinate vector in $\R^d$.  For
$\eta\in\{-1,1\}$, define $P_\eta$ by
 \[
 P_\eta(\{re_1\})=\frac{1+\eta\beta}{2},
 \qquad
 P_\eta(\{-re_1\})=\frac{1-\eta\beta}{2}.
\]
 Its mean is $\mu_\eta=\eta r\beta e_1$.  Since
$\|X-\mu_\eta\|_2\le2r=\wtG_2$ almost surely,
 \[
 \E_{P_\eta}\|X-\mu_\eta\|_2^j
 \le\wtG_2^j\le\wtG_j^j,
 \qquad j\in\{2,k\}.
\]
 Thus both distributions belong to
$\mathcal P_k(\wtG_2,\wtG_k)$.

 Scaling and embedding preserve total variation distance, so the
calculation in~\citet[Example~8.3.1]{duchi2023notes} gives
 \[
 \|P_+^{\otimes N}-P_-^{\otimes N}\|_{\mathrm{TV}}
 \le\beta\sqrt{\frac{3N}{2}}
 =\frac14.
\]
 The two means are separated by $2r\beta$.  An estimate with error
less than $r\beta$ therefore identifies the correct mean by
choosing the nearer of $\mu_+$ and $\mu_-$.  Le Cam's two-point
inequality yields
 \begin{align*}
 &\max_{\eta\in\{-1,1\}}
 \Pr_{Z\sim P_\eta^{\otimes N},\,\mathcal A}
 \left(\|\mathcal A(Z)-\mu_\eta\|_2\ge r\beta\right)\\
 &\qquad\ge
 \frac{1-\|P_+^{\otimes N}-P_-^{\otimes N}\|_{\mathrm{TV}}}{2}
 \ge\frac38>\frac13.
\end{align*}
 Since $r\beta=\wtG_2/(4\sqrt{6N})$, the probability bound holds
with $c=1/(4\sqrt6)$.

 For the distribution $P$ supplied by this bound,
 \begin{align*}
 &\E_{Z\sim P^{\otimes N},\,\mathcal A}
 \bigl[\|\mathcal A(Z)-\mu_P\|_2^2\bigr]\\
 &\qquad\ge
 (r\beta)^2
 \Pr_{Z\sim P^{\otimes N},\,\mathcal A}
 \left(\|\mathcal A(Z)-\mu_P\|_2\ge r\beta\right)\\
 &\qquad\ge\frac{(r\beta)^2}{3}
 =\frac{c^2\wtG_2^2}{3N}.
\end{align*}
 Taking square roots  gives
the second claim.
\end{proof}

 Finally, we prove the Gaussian privacy term, i.e.  the second term in~\eqref{eq:norm-mean-rmse-lower}.  For this, we first need a bound that allows arbitrary
mean locations and does not require the error to be small relative to
the standard deviation.  We give the corresponding user-level version
of the Gaussian fingerprinting argument of~\citet{klsu19,agarwal2024person}.

 \begin{lemma}[{User-level Gaussian mean-estimation lower bound}]
\label[lemma]{lem:unbounded-gaussian}
 Let $n,m,d\ge1$ be integers, and let $\sigma>0$ be known.
Suppose $0<\eps\le1$, $0<\delta<1/2$, and
 \begin{equation}
 \delta\bigl(\sqrt d+\sqrt{\log(e/\delta)}\bigr)\le c\eps
 \label{eq:gaussian-delta}
\end{equation}
 for a sufficiently small universal constant $c>0$.
 For each $\theta\in\R^d$, define the dataset distribution
 \[
 \mathsf P_\theta
 :=\left(\mathcal N(\theta,\sigma^2I_d)^{\otimes m}\right)^{\otimes n},
\]
 corresponding to $n$ users with $m$ independent samples each. 
 For every user-level $(\eps,\delta)$-DP estimator $\widehat\theta$,
there exists $\theta\in\R^d$ such that
 \begin{equation}
 \Pr_{\mathsf Z\sim\mathsf P_\theta,\,\widehat\theta}
 \left(
 \|\widehat\theta(\mathsf Z)-\theta\|_2
 > c'\frac{\sigma d}{n\eps\sqrt m}
 \right)>\frac13,
 \label{eq:unbounded-gaussian-prob}
\end{equation}
 where $c'>0$ is a universal constant.  Consequently, every such
estimator satisfies
 \[
 \sup_{\theta\in\R^d}
 \left(
 \E_{\mathsf Z\sim\mathsf P_\theta,\,\widehat\theta}
 \bigl[\|\widehat\theta(\mathsf Z)-\theta\|_2^2\bigr]
 \right)^{1/2}
 \ge \frac{c'}{\sqrt3}\frac{\sigma d}{n\eps\sqrt m}.
\]
 The probability and expectation include the randomness of the
dataset and of the estimator.
\end{lemma}

\begin{proof}
 We use the Gaussian fingerprinting identity to show that the
assumed probability guarantee forces a correlation that privacy
cannot permit.

 Set $\tau:=\sigma/\sqrt m$.  By~\citet[Proposition~3]{levy2021user},
the user-level estimator induces an item-level $(\eps,\delta)$-DP
estimator on $n$ observations from $\mathcal N(\theta,\tau^2I_d)$,
with the same output distribution.  Scaling the observations and
output by $1/\tau$, it suffices to prove the probability bound
$c'd/(n\eps)$ for an item-level estimator $\mathcal A$ of the mean
of $\mathcal N(\mu,I_d)$.

 Fix an arbitrary $e_0>0$, and suppose that,
 \[
 \Pr_{\mathsf X\sim\mathcal N(\mu,I_d)^{\otimes n},\,\mathcal A}
 \left(\|\mathcal A(\mathsf X)-\mu\|_2\le e_0\right)
 \ge\frac23
 \qquad\text{for every }\mu\in\R^d.
\]
 Set $a:=24e_0/\sqrt d$ and let $\mathcal M$ be $\mathcal A$
followed by Euclidean projection onto $[-a,a]^d$.  This preserves
privacy and does not increase the error for means in the cube.
 Give $\mu$ the uniform prior on $[-a,a]^d$ and draw
$\mathsf X\mid\mu\sim\mathcal N(\mu,I_d)^{\otimes n}$.
 All subsequent expectations include the prior, the samples,
and the estimator's randomness.

 When $\|\mathcal M(\mathsf X)-\mu\|_2\le e_0$, we have
$\langle\mu,\mathcal M(\mathsf X)\rangle
\ge\|\mu\|_2^2-e_0\|\mu\|_2$.
 On every outcome, this inner product is at least
$-a\|\mu\|_1$.  The assumed success probability therefore gives
 \[
 \E[\langle\mu,\mathcal M(\mathsf X)\rangle\mid\mu]
 \ge\frac23\|\mu\|_2^2-e_0\|\mu\|_2
      -\frac a3\|\mu\|_1.
\]
Define
 \[
 g_i:=X_i-\mu,\qquad
 V_i:=\sum_{j=1}^d
 (a^2-\mu_j^2)(\mathcal M_j(\mathsf X)-\mu_j)g_{i,j}.
\]
 The identity in the proof of~\citet[Lemma~59]{klsu19},
applied coordinatewise, gives
 \begin{equation}
 \begin{aligned}
 \sum_{i=1}^n\E V_i
 &=2\E\langle\mu,\mathcal M(\mathsf X)\rangle\\
 &\ge\frac{da^2}{9}-2e_0a\sqrt d
 =\frac{da^2}{36}.
 \end{aligned}
 \label{eq:gaussian-accuracy-correlation}
\end{equation}
 Here we used
$\E\|\mu\|_2^2=da^2/3$,
$\E\|\mu\|_1=da/2$,
$\E\|\mu\|_2\le a\sqrt d$, and $a\sqrt d=24e_0$.

 For the complementary privacy bound, replace $X_i$ by an
independent draw from $\mathcal N(\mu,I_d)$ and call the resulting
dataset $\mathsf X^{(i)}$.  Let $V_i'$ be defined as $V_i$ with
$\mathcal M(\mathsf X^{(i)})$ in place of $\mathcal M(\mathsf X)$.
 Conditional on $\mu$ and the replacement output, $V_i'$ is a
centered Gaussian.  Since both the output and the mean lie in
the cube,
 \[
 \E V_i'=0,\qquad
 \E|V_i'|\le2a^3\sqrt d,\qquad
 |V_i|\le2a^3\sqrt d\,\|g_i\|_2.
\]
 The truncated-statistic comparison
in~\citet[equation~(18)]{klsu19} consequently gives
 \[
 \E V_i
 \le4\eps a^3\sqrt d+2\delta B
     +2\int_B^\infty\Pr(V_i>t)\,dt
 \qquad(B>0).
\]
 Take
 \[
 t_0:=\sqrt{2\log(1/\delta)},
 \qquad
 B:=2a^3\sqrt d\,(\sqrt d+t_0).
\]
 The Gaussian norm bound
$\Pr(\|g_i\|_2>\sqrt d+u)\le e^{-u^2/2}$ implies
 \[
 \int_B^\infty\Pr(V_i>t)\,dt
 \le2a^3\sqrt d\int_{t_0}^\infty e^{-u^2/2}\,du
 \le2a^3\sqrt d\,\frac{\delta}{t_0}.
\]
 Summing over $i$ yields
 \begin{equation}
 \sum_{i=1}^n\E V_i
 \le Cna^3\sqrt d
 \left[
 \eps+\delta\bigl(\sqrt d+\sqrt{\log(e/\delta)}\bigr)
 \right].
 \label{eq:gaussian-privacy-correlation}
\end{equation}

 Comparing~\eqref{eq:gaussian-accuracy-correlation}
and~\eqref{eq:gaussian-privacy-correlation}, cancelling $a^2$,
and using~\eqref{eq:gaussian-delta} gives
 \[
 \frac d{36}
 \le Cna\sqrt d
 \left[
 \eps+\delta\bigl(\sqrt d+\sqrt{\log(e/\delta)}\bigr)
 \right]
 \le C_1n\eps e_0.
\]

Thus, for every $e_0>0$, the uniform accuracy guarantee above
can hold only if
 \[
 e_0\ge c_1\frac d{n\eps}
\]
 for a universal constant $c_1>0$.

 Set $c':=c_1/2$.  Since $c'd/(n\eps)$ is strictly below this
necessary threshold, the uniform accuracy guarantee cannot
hold at error $c'd/(n\eps)$.  Therefore there exists
$\mu\in\R^d$ such that
 \[
 \Pr_{\mathsf X\sim\mathcal N(\mu,I_d)^{\otimes n},\,\mathcal A}
 \left(
 \|\mathcal A(\mathsf X)-\mu\|_2
 > c'\frac d{n\eps}
 \right)>\frac13.
\]
 Rescaling by $\tau=\sigma/\sqrt m$ and setting $\theta=\tau\mu$
proves~\eqref{eq:unbounded-gaussian-prob}.

 For the mean $\theta$ supplied by that bound, set
$t:=c'\sigma d/(n\eps\sqrt m)$. Then
 \begin{align*}
 &\E_{\mathsf Z\sim\mathsf P_\theta,\,\widehat\theta}
 \bigl[\|\widehat\theta(\mathsf Z)-\theta\|_2^2\bigr]\\
 &\qquad\ge
 t^2\Pr_{\mathsf Z\sim\mathsf P_\theta,\,\widehat\theta}
 \left(\|\widehat\theta(\mathsf Z)-\theta\|_2\ge t\right)
 \ge\frac{t^2}{3}.
\end{align*}
 Taking square roots and then the supremum over $\theta$
proves the second claim.
\end{proof}

 The preceding Lemma~\ref{lem:unbounded-gaussian} gives a private lower bound
for Gaussian distributions with covariance $\sigma^2I_d$.
 To use it in Theorem~\ref{thm:norm-mean-lower}, we must choose
$\sigma$ so that these distributions belong to
$\mathcal P_k(\wtG_2,\wtG_k)$.
 The next lemma verifies that we can take
$\sigma\asymp_k\wtG_2/\sqrt d$ and apply the preceding result.
 At this scale, its lower bound
$\sigma d/(n\eps\sqrt m)$ becomes
$\wtG_2\sqrt d/(n\eps\sqrt m)$, giving the second term
in~\eqref{eq:norm-mean-rmse-lower}.
\begin{lemma}[User-level Gaussian norm-moment lower bound]
\label[lemma]{lem:user-gaussian-lower}
 Under the hypotheses of Theorem~\ref{thm:norm-mean-lower},
for every user-level $(\eps,\delta)$-DP estimator $\widehat\mu$,
there exists a Gaussian distribution
$P\in\mathcal P_k(\wtG_2,\wtG_k)$ such that
 \begin{equation}
 \Pr_{\mathsf Z\sim(P^{\otimes m})^{\otimes n},\,\widehat\mu}
 \left(
 \|\widehat\mu(\mathsf Z)-\mu_P\|_2
 \ge c_k\frac{\wtG_2\sqrt d}{n\eps\sqrt m}
 \right)\ge\frac13.
 \label{eq:user-gaussian-prob-lower}
\end{equation}
 For this same $P$,
 \begin{equation}
 \E_{\mathsf Z\sim(P^{\otimes m})^{\otimes n},\,\widehat\mu}
 \left[\|\widehat\mu(\mathsf Z)-\mu_P\|_2^2\right]
 \ge c'_k\frac{\wtG_2^2d}{n^2m\eps^2}.
 \label{eq:user-gaussian-mse-lower}
\end{equation}
 Here $c_k,c'_k>0$ depend only on $k$.  The probability and
expectation include the randomness of the dataset and of
the estimator.
\end{lemma}

\begin{proof}

We first choose $\sigma$ so that
$\mathcal N(\theta,\sigma^2I_d)\in\mathcal P_k(\wtG_2,\wtG_k)$
for every $\theta\in\R^d$.  We then apply
Lemma~\ref{lem:unbounded-gaussian} to obtain one of these
distributions for which the estimator has the required error.

 Let $g=(g_1,\ldots,g_d)\sim\mathcal N(0,I_d)$ and set
 \[
 K_k:=(\E|g_1|^k)^{1/k},
 \qquad
 \sigma:=\frac{\wtG_2}{K_k\sqrt d}.
\]
 Since $k\ge2$, we have $K_k\ge1$.  Convexity of
$t\mapsto t^{k/2}$ gives
 \[
 \E\|g\|_2^k
 =\E\left(\sum_{j=1}^d g_j^2\right)^{k/2}
 \le d^{k/2-1}\sum_{j=1}^d\E|g_j|^k
 =d^{k/2}K_k^k.
\]
 Thus, for every $\theta\in\R^d$, the random vector
$X=\theta+\sigma g$ satisfies
 \begin{align*}
 \E\|X-\theta\|_2^2
 &=\sigma^2d
 =\frac{\wtG_2^2}{K_k^2}
 \le\wtG_2^2,\\
 \E\|X-\theta\|_2^k
 &\le\sigma^k d^{k/2}K_k^k
 =\wtG_2^k
 \le\wtG_k^k.
\end{align*}
 Hence $\mathcal N(\theta,\sigma^2I_d)$ belongs to the required
moment class for every mean $\theta$.

 We next check the privacy condition in
Lemma~\ref{lem:unbounded-gaussian}.
 First suppose $\delta>0$.  Assumption~\eqref{eq:lower-delta}
gives $\delta\le c\eps^2/d$.  Using
$\delta\log(e/\delta)\le1$, $d\ge1$, and $\eps\le1$, we obtain
 \begin{align*}
 \delta\bigl(\sqrt d+\sqrt{\log(e/\delta)}\bigr)
 &\le\delta\sqrt d+\sqrt\delta\\
 &\le\frac{c\eps^2}{\sqrt d}
      +\frac{\sqrt c\,\eps}{\sqrt d}\\
 &\le(c+\sqrt c)\eps.
\end{align*}
 Choosing the constant $c$ in~\eqref{eq:lower-delta} sufficiently
small therefore ensures~\eqref{eq:gaussian-delta}.
 If $\delta=0$, the same application is valid with any positive
$\delta'$ satisfying~\eqref{eq:lower-delta}, since an
$(\eps,0)$-DP estimator is also $(\eps,\delta')$-DP.

We have verified all the conditions of
Lemma~\ref{lem:unbounded-gaussian}.
 Thus, Lemma~\ref{lem:unbounded-gaussian} supplies a mean $\theta\in\R^d$ such that,
with $P:=\mathcal N(\theta,\sigma^2I_d)$,
 \[
 \Pr_{\mathsf Z\sim(P^{\otimes m})^{\otimes n},\,\widehat\mu}
 \left(
 \|\widehat\mu(\mathsf Z)-\mu_P\|_2
 \ge b\frac{\sigma d}{n\eps\sqrt m}
 \right)\ge\frac13,
\]
 where $b>0$ is the universal constant in that lemma.
 Our moment calculation ensures that this $P$ belongs to
$\mathcal P_k(\wtG_2,\wtG_k)$.  Substituting
$\sigma=\wtG_2/(K_k\sqrt d)$ gives
 \[
 b\frac{\sigma d}{n\eps\sqrt m}
 =\frac{b}{K_k}\frac{\wtG_2\sqrt d}{n\eps\sqrt m}.
\]
 Thus~\eqref{eq:user-gaussian-prob-lower} holds with $c_k=b/K_k$.

 For this same $P$, the squared error is at least the square of the
displayed threshold on an event of probability at least $1/3$.  Hence
 \[
 \E_{\mathsf Z\sim(P^{\otimes m})^{\otimes n},\,\widehat\mu}
 \left[\|\widehat\mu(\mathsf Z)-\mu_P\|_2^2\right]
 \ge\frac{b^2\sigma^2d^2}{3n^2\eps^2m}
 =\frac{b^2}{3K_k^2}
   \frac{\wtG_2^2d}{n^2m\eps^2}.
\]
 This proves~\eqref{eq:user-gaussian-mse-lower} with
$c'_k=b^2/(3K_k^2)=c_k^2/3$.
\end{proof}

 The next result shows that $\Omega(\log(1/\delta)/\eps)$ users are necessary for any $(\eps, \delta)$-DP algorithm to achieve non-trivial accuracy over every distribution in $\mathcal P_k(\wtG_2,\wtG_k)$.  

\begin{lemma}[{Restatement of \citet[Lemma~5.1]{narayanan2022few}}]
\label[lemma]{lem:location-obstruction}
Suppose $0<\eps\le1$, $0<\delta<1/2$, $\delta\le\eps^2$, and
\[
 n\le\frac{\log(1/\delta)}{3\eps}.
\]
For every user-level $(\eps,\delta)$-DP mean estimator based on $n$
users with $m$ samples each, and every finite $R>0$, there exists a
point-mass distribution $P$ such that
\[
 \Pr\!\left(\|\widehat\mu-\mu_P\|>R\right)>\frac13.
\]
 Consequently, no finite error bound holds uniformly over
$\mathcal P_k(\wtG_2,\wtG_k)$, since every point mass belongs to this
class.
\end{lemma}

 \subsubsection{Proof of the mean estimation lower bound}

 We combine the non-private, Gaussian, and heavy-tailed lower
bounds obtained in the previous section. 

\begin{proof}[Proof of Theorem~\ref{thm:norm-mean-lower}]
 Fix a user-level $(\eps,\delta)$-DP estimator $\widehat\mu$.
 Write
 \[
 x:=\frac{\sqrt d}{nm\eps},
 \qquad
 T_1:=\frac{\wtG_2}{\sqrt{nm}},
 \qquad
 T_2:=\frac{\wtG_2\sqrt d}{n\eps\sqrt m},
\]
and
 \[
 T_3:=\min\left\{
 \wtG_2x^{1/2},\wtG_kx^{1-1/k}
 \right\}.
\]
 Thus $\err(\wtG_2,\wtG_k)=T_1+T_2+T_3$.

 Lemma~\ref{lem:norm-nonprivate-lower}, applied to the $nm$
independent samples, gives the lower bound for $T_1$.
 Lemma~\ref{lem:user-gaussian-lower} gives the lower bound for
$T_2$.  Both hold with probability at least $1/3$ for a suitable
distribution in $\mathcal P_k(\wtG_2,\wtG_k)$.

 To obtain the lower bound for $T_3$, let $C$ be the universal
constant required by Lemma~\ref{lem:sparse-norm-lower}.
 If $\sqrt d/\eps<C$, then
 \[
 T_3\le\wtG_2\sqrt x
 =T_1\sqrt{\frac{\sqrt d}{\eps}}
 \le\sqrt C\,T_1.
\]
 Thus the lower bound for $T_1$ also gives a lower bound for
$T_3$, with a smaller constant.

 We may therefore restrict attention to $\sqrt d/\eps\ge C$.
 If $x>1$, then, using $m\ge1$,
 \[
 T_3\le\wtG_2\sqrt x
 \le\wtG_2\sqrt m\,x
 =T_2.
\]
 In this case, the Gaussian lower bound already controls $T_3$.

 It remains to consider $\sqrt d/\eps\ge C$ and $x\le1$.
 By Lemma~\ref{lem:blocking-lower-bound}, grouping $N=nm$ items
into $n$ blocks of size $m$ turns $\widehat\mu$ into an
item-level $(\eps,\delta)$-DP estimator.  Moreover,
\eqref{eq:lower-delta} ensures $\delta\le1/(10^5N)$ when its
universal constant is sufficiently small.
 Lemma~\ref{lem:sparse-norm-lower} therefore applies.
 Since $q_N=\min\{1,x\}=x$, it gives precisely the lower bound
for $T_3$.

 We have now shown that, for each $j\in\{1,2,3\}$, there is a
distribution $P_j\in\mathcal P_k(\wtG_2,\wtG_k)$ such that
 \[
 \Pr_{\mathsf Z\sim(P_j^{\otimes m})^{\otimes n},\,\widehat\mu}
 \left(
 \|\widehat\mu(\mathsf Z)-\mu_{P_j}\|_2\ge b_kT_j
 \right)\ge\frac13,
\]
 where $b_k>0$ is a common constant depending only on $k$.
 Choosing $j_\star \in [3]$ maximizing $T_j$ and taking $P=P_{j_\star}$ 
proves the probability bound
\eqref{eq:norm-mean-prob-lower} with constant $c_k=b_k/3$.

 For the same distribution,
 \[
 \E_{\mathsf Z\sim(P^{\otimes m})^{\otimes n},\,\widehat\mu}
 \left[\|\widehat\mu(\mathsf Z)-\mu_P\|_2^2\right]
 \ge\frac{c_k^2}{3}\err(\wtG_2,\wtG_k)^2.
\]
 Taking square roots gives the
claimed root-mean-square lower bound.

 Finally, Lemma~\ref{lem:location-obstruction} gives the separate
impossibility result~\eqref{eq:unbounded-location-impossibility}.
\end{proof}

%% file: appendix/04_sco.tex
 \section{Proofs for Section~\ref*{sec:main-sco}: Stochastic convex optimization}
\label{app:sco}
 \subsection{Upper bound}
 We follow the four steps of Section~\ref{sec:main-sco}: population
localization, SGD with biased gradients, the private regularized ERM
solver, and the resulting SCO guarantee.
\input{appendix/04a_population_localization}
\input{appendix/04b_biased_sgd}
\input{appendix/04c_regularized_erm}
\input{appendix/04d_sco_upper}

\input{appendix/04e_sco_lower}

%% file: appendix/04a_population_localization.tex
\subsubsection{Population localization reduces SCO to regularized ERM}
\label{app:localization}\label{sec:sco-wrapper}
 We first briefly recall the setup and notation. Let $\mathcal W\subset\R^d$ be nonempty, compact, and convex, with
diameter $D>0$, and define $F(w)=\E_{z\sim P} f(w,z)$. Assume sample losses are
convex and differentiable on a neighborhood of $\mathcal W$, gradients
are measurable, and $F$ is finite at one point.
 For each sample, define
\begin{equation}
 {\mathcal L(z):=\sup_{w\in\mathcal W}\|\nabla f(w,z)\|.}
 \label{eq:app-lipschitz}
\end{equation}
 The moment assumption~\eqref{eq:momentbound} is
\begin{equation}
 \E_{z\sim P}\mathcal L(z)^2\le G_2^2,\qquad
 \E_{z\sim P}\mathcal L(z)^k\le G_k^k,\qquad 0<G_2\le G_k.
 \label{eq:sco-moments}
\end{equation}
For a fresh phase dataset $Z$ of $s$ users, a center $w_0\in\mathcal W$,
and $\lambda>0$, define
\[
 \begin{aligned}
 \widehat F_Z(w)&=\frac1{sm}\sum_{u=1}^s\sum_{j=1}^m f(w,z_{u,j}),\\
 \widehat F_{\lambda,Z}(w)&=\widehat F_Z(w)+\frac\lambda2\|w-w_0\|^2,
 \qquad \widehat w_{\lambda,Z}=\arg\min_{w\in\mathcal W}\widehat F_{\lambda,Z}(w).
 \end{aligned}
\]
{The logarithmic factor $\Lambda$ is the one defined in the preliminaries, using the full user count $n$, not the phase size $s$.}
{For any phase size $s\le n$, $\mathcal E_{s,m}(G_2,G_k)$ is the rate in~\eqref{eq:main-rate} with $n$ replaced by $s$ and raw moment bounds $G_2,G_k$.}
The constant $C_k$ is fixed sufficiently large in the proofs below.
 The localization meta-Theorem~\ref{thm:main-poploc-meta} requires a private ERM solver satisfying
 \begin{equation}
 \Pr\left(\|A_{\rm ERM}(Z;w_0,\lambda)-\widehat w_{\lambda,Z}\|
 \le C_k\Lambda^4\mathcal E_{s,m}(G_2,G_k)/\lambda\right)\ge0.8,
 \label{eq:erm-distance-target}
\end{equation}
 where $\Lambda = \log(enmd/(\eps \delta \beta))$ for some failure probability parameter $\beta \in (0, 1/4)$. 
The probability is over the fresh i.i.d. phase dataset and the algorithm.
 The guarantee~\eqref{eq:erm-distance-target} is required for every incoming regularization center $w_0$, which may depend on
earlier disjoint phases.  We have the following user-level analogue of the population-localization reduction of~\citet{asi2024heavy}, as presented in Theorem~2.1 of~\citet{lowy2026pure}. 

\begin{theorem}[{Restatement of Theorem~\ref{thm:main-poploc-meta}; adapted from~\citealp{asi2024heavy,lowy2026pure}}]
\label[theorem]{thm:poploc-meta-detail}
Assume~\eqref{eq:sco-moments}. Let $A_{\rm ERM}$ be a $(\eps,\delta)$-user-level DP solver with outputs in $\W$, and suppose that for every center $w_0\in\mathcal W$ and {each regularization parameter used by Algorithm~\ref{alg:poploc}}, on every fresh $s$-user block it satisfies~{ \eqref{eq:erm-distance-target}}. If $n\ge C_k\Lambda^6/\eps$, then Algorithm~\ref{alg:poploc} is $(\eps,\delta)$-user-level DP and, with probability at least $1-\beta$,
\begin{equation}
 F(\widehat w)-F^*\le D\min\bigl\{G_2,\ C_k\readingstart{logarithmic-factor}\revisionstart\Lambda^6\revisionend{}\readingend\err(G_2,G_k)\bigr\}.
 \label{eq:poploc-meta-detail}
\end{equation}
\end{theorem}

 Algorithm~\ref{alg:poploc} is a user-level variant of the item-level population localization algorithm of~\cite{asi2024heavy}.

\begin{algorithm}[H]
\caption{Population localization meta-algorithm}
\label{alg:poploc}
\textbf{Input:} Data {$\mathsf Z$}  from $n$ users with $m$ observations each; moment bounds $G_2,G_k$; privacy parameters $\varepsilon,\delta$; failure probability $\beta$; $(\eps, \delta)$-user-level-DP regularized ERM solver $A_{\rm ERM}$. Assume $n\ge C_k\Lambda^6/\varepsilon$.

\begin{enumerate}[leftmargin=*,itemsep=3pt,topsep=3pt]
\item Set
$I=1+\lceil\log_{32}\sqrt{nm}\rceil$,
$J=\lceil200\log(2I/\beta)\rceil$, and
$s=\lfloor n/(IJ)\rfloor$. Initialize $w_0\in\W$.

\item Partition the users into $IJ$ disjoint datasets
$Z^{i,j}$, each containing $s$ users and all $m$ observations
of each user. Discard unused users.

\item Set
$\Delta=6C_k\Lambda^4\mathcal E_{s,m}(G_2,G_k)$
and $\lambda_1=\Delta/D$.
If $\Delta\ge G_2$, return $w_0$.

\item For $i=1,\ldots,I$:
\begin{enumerate}[label=(\alph*),leftmargin=*,itemsep=3pt]
\item Set $\lambda_i=32^{i-1}\lambda_1$ and
$\rho_i=\Delta/(3\lambda_i)$.

\item For each $j\in[J]$, independently compute
\[
v_{i,j}\leftarrow
A_{\rm ERM}(Z^{i,j};w_{i-1},\lambda_i,\varepsilon,\delta).
\]

\item \textit{Geometric aggregation}~\citep[Fact~2]{asi2024heavy}: Set $w_i$ to the first $v_{i,j}$ with more than
$J/2$ candidates within distance $2\rho_i$.
If no such candidate exists, set $w_i=w_0$.
\end{enumerate}

\item Return $w_I$.
\end{enumerate}
\end{algorithm}
 \readingstart{geometric-aggregation}\revisionstart \begin{lemma}[{Geometric aggregation}]
\label[lemma]{lem:geometric-aggregation}
 Let $v_1,\ldots,v_J$ be points in a metric space.  If more than $J/2$ lie within distance $\rho$ of $w^*$, then there exists a candidate with more than $J/2$ candidates within distance $2\rho$, and every such candidate is within $3\rho$ of $w^*$.  If the candidates are independent and each is within $\rho$ of $w^*$ with probability at least $p>1/2$, this conclusion fails with probability at most $\exp[-2J(p-1/2)^2]$.
\end{lemma}
\begin{proof}
 Every good candidate is within $2\rho$ of every other good candidate, so a candidate satisfying the selection rule exists.  The $2\rho$ neighborhood of any selected candidate intersects the good set, since both sets have more than $J/2$ elements.  The triangle inequality gives distance at most $2\rho+\rho=3\rho$ from $w^*$.  Hoeffding's inequality bounds the probability that at most $J/2$ independent candidates are good.
\end{proof}\readingend\revisionend{}

 The proof of Theorem~\ref{thm:poploc-meta-detail} requires the following standard result about generalization of regularized ERM: 
\begin{lemma}[{Distance between empirical and population minimizers}]
\label[lemma]{lem:population-bridge}
 Fix $w_0$ independently of the fresh block $Z$.  Let $w_\lambda^{\rm pop}$
minimize $F_\lambda(w):=F(w)+\lambda\|w-w_0\|^2/2$ on $\mathcal W$. Then
 \begin{equation}
 \E_Z\|\widehat w_{\lambda,Z}-w_\lambda^{\rm pop}\|^2
 \le \frac{4G_2^2}{\lambda^2sm}.
 \label{eq:emp-pop-bridge}
\end{equation}
\end{lemma}
\begin{proof}
     Indeed, by $\lambda$-strong convexity of $F_\lambda$ and the regularized
ERM bound of~\citet[Eq.~(15), proof of Proposition~30]{lowy2023large}
applied to the $sm$ independent observations,
 \[
 \frac{\lambda}{2}\E_Z\|\widehat w_{\lambda,Z}-w_\lambda^{\rm pop}\|^2
 \le \E_Z\bigl[F_\lambda(\widehat w_{\lambda,Z})
               -F_\lambda(w_\lambda^{\rm pop})\bigr]
 \le \frac{2G_2^2}{\lambda sm}.
\]
\end{proof}

 We are now ready to prove~Theorem~\ref{thm:poploc-meta-detail}.
\begin{proof}[Proof of Theorem~\ref{thm:poploc-meta-detail}]
 Each user appears in at most one ERM call, so adaptive parallel composition
gives user-level $(\eps,\delta)$-DP.

 Suppose $\Delta<G_2$.  Conditional on the preceding phases, let $w_i^*$
be the population minimizer regularized around $w_{i-1}$ with parameter
$\lambda_i$.  Each candidate is within $\rho_i/2$ of its empirical
minimizer with probability at least $0.8$ by~\eqref{eq:erm-distance-target}.
 By Lemma~\ref{lem:population-bridge} and Markov's inequality, that
minimizer is within $\rho_i/2$ of $w_i^*$ except with probability at most
$16G_2^2/(sm\lambda_i^2\rho_i^2)\le0.01$, since
$\mathcal E_{s,m}(G_2,G_k)\ge G_2/\sqrt{sm}$ and $C_k$ is sufficiently large.
 Thus each of the $J$ independent candidates succeeds with probability
at least $0.79$.  With the stated $J$, \readingstart{geometric-aggregation}geometric
aggregation (Lemma~\ref{lem:geometric-aggregation})\readingend{} gives,
with probability at least $1-\beta$,
 \begin{equation}
 \|w_i-w_i^*\|\le3\rho_i=\Delta/\lambda_i
 \qquad\text{for all }i\in[I].
 \label{eq:wrapper-invariant}
\end{equation}
 The telescoping bound in the proof of~\citet[Proposition~2]{asi2024heavy}
now gives
 \[
 F(w_I)-F^*\lesssim
 \frac{G_2\Delta}{\lambda_I}+\lambda_1D^2+
 \frac{\Delta^2}{\lambda_1}\lesssim D\Delta,
\]
 using $\lambda_i=32^{i-1}\Delta/D$ and
$G_2/32^{I-1}\le G_2/\sqrt{nm}\le\Delta$.
 If $\Delta\ge G_2$, the same bound holds since $F(w_0)-F^*\le DG_2$.

 Finally, $n/s\le2IJ=O(\Lambda^2)$, so~\eqref{eq:main-rate} gives
 \begin{equation}
 \mathcal E_{s,m}(G_2,G_k)\le(n/s)\err(G_2,G_k),
 \qquad \Delta\lesssim_k\Lambda^6\err(G_2,G_k).
 \label{eq:localization-block-rate}
\end{equation}
 Together with $F(w)-F^*\le DG_2$ on $\mathcal W$, this proves the claim.
\end{proof}

%% file: appendix/04b_biased_sgd.tex
 \subsubsection{SGD with noisy biased gradients}
 We use a user-level DP SGD algorithm as our $\mathcal{A}_{\text{ERM}}$ to solve each regularized ERM problem. 
 To verify the ERM guarantee~\eqref{eq:erm-distance-target}, we need
to know how clipping and Gaussian noise affect the SGD output.
 The following lemma is a refinement of lemmata in~\citet{lowy2023large,asi2024heavy}.  It 
bounds the distance between the average SGD iterate and the ERM in terms of noise and \textit{average} squared bias, rather than the worst-case bias across all iterations. 

\begin{lemma}[{Restatement of Lemma~\ref{lem:main-biased-gd}}]
\label[lemma]{lem:biased-gd}
{Let $(\mathcal F_t)$ be a filtration such that $Z,w_0,w_t$
are $\mathcal F_t$-measurable. For gradient estimates $g_t$ and updates
\[
 w_{t+1}=\Pi_\W\!\left(
 w_t-\eta_t[g_t+\lambda(w_t-w_0)]
 \right),
\]
define the bias $b_t:=\E[g_t\mid\mathcal F_t]-\nabla\widehat F_Z(w_t)$
and noise $v_t:=g_t-\E[g_t\mid\mathcal F_t]$.
Assume $\E\|\nabla\widehat F_Z(w_t)\|^2\le G_2^2$ and
$\E[\|v_t\|^2\mid\mathcal F_t]\le\Sigma^2$ for every $t$.
Let $B^2=\frac1T\sum_{t=1}^T\E\|b_t\|^2$,
$\eta_t=4/(\lambda(t+1))$, and
$\bar w:=2\sum_{t=1}^Ttw_t/[T(T+1)]$.
Then, with probability at least $0.9$,
\begin{equation}
 \|\bar w-\widehat w_{\lambda,Z}\|
 \lesssim
 \frac1\lambda\left(
 B+\frac{G_2+\Sigma+\lambda D}{\sqrt T}
 \right).
 \label{eq:biased-gd-distance}
\end{equation}}
\end{lemma}

\begin{proof}
{By definition, $b_t$ is $\mathcal F_t$-measurable and $\E[v_t\mid\mathcal F_t]=0$. Nonexpansiveness of projection, strong convexity of $\widehat F_{\lambda,Z}$, and Young's inequality give}
\begin{align}
{{\widehat F_{\lambda,Z}(w_t)-\widehat F_{\lambda,Z}(\widehat w_{\lambda,Z})}}
{{\le{}}}&{{\left(\frac1{2\eta_t}-\frac\lambda4\right)
 \|w_t-\widehat w_{\lambda,Z}\|^2
 -\frac{\|w_{t+1}-\widehat w_{\lambda,Z}\|^2}{2\eta_t}}}\notag\\
&{{+\frac{\|b_t\|^2}{\lambda}
 -\langle v_t,w_t-\widehat w_{\lambda,Z}\rangle
 +\frac{\eta_t}{2}\|g_t+\lambda(w_t-w_0)\|^2.}}
\label{eq:one-step-biased-gd}
\end{align}
{Both $w_t$ and the empirical minimizer are $\mathcal F_t$-measurable, so the inner product with $v_t$ has expectation zero. Also, using the gradient-moment assumption and $\|w_t-w_0\|\le D$,}
\begin{equation}
{{\E\|g_t+\lambda(w_t-w_0)\|^2
 \lesssim G_2^2+\lambda^2D^2+\E\|b_t\|^2+\Sigma^2.}}
\label{eq:noisy-gradient-second-moment}
\end{equation}
{Multiply~\eqref{eq:one-step-biased-gd} by $t$, take expectations, and sum. For $\eta_t=4/(\lambda(t+1))$, the squared-distance terms telescope: the positive coefficient is $\lambda t(t-1)/8$ and the preceding negative coefficient is its opposite. Dropping the final nonpositive term and using~\eqref{eq:noisy-gradient-second-moment} gives}
\begin{align}
{{\sum_{t=1}^T t\,\E[\widehat F_{\lambda,Z}(w_t)-\widehat F_{\lambda,Z}(\widehat w_{\lambda,Z})]}}
{{\lesssim{}}}&{{\frac{T}{\lambda}\sum_{t=1}^T\E\|b_t\|^2}}
\notag\\&{{+\frac{T}{\lambda}(G_2^2+\Sigma^2+\lambda^2D^2).}}
\label{eq:weighted-gd-sum}
\end{align}
{Convexity and the weights defining $\bar w$ now imply}
\begin{equation}
{{\E[\widehat F_{\lambda,Z}(\bar w)-\widehat F_{\lambda,Z}(\widehat w_{\lambda,Z})]
 \lesssim \frac{B^2}{\lambda}
 +\frac{G_2^2+\Sigma^2}{\lambda T}+\frac{\lambda D^2}{T}.}}
\label{eq:biased-gd-bound}
\end{equation}
 By strong convexity and optimality of $\widehat w_{\lambda,Z}$,
 \[
 \E\|\bar w-\widehat w_{\lambda,Z}\|^2
 \le\frac2\lambda\E\bigl[\widehat F_{\lambda,Z}(\bar w)
             -\widehat F_{\lambda,Z}(\widehat w_{\lambda,Z})\bigr].
\]
 Applying Markov's inequality to the squared distance and using
\eqref{eq:biased-gd-bound} proves~\eqref{eq:biased-gd-distance}.
\end{proof}

%% file: appendix/04c_regularized_erm.tex
 \subsubsection{Solving regularized ERM with a fused private gradient estimator}
\label{app:erm}\label{app:fused-algorithm}
 Algorithm~\ref{alg:fused} gives the user-level DP ERM solver required by
population localization.  It builds on FriendlyCore and DP-SGD, clipping
user gradients around the average gradient of the users selected by the
filter and adding Gaussian noise to their clipped average.

\begin{algorithm}[H]
\caption{User-level DP SGD with the fused adaptive gradient estimator}
\label{alg:fused}
\textbf{Procedure:}
$\mathsf{FusedERM}(Z;w_0,\lambda,G_2,G_k,\varepsilon,\delta,\Lambda)$,
where $Z$ contains $s$ users with $m$ observations each.
\begin{enumerate}[leftmargin=*,itemsep=3pt,topsep=3pt]
\item Choose $T,R,\delta_0$ as in
Proposition~\ref{prop:private-erm-solver}. Set
$\tau=2\sqrt{160}e^{\varepsilon/2}G_2/\sqrt m$,
$L_0=\log(16T/\delta_0)$, and
$\sigma=C(R+\tau L_0)\sqrt T L_0/(s\varepsilon)$.
\item  Set $w_1=w_0$ and draw
$\widehat\theta=4s/5+\operatorname{Lap}(8/\varepsilon)$.
\item For $t=1,\ldots,T$:
\begin{enumerate}[label=(\alph*),leftmargin=*,itemsep=3pt]
\item  For each $i\in[s]$, compute
$y_{t,i}:=m^{-1}\sum_{a=1}^m\nabla f(w_t,z_{i,a})$ and set
 \[
 \widehat q_t=\frac1s\sum_{i,j=1}^s
 \1\{\|y_{t,i}-y_{t,j}\|\le\tau\}
 +\operatorname{Lap}(16/\varepsilon).
\]
 \revisionstart If $\widehat q_t<\widehat\theta$, halt and return $w_0$.\revisionend{}
\item  For each $i\in[s]$, set
 \[
 h_{t,i}=\sum_{j=1}^s\1\{\|y_{t,i}-y_{t,j}\|\le2\tau\},
 \qquad
 p_{t,i}=
 \begin{cases}
 0, & h_{t,i}<s/2,\\[2pt]
 \dfrac{h_{t,i}-s/2}{s/6},
    & s/2\le h_{t,i}<2s/3,\\[4pt]
 1, & h_{t,i}\ge2s/3.
 \end{cases}
\]
 Include each user $i$ in $\cS_t$ independently with probability
$p_{t,i}$.
\item  \revisionstart If $\cS_t=\varnothing$, halt and return $w_0$.\revisionend{}
 Otherwise set $Q_t=|\cS_t|^{-1}\sum_{i\in\cS_t}y_{t,i}$.
\item  Draw fresh $v_t\sim\mathcal N(0,\sigma^2I_d)$ and set
 \[
 \begin{aligned}
 g_t&=s^{-1}\sum_{i=1}^s\Pi_{B(Q_t,R)}y_{t,i}+v_t,\\
 w_{t+1}&=\Pi_\W\!\left(w_t-\frac4{\lambda(t+1)}
 [g_t+\lambda(w_t-w_0)]\right).
 \end{aligned}
\]
\end{enumerate}
\item Return $2\sum_{t=1}^Ttw_t/[T(T+1)]$.
\end{enumerate}
\end{algorithm}

 We first state the privacy and utility guarantees of Algorithm~\ref{alg:fused} for solving regularized ERM.  

\begin{proposition}[{Precise statement of Proposition~\ref{prop:main-erm}}]
\label[proposition]{prop:private-erm-solver}
 There exist a universal constant $C>0$ and a constant $C_k>0$
depending only on $k$ such that the following holds.
 \revisionstart Assume~\eqref{eq:momentbound}, $0<\eps\le1$, $0<\delta<1$,
$s\le n$, $s\ge C_k\Lambda/\eps$, and
$0<\lambda D\le32G_2\sqrt{nm}$.\revisionend{}
 \revisionstart Let $Z\sim(P^m)^s$, with the incoming center $w_0\in\W$
independent of $Z$.\revisionend{}
 In Algorithm~\ref{alg:fused}, use $C$ in the Gaussian noise scale
and set
 \[
 T=\left\lceil C_k
 \frac{G_2^2+\lambda^2D^2}{\mathcal E_{s,m}(G_2,G_k)^2}
 \right\rceil.
\]
 Define $a:=C_k\Lambda\sqrt d/(s\eps)$.
 Set $\gamma=\nu=\beta_c=1/100$,
$\ell=\lceil\log_4(4T/\beta_c)\rceil$, and
\revisionstart \begin{equation}
 \delta_0:=\min\left\{\frac\delta{10},\,
 \frac{\gamma}{4(1+e)ms^2},\,
 \frac{\beta_c}{4T\ell(40e)^\ell}\right\}.
 \label{eq:internal-delta}
\end{equation}\revisionend{}
 Select $j\in\{2,k\}$ to minimize $G_j(a/m)^{1-1/j}$ and set
\begin{equation}
 R=\max\left\{C_k\Lambda G_2/\sqrt m,\quad
 G_jm^{-(1-1/j)}a^{-1/j}\right\}.
 \label{eq:R-sco}
\end{equation}
 Then Algorithm~\ref{alg:fused} is $(\eps,\delta)$-user-level DP
and satisfies~\eqref{eq:main-erm-target}.
\end{proposition}

 We first prove the privacy guarantee.  We then establish the
lemmas needed for the utility analysis and complete the proof
of the ERM guarantee.
 \begin{proof}[Proof of the privacy part of Proposition~\ref{prop:private-erm-solver}]
 Fix $w_0$ and neighboring datasets $Z,Z'$.
 Following the proof of~\citet[Lemma~3.6]{asi2024user}, we
compare the concentration-test decisions and the noisy-gradient
releases.  AboveThreshold contributes privacy parameter $\eps/2$.
 For the gradient releases, we use a coupling of the selected
sets to bound the change in the clipping center, then apply
the Gaussian mechanism and composition.

  We prove the stronger $(\eps,\delta_0)$ guarantee for the
transcript of noisy gradients and the stopping decision.
 Here $\delta_0$ is the smaller privacy parameter chosen
in~\eqref{eq:internal-delta} and used in the algorithm's noise
calibration.  This stronger guarantee is needed in the
adaptive-query bounds below for utility; it implies $(\eps,\delta)$-privacy
because $\delta_0\le\delta$.

 For a fixed query $w$, the concentration score is
 \[
 q(Z,w):=\frac1s\sum_{i,j=1}^s
 \1\!\left\{
 \|\nabla\widehat F_{Z_i}(w)
   -\nabla\widehat F_{Z_j}(w)\|\le\tau
 \right\}.
\]
 Replacing one user changes only comparisons involving that user,
so
 \[
 |q(Z,w)-q(Z',w)|\le2.
\]
 Thus the threshold noise $\operatorname{Lap}(8/\eps)$ and
score noise $\operatorname{Lap}(16/\eps)$ implement
AboveThreshold with privacy parameter $\eps/2$
\citep[Section~3.6]{dwork2014algorithmic}.

 To compare the clipping centers on neighboring datasets, we
will need a large group of user gradients lying near a common
gradient.  We first show that passing the concentration test
ensures this property, except with a small probability.

 There is one threshold noise, with scale $8/\eps$, and at
most $T$ score noises, each with scale $16/\eps$.
 The Laplace tail bound and a union bound therefore show that,
with probability at least
 \[
 1-(T+1)e^{-\eps s/640}\ge1-\frac{\delta_0}{64},
\]
 we have
 \begin{equation}
\label{eq: good event}
 \left|\widehat\theta-\frac{4s}{5}\right|\le\frac{s}{40},
 \qquad
 |\widehat q_t-q(Z,w_t)|\le\frac{s}{40}
 \quad\text{at every reached round }t.
\end{equation}
 Here we use $s\ge C L_0/\eps$, where
$L_0=\log(16T/\delta_0)$.  \readingstart{erm-size-check}The bounds $\mathcal E_{s,m}\ge G_2/\sqrt{sm}$ and
$\lambda D\le32G_2\sqrt{nm}$ imply $T=O_k((nm)^2)$.
Consequently, \eqref{eq:internal-delta} gives
$L_0=\log(16T/\delta_0)=O_k(\Lambda)$, so
$s\ge C_k\Lambda/\eps$ implies the required
$s\ge CL_0/\eps$ for sufficiently large $C_k$.\readingend

 On the good event~\eqref{eq: good event}, 
 \[
 \widehat q_t-q(Z,w_t)\le\frac{s}{40},
 \qquad
 \widehat\theta-\frac{4s}{5}\ge-\frac{s}{40}.
\]
 Equivalently,
 \[
 q(Z,w_t)\ge\widehat q_t-\frac{s}{40},
 \qquad
 \widehat\theta\ge\frac{4s}{5}-\frac{s}{40}.
\]
 If the round passes the test, then
$\widehat q_t\ge\widehat\theta$.  Combining these inequalities gives
 \[
 q(Z,w_t)
 \ge\widehat q_t-\frac{s}{40}
 \ge\widehat\theta-\frac{s}{40}
 \ge\frac{4s}{5}-\frac{2s}{40}
 =\frac{3s}{4}.
\]
 By definition, $q(Z,w_t)$ is the average number of user
gradients within distance $\tau$ of a user gradient.
 Thus, at every accepted query, there is an index $i\in[s]$
such that
 \[
 \#\left\{j\in[s]:
 \left\|\nabla\widehat F_{Z_i}(w_t)
       -\nabla\widehat F_{Z_j}(w_t)\right\|\le\tau
 \right\}\ge\frac{3s}{4}.
\]
 We now use this concentration property to control how the
clipping center changes when one user is replaced.
 The coupling requires concentration on both inputs, so we
first use the score's sensitivity to obtain a corresponding
bound on the neighboring dataset.

 Let $u$ be the replaced user, and fix a common transcript
preceding round $t$, so the query $w$ is the same on both inputs.
 On the event controlling the test noises, acceptance on $Z$
implies $q(Z,w)\ge3s/4$.  Since replacing one user changes the
score by at most $2$,
 \[
 q(Z',w)\ge q(Z,w)-2
 \ge\frac{3s}{4}-2
 \ge\frac{7s}{10},
\]
 where the last inequality uses $s\ge40$, as ensured by the
sample-size assumption.  Thus both inputs satisfy
 \[
 \min\{q(Z,w),q(Z',w)\}\ge\frac{7s}{10}.
\]
 We compare the gradient releases under this deterministic
score condition.  When combining the privacy guarantees below,
we account for the probability that the test-noise bound fails.

 We first show that this condition guarantees a large selected
set on each input.  Since the score is the average number of
neighbors within distance $\tau$, each dataset contains a
user whose gradient has at least $7s/10$ such neighbors.
 These neighboring gradients are pairwise within distance
$2\tau$, so each has filter count at least $7s/10>2s/3$
and is selected with probability one. Consequently,
 \[
 |\cS_t|\ge\frac{7s}{10},
 \qquad |\cS'_t|\ge\frac{7s}{10}.
\]

 For every unchanged user $i\ne u$,
 \[
 |h_{t,i}-h'_{t,i}|\le1,
 \qquad
 |p_{t,i}-p'_{t,i}|\le\frac6s.
\]
 For user $u$, the selection probabilities differ by at most one.
 Therefore,
 \[
 \sum_{i=1}^s|p_{t,i}-p'_{t,i}|\le7.
\]
 Applying~\citet[Lemma~3.4]{asi2024user} with
$\zeta=\delta_0/(64T)$ gives a coupling such that, with
probability at least $1-\zeta$,
 \[
 |\cS_t\triangle\cS'_t|\le C\log(1/\zeta).
\]

We next use the coupling to bound $\|Q_t-Q'_t\|$.
 The selected sets differ in only a few users, so it suffices
to show that all selected gradients lie near a common
reference vector.  We adapt the argument in the proof
of~\citet[Lemma~3.5]{asi2024user} to our filter thresholds.

 Choose an index $i_*$ in $Z$ whose gradient has at least
$7s/10$ neighbors within distance $\tau$.
 After excluding the replaced user $u$, at least
$7s/10-1>s/2$ of these neighbors remain unchanged between
$Z$ and $Z'$.
 Every selected user, on either input, has filter count
greater than $s/2$, and hence more than $s/2$ neighbors
within distance $2\tau$.
 These neighbors must include one of the unchanged users
within distance $\tau$ of
$\nabla\widehat F_{Z_{i_*}}(w)$.
 The triangle inequality therefore gives
 \[
 \begin{aligned}
 \|\nabla\widehat F_{Z_i}(w)
       -\nabla\widehat F_{Z_{i_*}}(w)\|
 &\le3\tau &&\text{for every }i\in\cS_t,\\
 \|\nabla\widehat F_{Z'_i}(w)
       -\nabla\widehat F_{Z_{i_*}}(w)\|
 &\le3\tau &&\text{for every }i\in\cS'_t.
 \end{aligned}
\]

We decompose the difference of the centers as
 \begin{align*}
 Q_t-Q'_t
 &=
 \frac1{|\cS_t|}
 \Biggl[
 \sum_{i\in\cS_t}
   \bigl(\nabla\widehat F_{Z_i}(w)
        -\nabla\widehat F_{Z_{i_*}}(w)\bigr)\\
 &\qquad\qquad-
 \sum_{i\in\cS'_t}
   \bigl(\nabla\widehat F_{Z'_i}(w)
        -\nabla\widehat F_{Z_{i_*}}(w)\bigr)
 \Biggr]\\
 &\quad+
 \left(\frac1{|\cS_t|}-\frac1{|\cS'_t|}\right)
 \sum_{i\in\cS'_t}
   \bigl(\nabla\widehat F_{Z'_i}(w)
        -\nabla\widehat F_{Z_{i_*}}(w)\bigr).
\end{align*}
 We will bound $\|Q_t - Q'_t\|$ on the coupling event.  For every $i\ne u$, the gradients on the two inputs agree.
 Thus the difference of sums in brackets equals
 \[
 \begin{aligned}
 &\sum_{i\in\cS_t\setminus\cS'_t}
   \bigl(\nabla\widehat F_{Z_i}(w)
        -\nabla\widehat F_{Z_{i_*}}(w)\bigr)\\
 &\quad-
 \sum_{i\in\cS'_t\setminus\cS_t}
   \bigl(\nabla\widehat F_{Z'_i}(w)
        -\nabla\widehat F_{Z_{i_*}}(w)\bigr)\\
 &\quad+
 \1\{u\in\cS_t\cap\cS'_t\}
   \bigl(\nabla\widehat F_{Z_u}(w)
        -\nabla\widehat F_{Z'_u}(w)\bigr).
 \end{aligned}
\]
 Every selected gradient is within $3\tau$ of
$\nabla\widehat F_{Z_{i_*}}(w)$, so the norm of this
expression is at most
 \[
 3\tau|\cS_t\setminus\cS'_t|
 +3\tau|\cS'_t\setminus\cS_t|
 +6\tau\,\1\{u\in\cS_t\cap\cS'_t\}
 \le 3\tau|\cS_t\triangle\cS'_t|+6\tau.
\]
 The remaining term in the decomposition satisfies
 \begin{align*}
 &\left|\frac1{|\cS_t|}-\frac1{|\cS'_t|}\right|
 \left\|
 \sum_{i\in\cS'_t}
   \bigl(\nabla\widehat F_{Z'_i}(w)
        -\nabla\widehat F_{Z_{i_*}}(w)\bigr)
 \right\|\\
 &\qquad\le
 \frac{\bigl||\cS_t|-|\cS'_t|\bigr|}
      {|\cS_t|\,|\cS'_t|}
 \cdot 3\tau|\cS'_t|
 =
 \frac{3\tau\bigl||\cS_t|-|\cS'_t|\bigr|}{|\cS_t|}.
\end{align*}
Consequently,
 \begin{align*}
 \|Q_t-Q'_t\|
 &\le
 \frac{
  3\tau|\cS_t\triangle\cS'_t|+6\tau
  +3\tau\bigl||\cS_t|-|\cS'_t|\bigr|
 }{\min\{|\cS_t|,|\cS'_t|\}}\\
 &\le
 \frac{9\tau\bigl(1+|\cS_t\triangle\cS'_t|\bigr)}
      {\min\{|\cS_t|,|\cS'_t|\}}\\
 &\le\frac{C\tau\log(1/\zeta)}s
\end{align*}
 on the coupling event.  Here we used
$\bigl||\cS_t|-|\cS'_t|\bigr|
\le|\cS_t\triangle\cS'_t|$,
the lower bounds $|\cS_t|,|\cS'_t|\ge7s/10$,
and the coupling bound on $|\cS_t\triangle\cS'_t|$.

 Now, to apply the Gaussian mechanism, we must bound the change
in the average of all clipped gradients:
 \[
 A_t=\frac1s\sum_{i=1}^s
 \Pi_{B(Q_t,R)}\nabla\widehat F_{Z_i}(w),
 \qquad
 A'_t=\frac1s\sum_{i=1}^s
 \Pi_{B(Q'_t,R)}\nabla\widehat F_{Z'_i}(w).
\]
 For an unchanged user, only the clipping center changes.
 Using
$\Pi_{B(Q,R)}x=Q+\Pi_{B(0,R)}(x-Q)$
and nonexpansiveness of projection, we obtain
 \[
 \|\Pi_{B(Q_t,R)}x-\Pi_{B(Q'_t,R)}x\|
 \le2\|Q_t-Q'_t\|.
\]
 For the replaced user, the two projected gradients lie
within distance $R$ of their respective centers, so their
distance is at most $2R+\|Q_t-Q'_t\|$.
 Averaging these bounds gives
 \begin{align*}
 \|A_t-A'_t\|
 &\le\frac{2(s-1)}s\|Q_t-Q'_t\|
      +\frac{2R+\|Q_t-Q'_t\|}{s}\\
 &\le2\|Q_t-Q'_t\|+\frac{2R}s\\
 &\le\frac{C(R+\tau L_0)}s
\end{align*}
 on the coupling event, since
$\log(1/\zeta)=\log(64T/\delta_0)=O(L_0)$.

 We now turn this bound into a privacy guarantee for one
gradient release. Set
 \[
 \eps_t=\frac{\eps}{8\sqrt{T\log(8/\delta_0)}},
 \qquad
 \delta_t=\frac{\delta_0}{64T}.
\]

 To apply the Gaussian mechanism with parameters
$(\eps_t,\delta_t)$, it suffices that
 \[
 \sigma\ge
 \frac{C(R+\tau L_0)\sqrt{\log(1/\delta_t)}}{s\eps_t}.
\]
 Substituting
$\eps_t=\eps/[8\sqrt{T\log(8/\delta_0)}]$
and $\delta_t=\delta_0/(64T)$ gives
 \begin{align*}
 \frac{R+\tau L_0}{s\eps_t}\sqrt{\log(1/\delta_t)}
 &=
 \frac{8(R+\tau L_0)
       \sqrt{T\log(8/\delta_0)\log(64T/\delta_0)}}
      {s\eps}\\
 &\lesssim
 \frac{(R+\tau L_0)\sqrt T\,L_0}{s\eps},
\end{align*}
 where we used $L_0=\log(16T/\delta_0)$.
 Thus the choice of $\sigma$ in Algorithm~\ref{alg:fused}
satisfies the required bound for sufficiently large $C$.
 For any fixed selected sets $\cS_t,\cS'_t$ satisfying
 \[
 \|A_t-A'_t\|\le\frac{C(R+\tau L_0)}s,
\]
 the Gaussian mechanism gives
 \[
 \Pr(A_t+v_t\in E\mid\cS_t,\cS'_t)
 \le
 e^{\eps_t}\Pr(A'_t+v'_t\in E\mid\cS_t,\cS'_t)
 +\delta_t
\]
 for every measurable $E\subseteq\mathbb R^d$
\citep[Theorem~3.22]{dwork2014algorithmic}.
 These conditional probabilities are over the Gaussian noise,
with the preceding transcript fixed.

 The distance bound holds with probability at least $1-\zeta$
under the selection coupling.  On the remaining pairs, we
bound the conditional probability on the left by $1$. Thus
 \begin{align*}
 \Pr(A_t+v_t\in E)
 &=
 \E\!\left[\Pr(A_t+v_t\in E\mid\cS_t,\cS'_t)\right]\\
 &\le
 e^{\eps_t}
 \E\!\left[\Pr(A'_t+v'_t\in E\mid\cS_t,\cS'_t)\right]
 +\delta_t+\zeta\\
 &=
 e^{\eps_t}\Pr(A'_t+v'_t\in E)+\delta_t+\zeta,
\end{align*}
 where the expectations are over the selection coupling.
 The same argument gives the inequality with $Z$ and $Z'$
reversed.

 To apply composition at every history, fix the neighboring
pair $Z,Z'$ and add the following stopping rule to both
executions: after an accepted test at $w$, stop before
releasing a gradient if
 \[
 \min\{q(Z,w),q(Z',w)\}<7s/10.
\]
 At a common history, this rule either stops both executions
or permits a release satisfying the preceding
$(\eps_t,\delta_t+\zeta)$ comparison.

 Advanced composition with slack $\delta_0/8$ gives privacy
parameters $(\eps/2,5\delta_0/32)$ for the gradient releases,
since
$T(\delta_t+\zeta)+\delta_0/8=5\delta_0/32$
\citep[Corollary~3.21]{dwork2014algorithmic}.
 Combining this with the $\eps/2$ guarantee for AboveThreshold,
as in the proof of~\citet[Lemma~3.6]{asi2024user}, gives
$(\eps,5\delta_0/32)$ for the modified transcripts on $Z,Z'$.

We now show that the additional stopping rule changes
the execution with probability at most $\delta_0/64$
on either input.  On the event controlling the test noises,
every accepted query satisfies
 \[
 \min\{q(Z,w),q(Z',w)\}
 \ge 3s/4-2\ge7s/10,
\]
 so the additional stopping rule never activates.
 Thus, using the same randomness, the original and modified
transcripts differ with probability at most $\delta_0/64$
on each input.  The original transcripts therefore satisfy
the privacy inequality in both directions with additive
error at most
 \[
 \frac{5\delta_0}{32}
 +\frac{(1+e^\eps)\delta_0}{64}
 <\delta_0,
\]
 where we used $\eps\le1$.

For every fixed $w_0$, the iterates and final output,
including the return of $w_0$ on early termination,
are post-processing of the transcript.
 Since $\delta_0\le\delta$, the solver is
$(\eps,\delta)$-user-level DP.
\end{proof}

 We now turn to proving the utility guarantee of
Proposition~\ref{prop:private-erm-solver}.  We first bound
the empirical clipping bias when the clipping center is accurate.

\begin{lemma}[Clipping bias under an accurate center]
\label[lemma]{lem:bias-from-tail}
 Let $Z=(Z_1,\ldots,Z_s)$ be any dataset of $s$ users with $m$
observations each.  Fix $w\in\mathcal W$ and $R>0$.
 If a center $Q$ satisfies $\|Q-\nabla F(w)\|\le R/2$, then
 \begin{equation}
 \left\|\frac1s\sum_{i=1}^s
 \Pi_{B(Q,R)}\nabla\widehat F_{Z_i}(w)
 -\nabla\widehat F_Z(w)\right\|
 \le\frac1s\sum_{i=1}^s
 \left(\|\nabla\widehat F_{Z_i}(w)-\nabla F(w)\|-R/2\right)_+.
 \label{eq:empirical-bias-tail}
\end{equation}
\end{lemma}

\begin{proof}
For each user the length of the clipping displacement is
$(\|\nabla\widehat F_{Z_i}(w)-Q\|-R)_+$, which is at most
$(\|\nabla\widehat F_{Z_i}(w)-\nabla F(w)\|-R/2)_+$
by the center assumption. Average these inequalities and apply
the triangle inequality.
\end{proof}

 We next bound the expected square of the right-hand side
of~\eqref{eq:empirical-bias-tail} at a fixed query.

 \begin{lemma}[Average squared tail contribution at a fixed query]
\label[lemma]{lem:fixed-query-empirical-tail}
 For every fixed $w\in\mathcal W$ and $j\in\{2,k\}$,
 \[
 \E\left[\left(\frac1s\sum_{i=1}^s
 (\|\nabla\widehat F_{Z_i}(w)-\nabla F(w)\|-R/2)_+\right)^2\right]
 \le\psi_j(R)^2+\frac{G_2^2}{ms},
\]
 where
 \[
 \psi_2(R)=\frac{2G_2^2}{mR},
 \qquad
 \psi_k(R)=C_k\left[
 \left(R+\frac{G_2^2}{mR}\right)e^{-c_kmR^2/G_2^2}
 +\frac{G_k^k}{m^{k-1}R^{k-1}}\right]\quad(k>2).
\]
 At $k=2$ only $\psi_2$ is used.
\end{lemma}

\begin{proof}
 For each user, define
 \[
 X_i:=\nabla\widehat F_{Z_i}(w)-\nabla F(w),
 \qquad U_i:=(\|X_i\|-R/2)_+.
\]
 We bound the mean and second moment of $U_i$, then use
independence across users to bound the second moment of their average.

 Since $U_i\le\|X_i\|$, independence of the $m$ observations
within each user and~\eqref{eq:momentbound} give
 \[
 \E U_i^2
 \le\E\|X_i\|^2
 =\frac1m\E_{z\sim P}
 \|\nabla f(w,z)-\nabla F(w)\|^2
 \le\frac{G_2^2}{m}.
\]
 For the mean, the inequality $(x-R/2)_+\le2x^2/R$ gives
 \[
 \E U_i
 \le\frac2R\E\|X_i\|^2
 \le\frac{2G_2^2}{mR} = \psi_2(R).
\]

 For $k>2$, the individual centered gradients also satisfy
 \[
 \E_{z\sim P}\|\nabla f(w,z)-\nabla F(w)\|^k
 \le2^k\E_{z\sim P}\|\nabla f(w,z)\|^k
 \le2^kG_k^k.
\]
 Thus we can apply~\eqref{eq:mean-fuk-nagaev} to their average
$X_i$.  Integrating that tail bound as in the proof of
Lemma~\ref{lem:clipping-bias-appendix} yields
 \[
 \E U_i
 \le\E\!\left[\|X_i\|\1\{\|X_i\|>R/2\}\right]
 \le C_k\left[
 \left(R+\frac{G_2^2}{mR}\right)e^{-c_kmR^2/G_2^2}
 +\frac{G_k^k}{m^{k-1}R^{k-1}}\right]
 =\psi_k(R),
\]
 where the factor $2^k$ is absorbed into $C_k$.

 Because $w$ is fixed and the user blocks are independent,
$U_1,\ldots,U_s$ are independent and identically distributed.
 Consequently, for either $j\in\{2,k\}$,
 \begin{align*}
 \E\left[\left(\frac1s\sum_{i=1}^s U_i\right)^2\right]
 &= (\E U_1)^2+\frac{\operatorname{Var}(U_1)}s\\
 &\le(\E U_1)^2+\frac{\E U_1^2}s\\
 &\le\psi_j(R)^2+\frac{G_2^2}{ms}.
\end{align*}
 This is the claimed bound.
\end{proof}

 The above lemma considers fixed queries.
 To handle private adaptive queries, we compare expectations on the
input user blocks with expectations on independent copies, using
stability of differential privacy as
in~\citet[Theorem~3.4]{bassily2023user}
and~\citet[Lemma~4.6]{asi2024user}.
 The following comparison follows by iterating the DP inequality,
as in the group-privacy argument
of~\citet[Section~2.3]{dwork2014algorithmic}.

 \begin{lemma}[{DP comparison with independent users}]
\label[lemma]{lem:bounded-stability}
 Let $\mathcal A$ be $(\eps_0,\delta_0)$-user-level DP and let
$Z=(Z_1,\ldots,Z_s)$ consist of independent user blocks, each
distributed as $P^m$.
 Fix a set $J$ of $\ell$ indices, and let $Z'_J$ be independent
copies of those blocks, independent also of $Z$ and the randomness
of $\mathcal A$.
 For any measurable $h(w,z_J)\in[0,M]$,
 \begin{equation}
 \E h(\mathcal A(Z),Z_J)
 \le e^{\ell\eps_0}\E h(\mathcal A(Z),Z'_J)
 +M\delta_0\sum_{a=0}^{\ell-1}e^{a\eps_0}.
 \label{eq:bounded-stability}
\end{equation}
\end{lemma}

\begin{proof}
 Replace the blocks $Z_J$ in the mechanism's input by the independent
copies $Z'_J$, and denote the resulting dataset by $Z^{(J)}$.
 We will use privacy to compare the two outputs while keeping
the blocks $Z_J$ used to evaluate $h$ unchanged.

 Fix realizations of $Z$ and $Z'_J$.  Since $Z$ and $Z^{(J)}$
differ in at most $\ell$ users, iterating the DP inequality gives,
for every $u\in[0,M]$,
 \begin{align*}
 \Pr_{\mathcal A}\!\left(h(\mathcal A(Z),Z_J)>u\right)
 &\le e^{\ell\eps_0}
 \Pr_{\mathcal A}\!\left(h(\mathcal A(Z^{(J)}),Z_J)>u\right)
 +\delta_0\sum_{a=0}^{\ell-1}e^{a\eps_0}.
\end{align*}
 These probabilities are over the mechanism's randomness only.
 Because $0\le h\le M$, integrating over $u\in[0,M]$ expresses
each probability integral as an expectation and yields
 \[
 \E_{\mathcal A}h(\mathcal A(Z),Z_J)
 \le e^{\ell\eps_0}
 \E_{\mathcal A}h(\mathcal A(Z^{(J)}),Z_J)
 +M\delta_0\sum_{a=0}^{\ell-1}e^{a\eps_0}.
\]

 Now average over $Z$ and $Z'_J$.  The dataset $Z^{(J)}$ consists
of the unaffected users and the independent replacement users,
so it is independent of the removed blocks $Z_J$.
 Moreover, $Z^{(J)}$ has the same distribution as $Z$.
 Thus,
 \[
 (Z^{(J)},Z_J)\stackrel{d}{=}(Z,Z'_J),
 \qquad
 \E h(\mathcal A(Z^{(J)}),Z_J)
 =\E h(\mathcal A(Z),Z'_J).
\]
 Substituting this equality into the averaged inequality
proves~\eqref{eq:bounded-stability}.
\end{proof}

 The next lemma transfers the fixed-query tail bound of Lemma~\ref{lem:fixed-query-empirical-tail} to private
adaptive queries, by leveraging Lemma~\ref{lem:bounded-stability}. 
Together with Lemma~\ref{lem:bias-from-tail}, this will allow us
to bound the clipping bias along the private iterates.

 \begin{lemma}[{Average squared tail contribution at private queries}]
\label[lemma]{lem:approx-tail-transfer}
 Assume~\eqref{eq:momentbound}, let
$Z=(Z_1,\ldots,Z_s)\sim(P^m)^s$, and fix $R>0$.
 Let $(w_1,\ldots,w_T)\in\mathcal W^T$ be the output of an
$(\eps_0,\delta_0)$-user-level DP mechanism applied to $Z$, where
$0<\eps_0\le1$, $0<\gamma<1/10$, and
$\delta_0\le\gamma/[4(1+e)ms^2]$.
 Then, for every $j\in\{2,k\}$ and $0<\nu<1$, we have with probability
at least $1-\gamma-\nu$ over $Z$ and the mechanism's randomness that
 \begin{equation}
 \frac1T\sum_{t=1}^T
 \left[\frac1s\sum_{i=1}^s
 (\|\nabla\widehat F_{Z_i}(w_t)-\nabla F(w_t)\|-R/2)_+\right]^2
 \le\frac C\nu\left[\psi_j(R)^2+\frac{G_2^2}{ms}\right],
 \label{eq:adaptive-tail-average}
\end{equation}
 where
 \[
 \psi_2(R)=\frac{2G_2^2}{mR},
 \qquad
 \psi_k(R)=C_k\left[
 \left(R+\frac{G_2^2}{mR}\right)e^{-c_kmR^2/G_2^2}
 +\frac{G_k^k}{m^{k-1}R^{k-1}}\right]\quad(k>2).
\]
 When $k=2$, only $\psi_2$ is used.
\end{lemma}

\begin{proof}
 Lemma~\ref{lem:bounded-stability} applies to bounded functions.
 To apply it, we therefore define
 \[
 U(w,Z_i):=\min\left\{
 (\|\nabla\widehat F_{Z_i}(w)-\nabla F(w)\|-R/2)_+,\,
 G_2(1+\sqrt{s/\gamma})
 \right\}.
\]
 We first show that this replacement leaves all the original
contributions unchanged with probability at least $1-\gamma$.

 For each user, Jensen's inequality and~\eqref{eq:momentbound} give
 \[
 \E_Z\left[
 \left(\frac1m\sum_{a=1}^m
 \sup_{w\in\mathcal W}\|\nabla f(w,z_{i,a})\|\right)^2
 \right]\le G_2^2.
\]
 Thus Markov's inequality and a union bound over the $s$ users imply
 \[
 \Pr_Z\left(
 \max_{i\in[s]}\frac1m\sum_{a=1}^m
 \sup_{w\in\mathcal W}\|\nabla f(w,z_{i,a})\|
 >G_2\sqrt{s/\gamma}
 \right)\le\gamma.
\]
 On the complementary event where $ \max_{i\in[s]}\frac1m\sum_{a=1}^m
 \sup_{w\in\mathcal W}\|\nabla f(w,z_{i,a})\|
 \le G_2\sqrt{s/\gamma}$, we have for every $i$ and $w\in\mathcal W$, 
 \[
 \|\nabla\widehat F_{Z_i}(w)-\nabla F(w)\|
 \le \frac1m\sum_{a=1}^m
 \sup_{v\in\mathcal W}\|\nabla f(v,z_{i,a})\|
 +G_2
 \le G_2(1+\sqrt{s/\gamma}),
\]
 using $\|\nabla F(w)\|\le G_2$.
 On this event, for every $i\in[s]$ and $w\in\mathcal W$,
 \[
 (\|\nabla\widehat F_{Z_i}(w)-\nabla F(w)\|-R/2)_+
 \le G_2(1+\sqrt{s/\gamma}).
\]
 Hence
 \begin{align*}
 U(w,Z_i)
 &=\min\left\{
 (\|\nabla\widehat F_{Z_i}(w)-\nabla F(w)\|-R/2)_+,\,
 G_2(1+\sqrt{s/\gamma})
 \right\}\\
 &=(\|\nabla\widehat F_{Z_i}(w)-\nabla F(w)\|-R/2)_+ 
\end{align*}  for every $i\in[s]$ and $w\in\mathcal W$.
 Because this equality holds for every $w\in\mathcal W$, it holds
at each $w_t$, even though $w_t$ depends on $Z$.
 Thus, with probability at least $1-\gamma$,
 \[
 \frac1T\sum_{t=1}^T
 \left(\frac1s\sum_{i=1}^s U(w_t,Z_i)\right)^2
 =
 \frac1T\sum_{t=1}^T
 \left[\frac1s\sum_{i=1}^s
 (\|\nabla\widehat F_{Z_i}(w_t)-\nabla F(w_t)\|-R/2)_+\right]^2.
\]

 We next bound
 \[
 \E\left[\left(\frac1s\sum_{i=1}^s U(w_t,Z_i)\right)^2\right].
\]
 Let $Z'_1,\ldots,Z'_s$ be independent draws from $P^m$,
independent of $Z$ and the mechanism's randomness.
 Conditional on $w_t$, the fixed-query estimates established in
the proof of Lemma~\ref{lem:fixed-query-empirical-tail} give
 \[
 \E[U(w_t,Z'_i)\mid w_t]\le\psi_j(R),
 \qquad
 \E[U(w_t,Z'_i)^2\mid w_t]\le\frac{G_2^2}{m}.
\]

 For distinct users $i,r$, conditional independence also gives
 \[
 \E[U(w_t,Z'_i)U(w_t,Z'_r)\mid w_t]\le\psi_j(R)^2.
\]
 Indeed, the fresh blocks $Z'_i$ and $Z'_r$ are independent of each other
and jointly independent of $w_t$.  They therefore remain independent
conditional on $w_t$, so, for $i\ne r$,
 \begin{align*}
 \E[U(w_t,Z'_i)U(w_t,Z'_r)\mid w_t]
 &=\E[U(w_t,Z'_i)\mid w_t]\,
   \E[U(w_t,Z'_r)\mid w_t]\\
 &\le\psi_j(R)^2.
\end{align*}

 By the assumed privacy of $(w_1,\ldots,w_T)$, each $w_t$
is $(\eps_0,\delta_0)$-user-level DP.  By definition of $U$, for every $w\in\mathcal W$ and all user
blocks $Z_i,Z_r$, the following inequalities hold deterministically: 
 \begin{align*}
 0\le U(w,Z_i)^2
 &\le G_2^2(1+\sqrt{s/\gamma})^2
 \le \frac{4G_2^2s}{\gamma},\\
 0\le U(w,Z_i)U(w,Z_r)
 &\le G_2^2(1+\sqrt{s/\gamma})^2
 \le \frac{4G_2^2s}{\gamma}.
\end{align*}
 Apply Lemma~\ref{lem:bounded-stability} to the mechanism
$Z\mapsto w_t$, with $M=4G_2^2s/\gamma$.
 Taking $J=\{i\}$, $\ell=1$, and
$h(w,z_J)=U(w,z_i)^2$ gives
 \begin{align*}
 \E[U(w_t,Z_i)^2]
 &\le e^{\eps_0}\E[U(w_t,Z'_i)^2]+M\delta_0\\
 &\le e^{\eps_0}\frac{G_2^2}{m}
      +\frac{4\delta_0G_2^2s}{\gamma}.
\end{align*}
 For $i\ne r$, taking $J=\{i,r\}$, $\ell=2$, and
$h(w,z_J)=U(w,z_i)U(w,z_r)$ gives
 \begin{align*}
 \E[U(w_t,Z_i)U(w_t,Z_r)]
 &\le e^{2\eps_0}\E[U(w_t,Z'_i)U(w_t,Z'_r)]
      +(1+e^{\eps_0})M\delta_0\\
 &\le e^{2\eps_0}\psi_j(R)^2
      +\frac{4(1+e^{\eps_0})\delta_0G_2^2s}{\gamma}.
\end{align*}
 Expanding the square and substituting these bounds gives
 \begin{align*}
 \E\left[\left(\frac1s\sum_{i=1}^s U(w_t,Z_i)\right)^2\right]
 &=\frac1{s^2}\sum_{i=1}^s\E[U(w_t,Z_i)^2]
   +\frac1{s^2}\sum_{i\ne r}
     \E[U(w_t,Z_i)U(w_t,Z_r)]\\
 &\le \frac{e^{\eps_0}G_2^2}{ms}
      +e^{2\eps_0}\psi_j(R)^2
      +\frac{4(1+e^{\eps_0})\delta_0G_2^2s}{\gamma}\\
 &\le C\left[\psi_j(R)^2+\frac{G_2^2}{ms}\right],
\end{align*}
 where the last inequality uses $\eps_0\le1$ and the assumed
bound on $\delta_0$.

 Averaging over $T$ preserves this expectation bound.
 Markov's inequality therefore gives, with probability at least
$1-\nu$,
 \[
 \frac1T\sum_{t=1}^T
 \left(\frac1s\sum_{i=1}^s U(w_t,Z_i)\right)^2
 \le\frac C\nu\left[\psi_j(R)^2+\frac{G_2^2}{ms}\right].
\]
 With probability at least $1-\gamma$,
 \[
 U(w_t,Z_i)
 =(\|\nabla\widehat F_{Z_i}(w_t)-\nabla F(w_t)\|-R/2)_+
 \qquad\text{for all }t\in[T],\ i\in[s].
\]
 By a union bound, these equalities and the preceding Markov bound
hold simultaneously with probability at least $1-\gamma-\nu$.
 On this intersection,
 \begin{align*}
 &\frac1T\sum_{t=1}^T
 \left[\frac1s\sum_{i=1}^s
 (\|\nabla\widehat F_{Z_i}(w_t)-\nabla F(w_t)\|-R/2)_+\right]^2\\
 &\qquad=
 \frac1T\sum_{t=1}^T
 \left(\frac1s\sum_{i=1}^s U(w_t,Z_i)\right)^2
 \le\frac C\nu\left[\psi_j(R)^2+\frac{G_2^2}{ms}\right].
\end{align*}
 This proves~\eqref{eq:adaptive-tail-average}.
\end{proof}

 The next lemma ensures that with high probability Algorithm~\ref{alg:fused} completes all $T$ iterations
and satisfies the center condition  $\|Q_t -\nabla F(w_t)\|\le R/2$ in Lemma~\ref{lem:bias-from-tail}
at every iteration, for $R \gtrsim G_2/\sqrt{m}$.

\begin{lemma}[{Accurate centers along the adaptive trajectory}]
\label[lemma]{lem:adaptive-good-centers}
 There is a universal constant $C>0$ such that the following holds.
 Let
$Z=(Z_1,\ldots,Z_s)\sim(P^m)^s$, and let the incoming center
$w_0\in\mathcal W$ be independent of $Z$.
 Run Algorithm~\ref{alg:fused} with $\eps=\eps_0\in(0,1]$
and iteration budget $T\ge1$.
 Assume its transcript of queried points and the stopping decision
is $(\eps_0,\delta_0)$-user-level DP.
 For $0<\beta_c<1/10$, define
 \[
 r=\frac{\sqrt{160}\,e^{\eps_0/2}G_2}{\sqrt m},
 \qquad \tau=2r,
 \qquad
 \ell=\left\lceil\log_4(4T/\beta_c)\right\rceil.
\]
Assume $R\ge10r$, $s\ge40\ell$,
$s\ge C\log(T/\beta_c)/\eps_0$, and
\begin{equation}
 \delta_0\le\frac{\beta_c}{4T\ell(40e)^\ell}.
 \label{eq:center-delta}
\end{equation}
 Then, with probability at least $1-\beta_c$ over $Z$ and the
algorithm's randomness, Algorithm~\ref{alg:fused} completes
all $T$ iterations and
 \[
 \|Q_t-\nabla F(w_t)\|\le R/2
 \qquad\text{for every }t\in[T].
\]
\end{lemma}

\begin{proof}
 We first bound the number of users whose gradients differ from
$\nabla F(w_t)$ by more than $r$.  We will then use this bound
to verify that neither stopping condition occurs and that
$Q_t$ satisfies the claimed accuracy guarantee.

 For each round $t$ that the algorithm reaches, define
 \[
 \mathcal B_t
 :=\left\{i\in[s]:
 \|\nabla\widehat F_{Z_i}(w_t)-\nabla F(w_t)\|>r
 \right\}.
\]
 Set $\mathcal B_t=\varnothing$ if round $t$ is not reached.
 For a fixed $w\in\mathcal W$, the user-gradient second-moment
bound and Chebyshev's inequality give
 \[
 \Pr_{Z'\sim P^m}\!\left(
 \|\nabla\widehat F_{Z'}(w)-\nabla F(w)\|>r
 \right)
 \le\frac{G_2^2}{mr^2}
 =\frac{e^{-\eps_0}}{160}.
\]

 Fix $t\in[T]$ and $J\subseteq[s]$ with $|J|=\ell$.
 Writing $\mathsf{tr}$ for the private transcript, consider
the function
 \[
 h(\mathsf{tr},z_J)=
 \begin{cases}
 \displaystyle\prod_{i\in J}
 \1\{\|\nabla\widehat F_{z_i}(w_t)-\nabla F(w_t)\|>r\},
 & \text{if round $t$ is reached},\\[4pt]
 0, & \text{otherwise}.
 \end{cases}
\]
 Here $w_t$ is the query specified by the transcript.
 Let $Z'_J=(Z'_i)_{i\in J}$ consist of $\ell$ independent user
blocks, each distributed as $P^m$, and independent of $Z$
and the algorithm's randomness. 
 Conditional on $\mathsf{tr}$, these fresh blocks remain
independent and $w_t$ is fixed whenever round $t$ is reached.
 Thus,
 \[
 \E[h(\mathsf{tr},Z'_J)\mid\mathsf{tr}]
 \le\left(\frac{e^{-\eps_0}}{160}\right)^\ell.
\]
 By the tower property,
 \begin{align*}
 e^{\ell\eps_0}\E h(\mathsf{tr},Z'_J)
 &=e^{\ell\eps_0}
   \E\!\left[\E[h(\mathsf{tr},Z'_J)\mid\mathsf{tr}]\right]\\
 &\le e^{\ell\eps_0}
      \left(\frac{e^{-\eps_0}}{160}\right)^\ell
 =160^{-\ell}.
\end{align*}
 Then applying Lemma~\ref{lem:bounded-stability} to the transcript,
with $M=1$ and this set $J$ of $\ell$ users, gives
 \begin{align*}
 \E\1\{J\subseteq\mathcal B_t\}
 &=\E h(\mathsf{tr},Z_J)\\
 &\le e^{\ell\eps_0}\E h(\mathsf{tr},Z'_J)
      +\delta_0\sum_{a=0}^{\ell-1}e^{a\eps_0}\\
 &\le160^{-\ell}
      +\delta_0\sum_{a=0}^{\ell-1}e^{a\eps_0}.
\end{align*}

 For each realization of $\mathcal B_t$,
 \[
 \binom{|\mathcal B_t|}{\ell}
 =\sum_{\substack{J\subseteq[s]\\|J|=\ell}}
   \1\{J\subseteq\mathcal B_t\}.
\]
 Indeed, the sum counts precisely the $\ell$-element subsets
of $\mathcal B_t$.  Taking expectations and applying the
preceding bound to each summand gives
 \begin{align*}
 \E\binom{|\mathcal B_t|}{\ell}
 &=\sum_{\substack{J\subseteq[s]\\|J|=\ell}}
   \E\1\{J\subseteq\mathcal B_t\}\\
 &\le\binom{s}{\ell}
 \left(160^{-\ell}
       +\delta_0\sum_{a=0}^{\ell-1}e^{a\eps_0}\right).
\end{align*}

 If $|\mathcal B_t|\ge\lceil s/20\rceil$, then
 \[
 \binom{|\mathcal B_t|}{\ell}
 \ge\binom{\lceil s/20\rceil}{\ell}.
\]
 By Markov's inequality and the preceding expectation bound,
 \begin{align*}
 \Pr\!\left(|\mathcal B_t|\ge\lceil s/20\rceil\right)
 &\le
 \frac{\E\binom{|\mathcal B_t|}{\ell}}
      {\binom{\lceil s/20\rceil}{\ell}}\\
 &\le40^\ell\left(
 160^{-\ell}+\delta_0\sum_{a=0}^{\ell-1}e^{a\eps_0}
 \right)\\
 &\le4^{-\ell}+\ell(40e)^\ell\delta_0\\
 &\le\frac{\beta_c}{4T}+\frac{\beta_c}{4T}
 =\frac{\beta_c}{2T}.
\end{align*}
 The second inequality uses
 \[
 \frac{\binom{s}{\ell}}{\binom{\lceil s/20\rceil}{\ell}}
 =\prod_{a=0}^{\ell-1}\frac{s-a}{\lceil s/20\rceil-a}
 \le\left(\frac{s}{s/20-\ell+1}\right)^\ell
 \le40^\ell,
\]
 since $s\ge40\ell$.  The third uses $\eps_0\le1$, and the
last uses the choice of $\ell$ and~\eqref{eq:center-delta}. 

Define  \[
\mathcal G_t:=[s]\setminus\mathcal B_t.\]
 A union bound over $t\in[T]$ shows that, with probability
at least $1-\beta_c/2$, every reached round satisfies
 \[
 |\mathcal B_t|<\lceil s/20\rceil,
 \qquad
 |\mathcal G_t|>\frac{19s}{20}.
\]

 We next control the noise in the concentration tests.
 For a reached round, define
 \[
 q_t:=\frac1s\sum_{i,j=1}^s
 \1\!\left\{
 \|\nabla\widehat F_{Z_i}(w_t)
   -\nabla\widehat F_{Z_j}(w_t)\|\le\tau
 \right\}.
\]
 The threshold noise has scale $8/\eps_0$, and each score
noise has scale $16/\eps_0$.
 The Laplace tail bound and a union bound imply that, except
with probability at most
 \[
 (T+1)\exp(-\eps_0s/640)\le\beta_c/2,
\]
 we have
 \[
 |\widehat\theta-4s/5|\le s/40,
 \qquad
 |\widehat q_t-q_t|\le s/40
 \quad\text{at every reached round}.
\]
 The last probability bound follows from the sample-size
assumption.

 By a union bound, both preceding events hold with probability
at least $1-\beta_c$.  Consider any reached round on their high-probability
intersection.  For $i,j\in\mathcal G_t$,
 \begin{align*}
 \|\nabla\widehat F_{Z_i}(w_t)
   -\nabla\widehat F_{Z_j}(w_t)\|
 &\le\|\nabla\widehat F_{Z_i}(w_t)-\nabla F(w_t)\|\\
 &\quad+\|\nabla\widehat F_{Z_j}(w_t)-\nabla F(w_t)\|\\
 &\le2r=\tau.
\end{align*}
 It follows that
 \begin{align*}
 q_t
 &\ge\frac{|\mathcal G_t|^2}{s}
 >\left(\frac{19}{20}\right)^2s>\frac{9s}{10},\\
 \widehat q_t-\widehat\theta
 &\ge q_t-\frac{4s}{5}-\frac{s}{20}
 >\frac{s}{20}>0.
\end{align*}
 Thus the concentration test accepts.

 For every $i\in\mathcal G_t$, the filter count satisfies
 \[
 h_{t,i}
 =\sum_{j=1}^s
 \1\!\left\{
 \|\nabla\widehat F_{Z_i}(w_t)
   -\nabla\widehat F_{Z_j}(w_t)\|\le2\tau
 \right\}
 \ge|\mathcal G_t|>\frac{2s}{3},
\]
 so $p_{t,i}=1$.
 Hence $\mathcal G_t\subseteq\cS_t$, and in particular
$\cS_t\ne\varnothing$.

 Conversely, every selected user $i\in\cS_t$ has
$p_{t,i}>0$, and therefore $h_{t,i}>s/2$.
 Since fewer than $s/20$ users lie outside $\mathcal G_t$,
at least one of these neighbors belongs to $\mathcal G_t$.
 For such a user $j$,
 \begin{align*}
 \|\nabla\widehat F_{Z_i}(w_t)-\nabla F(w_t)\|
 &\le\|\nabla\widehat F_{Z_i}(w_t)
       -\nabla\widehat F_{Z_j}(w_t)\|\\
 &\quad+\|\nabla\widehat F_{Z_j}(w_t)-\nabla F(w_t)\|\\
 &\le2\tau+r=5r.
\end{align*}
 Averaging over the selected users gives
 \[
 \|Q_t-\nabla F(w_t)\|
 \le\frac1{|\cS_t|}\sum_{i\in\cS_t}
       \|\nabla\widehat F_{Z_i}(w_t)-\nabla F(w_t)\|
 \le5r\le R/2.
\]

 Thus, with probability at least $1-\beta_c$, neither stopping condition can occur at any reached
round.  The algorithm completes all $T$ iterations, and the
center bound holds at each one.
\end{proof}

 Combining the privacy guarantee of Proposition~\ref{prop:private-erm-solver}, the center guarantee of
Lemma~\ref{lem:adaptive-good-centers}, the adaptive tail bound
of Lemma~\ref{lem:approx-tail-transfer}, and the clipping bound
of Lemma~\ref{lem:bias-from-tail} gives the following bound on
the average squared clipping bias.  We will use it in the
utility proof of Proposition~\ref{prop:private-erm-solver}.

 \begin{lemma}[{Adaptive empirical clipping bias}]
\label[lemma]{prop:fused-adaptive-bias}
 Let $Z=(Z_1,\ldots,Z_s)\sim(P^m)^s$, with incoming center
$w_0\in\mathcal W$ independent of $Z$.
 Run Algorithm~\ref{alg:fused} with $0<\eps\le1$,
iteration budget $T\ge1$,
$\delta_0$ from~\eqref{eq:internal-delta}, and
 \[
 r=\frac{\sqrt{160}\,e^{\eps/2}G_2}{\sqrt m},
 \qquad \tau=2r,
 \qquad L_0=\log(16T/\delta_0).
\]
 For each performed update, define the bias
 \[
 b_t:=\frac1s\sum_{i=1}^s
 \Pi_{B(Q_t,R)}\nabla\widehat F_{Z_i}(w_t)
 -\nabla\widehat F_Z(w_t).
\]
 If $s\ge C L_0/\eps$ and $R\ge10r$, then, for each fixed
$j\in\{2,k\}$, with probability at least $0.97$ over $Z$
and the algorithm's randomness, the algorithm completes all
$T$ iterations and
 \begin{equation}
 \frac1T\sum_{t=1}^T\|b_t\|^2
 \le C_k\left[\psi_j(R)^2+\frac{G_2^2}{ms}\right],
 \label{eq:adaptive-bias-final}
\end{equation}
 where
 \[
 \psi_2(R)=\frac{2G_2^2}{mR},
 \qquad
 \psi_k(R)=C_k\left[
 \left(R+\frac{G_2^2}{mR}\right)e^{-c_kmR^2/G_2^2}
 +\frac{G_k^k}{m^{k-1}R^{k-1}}
 \right]\quad(k>2).
\]
 When $k=2$, only $\psi_2$ is used.
\end{lemma}

\begin{proof}
 Condition on $w_0$, and let $N\le T$ be the number of
performed updates.
 The preceding privacy proof establishes
$(\eps,\delta_0)$-user-level privacy of the noisy-gradient
transcript and stopping decision under $s\ge C L_0/\eps$.
 By post-processing, the queried points and stopping decision
have the same privacy guarantee.

 Set $\gamma=\nu=1/100$.
 Let $\mathsf{tr}$ denote the private transcript, which determines
$N$ and the queries $w_1,\ldots,w_N$ used in the performed updates.
 For each $t\in[T]$ and user block $V$, define
 \[
 U_t(\mathsf{tr},V):=
 \begin{cases}
 \displaystyle
 \min\left\{
 \left(\|\nabla\widehat F_V(w_t)-\nabla F(w_t)\|-R/2\right)_+,\,
 G_2(1+\sqrt{s/\gamma})
 \right\},
 & t\le N,\\[4pt]
 0, & t>N.
 \end{cases}
\]
 For each fixed $t$, the two functions to which we apply
Lemma~\ref{lem:bounded-stability} are
 \[
 h_1(\mathsf{tr},V)=U_t(\mathsf{tr},V)^2,
 \qquad
 h_2(\mathsf{tr},V,V')
 =U_t(\mathsf{tr},V)U_t(\mathsf{tr},V').
\]
 Both take values in $[0,M]$ with $M=4G_2^2s/\gamma$.

 Let $V,V'\sim P^m$ be independent of each other, of $Z$,
and of the algorithm's randomness.
 Conditional on $\mathsf{tr}$, the fixed-query moment bounds give
 \[
 \E[h_1(\mathsf{tr},V)\mid\mathsf{tr}]
 \le\frac{G_2^2}{m},
 \qquad
 \E[h_2(\mathsf{tr},V,V')\mid\mathsf{tr}]
 \le\psi_j(R)^2.
\]
 For $t\le N$, these follow from the fixed-query bounds and
independence of $V,V'$; for $t>N$, both left-hand sides are zero.
 Taking expectations gives the same unconditional bounds.

 Apply Lemma~\ref{lem:bounded-stability} with $\ell=1$ to
$h_1$ and with $\ell=2$ to $h_2$.
 Expanding the square and averaging over $t$ as in the proof
of Lemma~\ref{lem:approx-tail-transfer} gives
 \begin{align*}
 \E\left[
 \frac1T\sum_{t=1}^T
 \left(\frac1s\sum_{i=1}^s U_t(\mathsf{tr},Z_i)\right)^2
 \right]
 &\le e^{2\eps}\psi_j(R)^2
      +\frac{e^\eps G_2^2}{ms}
      +(1+e^\eps)\delta_0M\\
 &\le C\left[\psi_j(R)^2+\frac{G_2^2}{ms}\right].
\end{align*}
 Thus Markov's inequality bounds this average by
$C\nu^{-1}[\psi_j(R)^2+G_2^2/(ms)]$ with probability
at least $1-\nu$.

 The uniform gradient bound in that same proof gives, with
probability at least $1-\gamma$,
 \[
 U_t(\mathsf{tr},Z_i)
 =\left(
 \|\nabla\widehat F_{Z_i}(w_t)-\nabla F(w_t)\|-R/2
 \right)_+
 \qquad
 \text{for all }t\le N,\ i\in[s].
\]
 Since $U_t(\mathsf{tr},Z_i)=0$ for $t>N$, intersecting
these two events yields, with probability at least
$1-\gamma-\nu=0.98$,
 \[
 \frac1T\sum_{t=1}^{N}
 \left[\frac1s\sum_{i=1}^s
 \left(
 \|\nabla\widehat F_{Z_i}(w_t)-\nabla F(w_t)\|-R/2
 \right)_+\right]^2
 \le C_k\left[\psi_j(R)^2+\frac{G_2^2}{ms}\right].
\]

 The choice of $\delta_0$ and the bound $s\ge C L_0/\eps$
ensure the hypotheses of Lemma~\ref{lem:adaptive-good-centers}
with $\beta_c=1/100$.
 That lemma gives, with probability at least $0.99$,
 \[
 N=T,
 \qquad
 \|Q_t-\nabla F(w_t)\|\le R/2
 \quad\text{for every }t\in[T].
\]
 By a union bound, both conclusions hold with probability at
least $1-0.02-0.01=0.97$.
 On their intersection, Lemma~\ref{lem:bias-from-tail} gives
 \begin{align*}
 \frac1T\sum_{t=1}^T\|b_t\|^2
 &\le
 \frac1T\sum_{t=1}^T
 \left[\frac1s\sum_{i=1}^s
 \left(
 \|\nabla\widehat F_{Z_i}(w_t)-\nabla F(w_t)\|-R/2
 \right)_+\right]^2\\
 &\le C_k\left[\psi_j(R)^2+\frac{G_2^2}{ms}\right].
\end{align*}
 The bounds are uniform in $w_0$, so averaging over its
distribution proves the claim.
\end{proof}

 \begin{proof}[Completion of the proof of Proposition~\ref{prop:private-erm-solver}]
 \textbf{Utility.}
Fix the incoming center $w_0$.
 We prove~\eqref{eq:erm-distance-target} by combining
Lemma~\ref{prop:fused-adaptive-bias} with the one-step
SGD inequality~\eqref{eq:one-step-biased-gd}.
 We retain the realized biases in that inequality and control
the Gaussian inner-product term by an exponential-moment
bound.  All probabilities and expectations below are over
$Z\sim(P^m)^s$ and the algorithm's randomness, conditional
on the fixed $w_0$.

 At each performed update, the Gaussian noise satisfies
 \begin{equation}
 \Sigma^2:=\E[\|v_t\|^2\mid\mathcal F_t]
 =d\sigma^2
 =C^2\frac{d(R+\tau L_0)^2T L_0^2}{s^2\eps^2},
 \label{eq:fused-noise-general}
\end{equation}
 where $\mathcal F_t$ contains $Z$ and all randomness through
the filter in round $t$, before drawing $v_t$.
 In particular, $w_t$, $Q_t$, and $b_t$ are
$\mathcal F_t$-measurable, and
$\E[v_t\mid\mathcal F_t]=0$.

 The lower bound on $R$ in~\eqref{eq:R-sco} and
$L_0=O_k(\Lambda)$ give
 \begin{equation}
 R\ge C\tau L_0.
 \label{eq:R-coupling-floor}
\end{equation}
 Recalling $a=C_k\Lambda\sqrt d/(s\eps)$, we obtain
 \begin{equation}
 \Sigma^2\le C'\frac{dR^2T L_0^2}{s^2\eps^2},
 \qquad
 \frac{\Sigma}{\sqrt T}\le aR.
 \label{eq:fused-noise}
\end{equation}

 The parameter choices ensure $s\ge C L_0/\eps$ and
$R\ge10r$, so Lemma~\ref{prop:fused-adaptive-bias} applies.
 Let $N\le T$ be the number of performed updates and define
 \[
 B_0^2:=C_k\left[\psi_j(R)^2+\frac{G_2^2}{ms}\right].
\]
 The lemma gives
 \[
 \Pr\!\left(
 N=T,\quad
 \sum_{t=1}^{N}\|b_t\|^2\le TB_0^2
 \right)\ge0.97.
\]

 Define the weighted empirical objective error and Gaussian
inner-product sum by
 \[
 \begin{aligned}
 H_N&:=\sum_{t=1}^{N}t
 \left[
 \widehat F_{\lambda,Z}(w_t)
 -\widehat F_{\lambda,Z}(\widehat w_{\lambda,Z})
 \right],\\
 M_N&:=-\sum_{t=1}^{N}
 t\langle v_t,w_t-\widehat w_{\lambda,Z}\rangle.
 \end{aligned}
\]
 Multiply~\eqref{eq:one-step-biased-gd} by $t$ and sum over
$t=1,\ldots,N$.
 The squared-distance terms telescope, with zero initial
coefficient and a nonpositive final term.
 Using $t\eta_t\le4/\lambda$ and
$g_t=\nabla\widehat F_Z(w_t)+b_t+v_t$ gives
 \[
 H_N\le
 \frac{CT}{\lambda}\sum_{t=1}^{N}\|b_t\|^2
 +\frac C\lambda\sum_{t=1}^{N}
 \left[
 \|\nabla\widehat F_Z(w_t)\|^2+\|v_t\|^2+\lambda^2D^2
 \right]
 +M_N.
\]

 To bound the sum of squared gradient and noise norms, observe
that
 \[
 \E_Z\left[
 \sup_{w\in\mathcal W}\|\nabla\widehat F_Z(w)\|^2
 \right]
 \le
 \E_Z\left[
 \left(\frac1{sm}\sum_{i=1}^s\sum_{a=1}^m
 \sup_{w\in\mathcal W}\|\nabla f(w,z_{i,a})\|\right)^2
 \right]
 \le G_2^2.
\]
 The final inequality follows from Jensen's inequality and the
second-moment assumption.
 Also, the decision to perform update $t$ precedes the draw
of $v_t$.  Hence
 \[
 \E\sum_{t=1}^{N}\|v_t\|^2
 =\Sigma^2\E N\le T\Sigma^2.
\]
 Together these imply
 \[
 \E\sum_{t=1}^{N}
 \left[\|\nabla\widehat F_Z(w_t)\|^2+\|v_t\|^2\right]
 \le T(G_2^2+\Sigma^2).
\]
 By Markov's inequality, with probability at least $0.95$,
 \[
 \sum_{t=1}^{N}
 \left[\|\nabla\widehat F_Z(w_t)\|^2+\|v_t\|^2\right]
 \le20T(G_2^2+\Sigma^2).
\]

 It remains to control $M_N$.
 Conditional on $\mathcal F_t$, the Gaussian moment-generating
function gives, for every $\theta>0$,
 \[
 \E\!\left[
 \exp\!\left(
 -\theta t\langle v_t,w_t-\widehat w_{\lambda,Z}\rangle
 -\frac{\theta^2\sigma^2t^2}{2}
  \|w_t-\widehat w_{\lambda,Z}\|^2
 \right)
 \,\middle|\,\mathcal F_t
 \right]=1.
\]
 Iterating this identity over the performed updates yields
 \[
 \E\exp\!\left(
 \theta M_N-\frac{\theta^2\sigma^2}{2}
 \sum_{t=1}^{N}t^2
 \|w_t-\widehat w_{\lambda,Z}\|^2
 \right)\le1.
\]
 This remains valid for the stopped sum because each decision
to perform an update is made before its Gaussian draw, and
there are at most $T$ updates.

 Strong convexity and optimality of
$\widehat w_{\lambda,Z}$ imply
 \begin{align*}
 \sum_{t=1}^{N}t^2
 \|w_t-\widehat w_{\lambda,Z}\|^2
 &\le\frac2\lambda\sum_{t=1}^{N}t^2
 \left[
 \widehat F_{\lambda,Z}(w_t)
 -\widehat F_{\lambda,Z}(\widehat w_{\lambda,Z})
 \right]\\
 &\le\frac{2T}{\lambda}H_N.
\end{align*}
 Taking $\theta=\lambda/(2\sigma^2T)$ therefore gives
 \[
 \E\exp\!\left(\theta[M_N-H_N/2]\right)\le1.
\]
 Another application of Markov's inequality gives, with
probability at least $0.95$,
 \[
 M_N\le\frac12H_N+\frac{2\sigma^2T}{\lambda}\log20.
\]

 By a union bound, the preceding two inequalities and
 \[
 N=T,\qquad
 \sum_{t=1}^{N}\|b_t\|^2\le TB_0^2
\]
 hold simultaneously with probability at least
$1-0.05-0.05-0.03=0.87$.
 Substituting these bounds into the inequality for $H_N$,
moving $H_T/2$ to the left, and using
$\sigma^2\le\Sigma^2$ gives
 \[
 H_T\le
 \frac{CT^2B_0^2}{\lambda}
 +\frac{CT}{\lambda}
 \left(G_2^2+\Sigma^2+\lambda^2D^2\right).
\]
 On this event the algorithm completes all iterations, so
 \[
 w_{\rm out}=\frac{2}{T(T+1)}\sum_{t=1}^Ttw_t.
\]
 Convexity and strong convexity now give
 \begin{align*}
 \|w_{\rm out}-\widehat w_{\lambda,Z}\|^2
 &\le\frac2\lambda
 \left[
 \widehat F_{\lambda,Z}(w_{\rm out})
 -\widehat F_{\lambda,Z}(\widehat w_{\lambda,Z})
 \right]\\
 &\le\frac{4H_T}{\lambda T(T+1)}\\
 &\le\frac C{\lambda^2}
 \left[
 B_0^2+\frac{G_2^2+\Sigma^2+\lambda^2D^2}{T}
 \right].
\end{align*}
 Taking square roots and applying~\eqref{eq:fused-noise},
we obtain, with probability at least $0.87$,
 \[
 \|w_{\rm out}-\widehat w_{\lambda,Z}\|
 \le\frac C\lambda
 \left[
 B_0+aR+\frac{G_2+\lambda D}{\sqrt T}
 \right].
\]

 We finish by substituting the choices of $T$ and $R$.
 The iteration budget gives
 \[
 \frac{G_2+\lambda D}{\sqrt T}
 \lesssim_k\mathcal E_{s,m}(G_2,G_k).
\]
 For $j=k>2$, the lower bound
$R\ge C_k\Lambda G_2/\sqrt m$ ensures
 \[
 C_k\left(R+\frac{G_2^2}{mR}\right)
 e^{-c_kmR^2/G_2^2}
 \le\frac{G_2}{\sqrt{ms}}.
\]
 Using this bound when $j=k>2$, and
$\psi_2(R)=2G_2^2/(mR)$ when $j=2$, gives
 \[
 B_0\le C_k\left[
 \frac{G_2}{\sqrt{ms}}
 +\frac{G_j^j}{m^{j-1}R^{j-1}}
 \right].
\]
 The choice of $R$ in~\eqref{eq:R-sco} implies
 \[
 \frac{G_j^j}{m^{j-1}R^{j-1}}
 \le G_j(a/m)^{1-1/j},
 \qquad
 aR\le\frac{C_k\Lambda aG_2}{\sqrt m}
       +G_j(a/m)^{1-1/j}.
\]
 Since $j$ minimizes $G_j(a/m)^{1-1/j}$ over $\{2,k\}$,
 \begin{align*}
 B_0+aR
 &\le C_k\left[
 \frac{G_2}{\sqrt{ms}}
 +\frac{G_2\sqrt d\,\Lambda^2}{s\eps\sqrt m}
 +\min\left\{
 G_2(a/m)^{1/2},\,
 G_k(a/m)^{1-1/k}
 \right\}\right]\\
 &\lesssim_k\Lambda^2\mathcal E_{s,m}(G_2,G_k).
\end{align*}
 Combining these estimates proves
 \[
 \Pr\!\left(
 \|w_{\rm out}-\widehat w_{\lambda,Z}\|
 \le
 \frac{C_k\Lambda^4\mathcal E_{s,m}(G_2,G_k)}{\lambda}
 \right)\ge0.87\ge0.8.
\]
 This is~\eqref{eq:erm-distance-target}.
 The bounds hold uniformly in the incoming center $w_0$,
so they also hold when $w_0$ is random and independent of $Z$.
\end{proof}

%% file: appendix/04d_sco_upper.tex
 \subsubsection{The final SCO upper bound}
 The regularized ERM guarantee verifies the remaining hypothesis
of the population localization theorem.
\begin{theorem}[Restatement of Theorem~\ref{thm:main-sco}]
\label[theorem]{thm:sco}
Assume~\eqref{eq:sco-moments} and $n\ge C_k\Lambda^6/\eps$.
Algorithm~\ref{alg:poploc}, instantiated with Algorithm~\ref{alg:fused},
is $(\eps,\delta)$-user-level DP and, with probability at least $1-\beta$,
satisfies
\begin{equation}
 F(\widehat w)-F^*\le D\min\bigl\{G_2,\ C_k\readingstart{logarithmic-factor}\revisionstart\Lambda^6\revisionend{}\readingend\err(G_2,G_k)\bigr\}.
 \label{eq:sco-final}
\end{equation}
\end{theorem}

\begin{proof}[Proof of Theorem~\ref{thm:sco}]
 If $\Delta\ge G_2$, Algorithm~\ref{alg:poploc} returns $w_0$.
 Since $F$ is $G_2$-Lipschitz and~\eqref{eq:localization-block-rate}
bounds $\Delta$, we have
 \[
 F(w_0)-F^*\le DG_2\le D\Delta\le C_kD\Lambda^6\err(G_2,G_k).
\]
 Otherwise, it suffices to check the two parameter restrictions in
Proposition~\ref{prop:private-erm-solver}.  First,
$\lambda_iD=32^{i-1}\Delta\le32G_2\sqrt{nm}$ for every phase.
 Second, $IJ=O(\Lambda^2)$ and $n\ge C_k\Lambda^6/\eps$ imply
$s=\lfloor n/(IJ)\rfloor\ge C'_k\Lambda/\eps$.
 The proposition therefore supplies a private solver satisfying
\eqref{eq:main-erm-target} for every phase and every incoming center.
 Applying Theorem~\ref{thm:poploc-meta-detail} gives~\eqref{eq:sco-final}
and user-level privacy.
\end{proof}

\readingstart{moment-bounds}\revisionstart \begin{remark}[{Public moment bounds}]
\label{rem:moment-inputs}
The parameters $G_2$ and $G_k$ are public upper bounds.
Overestimating them preserves the utility guarantee with
the supplied bounds substituted for $G_2$ and $G_k$.
Underestimating them can cause the concentration test to
reject or increase clipping bias, invalidating the stated
accuracy guarantee; privacy remains valid for fixed public
inputs. A smaller valid moment order yields its corresponding
rate. Private moment estimation is possible under stronger
assumptions~\citep{kothari2022private}; obtaining parameter-free
algorithms with optimal rates under our assumptions remains
an important direction for future work.
\end{remark}\revisionend{}\readingend

%% file: appendix/04e_sco_lower.tex
\subsection{Lower bound}
\label{app:sco-lower}

\begin{theorem}[Restatement of Theorem~\ref{thm:main-sco-lower}]
\label[theorem]{thm:norm-sco-lower}
 Assume~\eqref{eq:lower-delta} and let $\mathcal W$ be a Euclidean
ball of diameter $D$.  For every user-level $(\eps,\delta)$-DP algorithm
$\mathcal A$ with output in $\mathcal W$, there is a distribution $P$
of linear losses satisfying~\eqref{eq:momentbound} such that
 \begin{equation}
 \E_{\mathsf Z\sim(P^{\otimes m})^{\otimes n},\,\mathcal A}
 \bigl[F(\mathcal A(\mathsf Z))-F^*\bigr]
 \ge c_k D\min\{G_2,\err(G_2,G_k)\}.
 \label{eq:norm-sco-lower}
\end{equation}
\end{theorem}

 We use the Bernoulli test, Gaussian fingerprinting argument, and
sparse-mixture reduction from the mean-estimation lower bound.  Two changes are needed for SCO: the moment bounds apply to the
gradients themselves, and the error is measured by excess population
risk.  To see what the latter requires, center the ball at zero and take
$f(w,z)=-\langle w,z\rangle$.  If $z\sim P$ has mean $\mu_P$, then
 \begin{equation}
 L_P(w):=F_P(w)-F_P^*
 =\frac D2\|\mu_P\|_2-\langle w,\mu_P\rangle,
 \qquad 0\le L_P(w)\le D\|\mu_P\|_2.
 \label{eq:sco-linear-risk}
\end{equation}
 Thus the Gaussian mean-estimation lower bound alone does not suffice:
the minimizer $(D/2)\mu_P/\|\mu_P\|_2$ for $\mu_P\ne0$ depends
only on the direction of $\mu_P$.  We first bound how much alignment privacy allows for Gaussian
data.  The resulting family will also supply the nonzero observations
in the heavy-tail construction.

 \begin{lemma}[{Gaussian linear losses}]
\label[lemma]{lem:gaussian-sco-direct}
 Fix $k\ge2$, $0<G_2\le G_k$, $0<\eps\le1$, and $0<\delta<1/2$
satisfying~\eqref{eq:gaussian-delta}.  There is a family $\mathcal Q$
of Gaussian laws satisfying~\eqref{eq:sco-moments} and a number
 \[
 \rho\asymp_k DG_2\min\left\{1,\frac{\sqrt d}{n\eps\sqrt m}\right\}
\]
 such that $L_Q(w)\le\rho$ for every $Q\in\mathcal Q$ and $w\in\mathcal W$.
 For every user-level $(\eps,\delta)$-DP algorithm $\mathcal A$, some
$Q\in\mathcal Q$ satisfies
 \[
 \E_{\mathsf Z\sim(Q^{\otimes m})^{\otimes n},\,\mathcal A}
 [L_Q(\mathcal A(\mathsf Z))]\ge\rho/8.
\]
\end{lemma}

\begin{proof}
 Let $K_k=(\E|g|^k)^{1/k}$ for $g\sim\mathcal N(0,1)$, and set
 \[
 \sigma:=\frac{G_2}{(1+K_k)\sqrt d},\qquad
 a:=c_0\sigma\min\left\{1,\frac{\sqrt d}{n\eps\sqrt m}\right\},
 \qquad \rho:=Da\sqrt d,
\]
 where $c_0\in(0,1)$ is a universal constant to be fixed below.
 Take $Q_\theta=\mathcal N(\theta,\sigma^2I_d)$ with
$\theta\in[-a,a]^d$.  For $j\in\{2,k\}$, the Gaussian moment bound
in the proof of Lemma~\ref{lem:user-gaussian-lower} gives
 \[
 (\E_{z\sim Q_\theta}\|z\|_2^j)^{1/j}
 \le\|\theta\|_2+\sigma K_k\sqrt d\le G_2\le G_j.
\]
 These are the required raw moments since $\nabla f(w,z)=-z$.
 Also, \eqref{eq:sco-linear-risk} gives $L_{Q_\theta}(w)\le\rho$.

 To obtain a lower bound, draw $\theta$ uniformly from $[-a,a]^d$,
then draw the users' samples from $Q_\theta$.  Write
$W=\mathcal A(\mathsf Z)$, let $Y_i$ be user $i$'s sample mean, and set
 \[
 g_i:=\frac{\sqrt m}{\sigma}(Y_i-\theta),\qquad
 V_i:=\sum_{j=1}^d(a^2-\theta_j^2)W_jg_{i,j}.
\]
 The Gaussian score for user $i$ is $(\sqrt m/\sigma)g_i$.
 Thus the fingerprinting identity used in~\eqref{eq:gaussian-accuracy-correlation},
applied to the bounded output $W$, gives
 \[
 2\E\langle W,\theta\rangle
 =\frac{\sqrt m}{\sigma}\sum_{i=1}^n\E V_i.
\]
 Here the expectation includes the prior, the samples, and the algorithm.
 To apply the privacy bound from the same proof, put $H=a^2D/2$.
 Then $|V_i|\le H\|g_i\|_2$.  If we replace user $i$ in the input
to $W$ but retain the original $g_i$, the resulting statistic is
centered Gaussian with standard deviation at most $H$ conditional
on $\theta$ and the replacement output.
 The truncation calculation in~\eqref{eq:gaussian-privacy-correlation}
therefore gives
 \[
 \E V_i\le C H\bigl[\eps+\delta(\sqrt d+\sqrt{\log(e/\delta)})\bigr]
 \le C' H\eps.
\]
 Since $n\eps\sqrt m\,a/\sigma\le c_0\sqrt d$, the identity now yields
$\E\langle W,\theta\rangle\le C''c_0Da\sqrt d\le\rho/8$
for sufficiently small $c_0$.  In contrast,
$\E\|\theta\|_2\ge\E\|\theta\|_1/\sqrt d=a\sqrt d/2$.
 Substituting both bounds into~\eqref{eq:sco-linear-risk} gives
 \[
 \E L_{Q_\theta}(W)\ge\frac{Da\sqrt d}{4}-\frac\rho8=\frac\rho8.
\]
 Hence the same lower bound holds for at least one fixed $\theta$.
\end{proof}

 We next use the sampling reduction from Lemma~\ref{lem:sparse-norm-lower}.
 For linear losses the reduction is particularly simple: multiplying
the mean by $pb$ multiplies every feasible point's excess risk by $pb$.

 \begin{lemma}[{Sparse mixtures of linear losses}]
\label[lemma]{lem:sparse-sco-lower}
 Under the privacy and moment parameter assumptions of
Lemma~\ref{lem:gaussian-sco-direct}, set
$q_N:=\min\{1,\sqrt d/(N\eps)\}$.  For every item-level
$(\eps,\delta)$-DP algorithm $\mathcal A$ on $N$ observations,
there is a distribution $P$ of linear losses satisfying
 \begin{equation}
 \E_{z\sim P}\left[\sup_{w\in\mathcal W}\|\nabla f(w,z)\|_2^j\right]
 \le G_j^j,\qquad j\in\{2,k\},
 \label{eq:sparse-sco-moments}
\end{equation}
 such that
 \begin{equation}
 \E_{Z\sim P^{\otimes N},\,\mathcal A}[L_P(\mathcal A(Z))]
 \ge c_k D\min\{G_2q_N^{1/2},G_kq_N^{1-1/k}\}.
 \label{eq:sparse-sco-risk}
\end{equation}
\end{lemma}

\begin{proof}
 Take $s:=\min\{N,\lfloor\sqrt d/\eps\rfloor\}$ and apply
Lemma~\ref{lem:gaussian-sco-direct} with $n=s$, $m=1$, and
$G_2=G_k=1$.  Since $s\le\sqrt d/\eps$, its risk scale is
$\rho\asymp_k D$.  For each $Q$ in that family let $P_Q$ be the law
of $bUX$, where $X\sim Q$ and $U\sim\operatorname{Bernoulli}(p)$
are independent and
 \[
 p:=q_N/200,\qquad b:=\min\{G_2p^{-1/2},G_kp^{-1/k}\}.
\]
 For $j\in\{2,k\}$ we have
$\E\|bUX\|_2^j=pb^j\E\|X\|_2^j\le G_j^j$,
which proves~\eqref{eq:sparse-sco-moments}.
 Since $\mu_{P_Q}=pb\mu_Q$, equation~\eqref{eq:sco-linear-risk} gives
$L_{P_Q}(w)=pbL_Q(w)$ for every $w$.

 Use the construction in the proof of Lemma~\ref{lem:sparse-norm-lower}
with $s$ samples from $Q$ and scale $b$.  The resulting algorithm
$\mathcal B$ returns the output of $\mathcal A$ unchanged if the
number $K$ of nonzero positions is at most $s$, and returns $0$
otherwise.  Each input sample affects at most one position for fixed
indicators, so $\mathcal B$ is item-level $(\eps,\delta)$-DP.
 In the coupling used for~\eqref{eq:sparse-mixture-transfer}, the
input to $\mathcal A$ agrees on $K\le s$ with
$Z\sim P_Q^{\otimes N}$.  On $K>s$ the excess risk of $\mathcal B$
is at most $\rho$.  Splitting the expectation over these two events gives,
for $X\sim Q^{\otimes s}$,
 \[
 \E L_Q(\mathcal B(X))
 \le \frac{\E L_{P_Q}(\mathcal A(Z))}{pb}+\rho\Pr(K>s),
 \qquad \Pr(K>s)\le\frac{Np}{s}\le\frac1{100}.
\]
 The last inequality uses $s\ge\min\{N,\sqrt d/\eps\}/2$.
 For the $Q$ supplied by Lemma~\ref{lem:gaussian-sco-direct}, the
left-hand side is at least $\rho/8$. Thus
 \[
 \E L_{P_Q}(\mathcal A(Z))\ge
 pb\rho\left(\frac18-\frac1{100}\right).
\]
 Substituting $pb=\min\{G_2\sqrt p,G_kp^{1-1/k}\}$ and $p=q_N/200$
proves~\eqref{eq:sparse-sco-risk}.
\end{proof}

\begin{proof}[Proof of Theorem~\ref{thm:norm-sco-lower}]
 Write $N=nm$ and let $T_1,T_2,T_3$ be the sampling, Gaussian, and
heavy-tail terms of~\eqref{eq:main-rate}, evaluated at $G_2,G_k$.
 Thus $\err(G_2,G_k)=T_1+T_2+T_3$.

 For $T_1$, use the two Bernoulli laws from the proof of
Lemma~\ref{lem:norm-nonprivate-lower}, scaled to
$z\in\{-G_2e_1,G_2e_1\}$.  Their means are $\eta h G_2e_1$, where
$\eta\in\{-1,1\}$ and $h=1/\sqrt{24N}$.  Both laws satisfy the raw
moment bounds since $\|z\|_2=G_2$.  Test $\eta$ by the sign of the
first coordinate of $W=\mathcal A(\mathsf Z)$, breaking ties arbitrarily.
 By the same two-point bound, this test errs with probability at least
$3/8$ for one law $P$.  On a testing error
$\langle W,\mu_P\rangle\le0$, so~\eqref{eq:sco-linear-risk} gives
$\E L_P(W)\ge(3/8)(DhG_2/2)\gtrsim DT_1$.

 Lemma~\ref{lem:gaussian-sco-direct} gives $c_kD\min\{G_2,T_2\}$.
 Its privacy condition follows from~\eqref{eq:lower-delta} as in
the proof of Lemma~\ref{lem:user-gaussian-lower}, including the case
$\delta=0$ by taking a positive $\delta'$ in that proof.
 For $T_3$, group the $N$ items into $n$ users as in
Lemma~\ref{lem:blocking-lower-bound}.  This preserves the output law
and makes the algorithm item-level $(\eps,\delta)$-DP, so
Lemma~\ref{lem:sparse-sco-lower} applies (with the same $\delta'$ if
$\delta=0$).  Its bound is $c_kD\min\{G_2,T_3\}$: if
$\sqrt d/(N\eps)\le1$, then $q_N=\sqrt d/(N\eps)$ and $T_3\le G_2$;
otherwise $q_N=1$ and $T_3\ge G_2$.

 Since $T_1\le G_2$, we have proved a lower bound of
$c_kD\min\{G_2,T_j\}$ for each $j$, possibly under different laws.
 Choose the law for the largest of these three bounds. Then
 \[
 \max_{1\le j\le3}\min\{G_2,T_j\}
 \ge\frac13\min\{G_2,T_1+T_2+T_3\},
\]
 which proves~\eqref{eq:norm-sco-lower}.
\end{proof}

%% file: appendix/05_directional.tex
 \section{Proofs for Section~\ref*{sec:main-directional}: Directional-moment mean estimation}
\label{app:directional}

\begin{theorem}[{Restatement of Theorem~\ref{thm:main-directional}}]
\label[theorem]{thm:directional-final}
Assume $k\ge 2$ and \eqref{eq:intro-directional-model}.
There is a computationally efficient user-level $(\varepsilon,\delta)$-DP estimator with constant-probability $\ell_2$-error at most $\alpha$ using
\begin{equation}
 n=\widetilde O_k\left[\frac d{m\alpha^2}
 +\frac{d\sqrt{\log(1/\delta)}}{\alpha\varepsilon\sqrt m}
 +\frac{d\sqrt{\log(1/\delta)}}{m\varepsilon\alpha^{k/(k-1)}}
 +\frac{\log(1/\delta)}\varepsilon\right].
 \label{eq:directional-final}
\end{equation}
This is tight up to logarithms if $\delta\le c\min\{1/(nm),\varepsilon^2/d,(\varepsilon\min\{1,\alpha\}/d)^{C_k}\}$ for sufficiently large $C_k$.
\end{theorem}

For the upper bound, the proof depends on the target accuracy.  For $k>2$ and $\alpha\le1/\sqrt m$, the estimator of~\citet{agarwal2024person} suffices.  When $\alpha\ge\sqrt{d/m}$, we can apply our norm-moment bound directly.  For the remaining accuracies, including all $0<\alpha<\sqrt{d/m}$ when $k=2$, we refine the estimate on fresh users, choosing the clipping radius from the preceding error bound.  We prove the $k=2$ clipping-bias estimate directly below; no higher-moment tail inequality is needed at this endpoint.
 All numbered citations to~\citet{agarwal2024person} in this appendix
refer to the full version (arXiv:2405.20405v3).
 We denote user $i$'s sample mean by $Y_i=m^{-1}\sum_{a=1}^m z_{i,a}$.

\subsection{Small target error when \texorpdfstring{$k>2$}{k > 2}}

For $k>2$ and $\alpha\le1/\sqrt m$, the desired upper bound
follows directly from \citet[Theorem~4.3]{agarwal2024person}.
Under the directional-moment assumption~\eqref{eq:intro-directional-model},
their algorithm is computationally efficient,
$(\eps,\delta)$-user-level DP, and estimates the mean to
$\ell_2$-error at most $\alpha$ with probability at least $2/3$.
It requires no bound on the location of the mean.
Suppressing logarithmic factors, its user complexity is
\[
n=\widetilde O_k\left(
\frac{d}{m\alpha^2}
+\frac{d}{\alpha\eps\sqrt m}
+\frac{d}{m\eps\alpha^{k/(k-1)}}
+\frac{d}{\eps}
\right).
\]
The first three terms are those in~\eqref{eq:directional-final},
up to logarithmic factors. Since $\alpha\sqrt m\le1$,
\[
\frac{d}{\eps}\le\frac{d}{\alpha\eps\sqrt m},
\]
so the additional $d/\eps$ term is absorbed by the second term.
Thus their algorithm, without modification, achieves the claimed
user complexity in this regime.
 
\readingstart{initialization}
\revisionstart
\subsection{Intermediate target error:
\texorpdfstring{$1/\sqrt m<\alpha<\sqrt{d/m}$}{1/sqrt(m) < alpha < sqrt(d/m)}}

In this regime, we modify the refinement procedure
of~\citet{agarwal2024person} to achieve the user count
in~\eqref{eq:directional-final} without the additional
$d/\eps$ requirement. We first construct a private coarse
estimate, then refine it using fresh users in each round.
For $k=2$, the same construction also covers
$0<\alpha\le1/\sqrt m$.

\subsubsection{Private initialization}

The refinement requires an initial estimate $u_0$ satisfying
$\|u_0-\mu\|\le\sqrt{d/m}/4$.
We construct this estimate using coordinatewise histograms.
The bins cover all of $\mathbb R$, so the construction requires
no prior bound on the location of $\mu$.
\revisionend{}

\revisionstart
\begin{lemma}[Location-free initialization]
\label[lemma]{lem:location-free-histogram}
Assume~\eqref{eq:intro-directional-model}, $k\ge2$,
$0<\eps\le1$, $0<\delta<1/2$, and $0<\eta<1/4$.
There is an $(\eps,\delta)$-user-level DP estimator $u_0$
using $\widetilde O(\sqrt d/\eps)$ users such that
\[
\Pr\left(\|u_0-\mu\|\le\frac14\sqrt{\frac dm}\right)
\ge1-\eta.
\]
The user count hides logarithmic factors in
$d$, $1/\eps$, $1/\delta$, and $1/\eta$,
and does not depend on $\|\mu\|$.
\end{lemma}
\revisionend{}

\begin{proof}
We use the coordinatewise histogram approach
of~\citet[Theorem~4.19]{agarwal2024person}, making the
construction on an unbounded domain explicit.
For each coordinate, concentration ensures that a bin near the
mean contains many observations. We add noise only to occupied
bins and suppress bins with small noisy counts. This suppression
is needed for privacy: a bin occupied by only one user must
rarely be released.

\paragraph{Grouping users.}
Partition the initialization users into $N$ disjoint groups
of $L=1024$ users each. Treat each group as one synthetic user
with $M:=Lm$ samples, and let $V_i$ be its sample average.
For every coordinate $j$, the moment assumption and $k\ge2$ give
\[
\E(z_j-\mu_j)^2
\le\bigl(\E|z_j-\mu_j|^k\bigr)^{2/k}\le1,
\qquad
\operatorname{Var}(V_{i,j})\le\frac1M.
\]
Grouping reduces the coordinate fluctuations enough to obtain
the required constant $1/4$ in the final error bound.

\paragraph{Histogram estimator.}
Set
\[
H:=\log(2/\delta),\qquad
\eps_h:=\frac{\eps}{4\sqrt{dH}},\qquad
\delta_h:=\frac{\delta}{2d},\qquad
h:=\frac8{\sqrt M}.
\]
For coordinate $j$, define
\[
B_\ell:=[\ell h,(\ell+1)h),\qquad
c_{j,\ell}:=\sum_{i=1}^N
\mathbf1\{V_{i,j}\in B_\ell\},
\qquad \ell\in\mathbb Z.
\]
For each occupied bin, independently draw
\[
W_{j,\ell}\sim\operatorname{Lap}(2/\eps_h),
\qquad
\widetilde c_{j,\ell}:=c_{j,\ell}+W_{j,\ell}.
\]
Retain the bin if
\[
\widetilde c_{j,\ell}\ge
\theta:=1+\frac2{\eps_h}\log\frac8{\delta_h}.
\]
Let $\widehat\ell_j$ maximize the retained noisy counts,
breaking ties deterministically, and return
\[
u_{0,j}:=(\widehat\ell_j+1/2)h.
\]
If no bin is retained, set $u_{0,j}:=0$.
Only occupied bins are processed, so each coordinate requires
at most $N$ noise draws.

\paragraph{Privacy of one coordinate.}
Fix neighboring synthetic datasets $D,D'$.
Let $\mathcal C$ be the set of bins occupied in both.
Restricted to $\mathcal C$, their count vectors satisfy
\[
\sum_{\ell\in\mathcal C}
|c_{j,\ell}(D)-c_{j,\ell}(D')|\le2.
\]
Thus adding $\operatorname{Lap}(2/\eps_h)$ noise to these
counts is $\eps_h$-DP.

There is at most one bin occupied in $D$ but not in $D'$,
and its count in $D$ is one. Its probability of being retained is
\[
\Pr(1+W_{j,\ell}\ge\theta)
=\frac12\exp\left(-\frac{\eps_h(\theta-1)}2\right)
=\frac{\delta_h}{16}.
\]
The same statement holds with $D,D'$ interchanged.

Write $\mathcal M$ for the coordinate estimator and
$\mathcal M_{\mathcal C}$ for the same estimator restricted
to bins in $\mathcal C$. Couple their noises on the common bins.
On either input, their outputs agree unless a bin outside
$\mathcal C$ is retained. Hence, for every measurable set $S$,
\begin{align*}
\Pr(\mathcal M(D)\in S)
&\le \Pr(\mathcal M_{\mathcal C}(D)\in S)+\delta_h/16\\
&\le e^{\eps_h}
     \Pr(\mathcal M_{\mathcal C}(D')\in S)+\delta_h/16\\
&\le e^{\eps_h}\Pr(\mathcal M(D')\in S)
     +(1+e^{\eps_h})\delta_h/16\\
&\le e^{\eps_h}\Pr(\mathcal M(D')\in S)+\delta_h.
\end{align*}
The last inequality uses $\eps_h\le1$.
Thus each coordinate estimator is $(\eps_h,\delta_h)$-DP.

\paragraph{Composition across coordinates.}
Advanced composition with slack $\delta/2$ gives total
failure parameter $d\delta_h+\delta/2=\delta$ and privacy
parameter at most
\begin{align*}
\sqrt{2dH}\,\eps_h+d\eps_h(e^{\eps_h}-1)
&\le\frac{\eps}{\sqrt8}+2d\eps_h^2\\
&=\frac{\eps}{\sqrt8}+\frac{\eps^2}{8H}
\le\eps.
\end{align*}
Here $e^x-1\le2x$ for $0\le x\le1$, $\eps\le1$,
and $H\ge\log4$.
Replacing one original user changes only one synthetic user.
The estimator is therefore also $(\eps,\delta)$-user-level
DP for the original data.

\paragraph{Accuracy.}
For each coordinate, let
\[
I_j:=
\left[\mu_j-\frac4{\sqrt M},
      \mu_j+\frac4{\sqrt M}\right].
\]
Chebyshev's inequality gives
\[
\Pr(V_{i,j}\notin I_j)
\le\frac{\operatorname{Var}(V_{i,j})}{16/M}
\le\frac1{16}.
\]
The interval $I_j$ has length $h$ and intersects at most two
bins. Therefore some bin $B_{\ell_j^\star}$ intersecting $I_j$
has probability at least $15/32$.

Let $D_j$ count the observations in bins disjoint from $I_j$.
Since $\E D_j\le N/16$, Hoeffding's inequality gives
\begin{align*}
\Pr(c_{j,\ell_j^\star}<N/3)
&\le\exp\left[-2N(15/32-1/3)^2\right]
\le e^{-N/128},\\
\Pr(D_j>N/8)
&\le\exp\left[-2N(1/8-1/16)^2\right]
=e^{-N/128}.
\end{align*}
Consequently, if $N\ge128\log(4d/\eta)$, then with
probability at least $1-\eta/2$, simultaneously for every $j$,
\[
c_{j,\ell_j^\star}\ge N/3,
\qquad D_j\le N/8.
\]

There are at most $Nd$ occupied bins. Conditional on the data,
the Laplace tail bound and a union bound give
\[
\Pr\left(
\max_{j,\ell:\,c_{j,\ell}>0}|W_{j,\ell}|>a
\,\middle|\,V_1,\ldots,V_N
\right)\le\eta/2,
\qquad
a:=\frac2{\eps_h}\log\frac{2Nd}{\eta}.
\]
Choose $N$ sufficiently large that
\[
N\ge128\log(4d/\eta),\qquad
a\le N/16,\qquad
\theta\le N/4.
\]
On the intersection of the two preceding events,
\[
\widetilde c_{j,\ell_j^\star}
\ge N/3-N/16=13N/48>N/4,
\]
so this bin is retained. Every bin disjoint from $I_j$ has
\[
\widetilde c_{j,\ell}
\le D_j+a\le N/8+N/16=3N/16.
\]
Thus the selected bin must intersect $I_j$. Its midpoint
therefore satisfies
\[
|u_{0,j}-\mu_j|
\le\frac4{\sqrt M}+\frac h2
=\frac8{\sqrt M}.
\]
With probability at least $1-\eta$, this holds for all coordinates,
and hence
\[
\|u_0-\mu\|
\le8\sqrt{\frac dM}
=\frac14\sqrt{\frac dm}.
\]

\paragraph{User count.}
The three requirements on $N$ hold, for example, for
\[
Q:=\log\frac{8d}{\eta\delta_h\eps_h},
\qquad
N:=\left\lceil\frac{512Q}{\eps_h}\right\rceil.
\]
Indeed, $Q\ge4$, $Q\ge\log(4d/\eta)$, and
\[
\log(2Nd/\eta)\le Q+\log(129Q)\le3Q.
\]
Therefore
\[
a\le\frac{6Q}{\eps_h}\le\frac N{16},
\qquad
\theta\le\frac{3Q}{\eps_h}\le\frac N4,
\]
and the Hoeffding requirement also holds.
Since $L=1024$ is constant, the number of original users is
\[
LN
=O\left(
\frac{\sqrt{d\log(2/\delta)}}{\eps}
\log\frac{8d}{\eta\delta_h\eps_h}
\right)
=\widetilde O(\sqrt d/\eps).
\]
\end{proof}
\readingend

\subsubsection{Iterative refinement}

Starting from the private coarse estimate, we apply
clip-and-noise steps on fresh users. We first bound the error
of one step in terms of its clipping radius and current center
error. For $k>2$, we use the clipping-bias bound
of~\citet{agarwal2024person}. For $k=2$, we prove the
corresponding bound directly from the covariance assumption.

\begin{lemma}[One Clip-and-Noise step]
\label[lemma]{lem:directional-one-step}
Let $u$ be independent of $s$ fresh users and satisfy
$\|u-\mu\|\le\Gamma\le\sqrt{d/m}$.
Define
\[
\widehat\mu^+
:=\frac1s\sum_{i=1}^s
\Pi_{B(u,R)}\!\left(\frac1m\sum_{j=1}^m z_{i,j}\right)+\xi,
\qquad
\xi\sim\mathcal N\!\left(
0,\frac{C^2R^2\log(1/\delta)}{s^2\eps^2}I_d
\right),
\]
where $\xi$ is independent of $u$ and the data.
For each fixed $u$, this estimator is
$(\eps,\delta)$-user-level DP with respect to the fresh users.
With probability at least $9/10$, its error
$\Gamma^+=\|\widehat\mu^+-\mu\|$ satisfies
\begin{equation}
\label{eq:one-step-CN}
\Gamma^+
\le\widetilde O_k\left[
\sqrt{\frac{d}{ms}}
+\frac{R\sqrt{d\log(1/\delta)}}{s\eps}
+\frac{d^{(k-1)/2}}{m^{k-1}R^{k-1}}
+\frac{d^{k/2}\Gamma}{m^{k-1}R^k}
\right],
\end{equation}
provided
\[
R\gtrsim\sqrt{\frac{d\log(em)}m}.
\]
\end{lemma}

\begin{proof}
Condition on the center $u$.
For $k>2$, the clipping-bias bound follows
from~\citet[Theorem~4.2]{agarwal2024person}.
We first establish the corresponding bound for $k=2$,
then bound sampling error and privacy noise for every $k\ge2$.

For $k=2$, let $Y$ be one user average and $X:=Y-\mu$.
The directional second-moment assumption gives
\[
\operatorname{Cov}(Y)\preceq I_d/m,
\qquad \E\|X\|^2\le d/m.
\]
Define
\[
h(Y):=(1-R/\|Y-u\|)_+,
\]
with $h(Y)=0$ when $Y=u$.
For every unit vector $v$, the identity
$Y-\Pi_{B(u,R)}Y=h(Y)(Y-u)=h(Y)(X+\mu-u)$ gives
\begin{align*}
\bigl|\langle v,\E[Y-\Pi_{B(u,R)}Y]\rangle\bigr|
&=\left|
\E[h(Y)\langle v,X\rangle]
+\langle v,\mu-u\rangle\,\E h(Y)
\right|\\
&\le \E[h(Y)|\langle v,X\rangle|]+\Gamma\E h(Y)\\
&\le \sqrt{\E[h(Y)^2]}
      \sqrt{\E[\langle v,X\rangle^2]}
      +\Gamma\E h(Y)\\
&\le \sqrt{\Pr(\|Y-u\|>R)}
      \sqrt{\E[\langle v,X\rangle^2]}
      +\Gamma\Pr(\|Y-u\|>R).
\end{align*}
The first inequality uses
$|\langle v,\mu-u\rangle|\le\|u-\mu\|\le\Gamma$.
The second is Cauchy--Schwarz, and the last follows from
\[
0\le h(Y)^2\le h(Y)\le\mathbf1\{\|Y-u\|>R\}.
\]
Also,
\[
\Pr(\|Y-u\|>R)
\le\frac{\E\|Y-u\|^2}{R^2}
=\frac{\E\|X\|^2+\|u-\mu\|^2}{R^2}
\le\frac{2d}{mR^2}.
\]
Since $\E\langle v,X\rangle^2\le1/m$, taking the supremum
over unit vectors $v$ gives
\begin{equation}
\|\E\Pi_{B(u,R)}Y-\mu\|
\le\frac{\sqrt{2d}}{mR}+\frac{2d\Gamma}{mR^2}.
\label{eq:directional-k2-bias}
\end{equation}
This is $O(\sqrt d/(mR)+d\Gamma/(mR^2))$, giving the last
two terms of~\eqref{eq:one-step-CN} when $k=2$.

The sampling and privacy arguments below apply to every $k\ge2$.
Write
\[
Y_i:=\frac1m\sum_{j=1}^m z_{i,j},
\qquad
V_i:=\Pi_{B(u,R)}Y_i,
\qquad
\overline V:=\frac1s\sum_{i=1}^s V_i.
\]
The directional-moment assumption implies
$\E\|Y_i-\mu\|^2\le d/m$.
Conditional on $u$, the $V_i$ are independent. Therefore,
using nonexpansiveness of projection,
\begin{align*}
\E\!\left[
\|\overline V-\E[V_1\mid u]\|^2
\,\middle|\,u\right]
&=\frac1s\E\!\left[
\|V_1-\E[V_1\mid u]\|^2
\,\middle|\,u\right]\\
&\le\frac1s\E\!\left[
\|V_1-\Pi_{B(u,R)}\mu\|^2
\,\middle|\,u\right]\\
&\le\frac1s\E\!\left[
\|Y_1-\mu\|^2\,\middle|\,u\right]
\le\frac{d}{ms}.
\end{align*}

For each fixed $u$, replacing one user changes $\overline V$
by at most $2R/s$. Thus the Gaussian mechanism with
\[
\sigma:=\frac{CR\sqrt{\log(1/\delta)}}{s\eps}
\]
ensures $(\eps,\delta)$-user-level DP for sufficiently large $C$.

For any $\eta\in(0,1)$, Chebyshev's inequality and Gaussian
concentration give
\begin{align*}
\Pr\left(
\|\overline V-\E[V_1\mid u]\|
>\sqrt{\frac{2d}{ms\eta}}
\,\middle|\,u\right)
&\le\frac{\eta}{2},\\
\Pr\left(
\|\xi\|>\sigma\left[\sqrt d+\sqrt{2\log(2/\eta)}\right]
\,\middle|\,u\right)
&\le\frac{\eta}{2}.
\end{align*}
Since $\widehat\mu^+=\overline V+\xi$, the triangle inequality
and a union bound yield, with probability at least $1-\eta$,
\[
\|\widehat\mu^+-\mu\|
\le
\underbrace{\|\E[V_1\mid u]-\mu\|}_{\text{clipping bias}}
+\sqrt{\frac{2d}{ms\eta}}
+\sigma\left[\sqrt d+\sqrt{2\log(2/\eta)}\right].
\]
For $k=2$, the clipping bias is bounded
by~\eqref{eq:directional-k2-bias}; for $k>2$, it is bounded
by~\citet[Theorem~4.2]{agarwal2024person}.
Taking $\eta=1/10$, substituting $\sigma$, and using $d\ge1$
therefore gives
\[
\|\widehat\mu^+-\mu\|
\le\widetilde O_k\left[
\sqrt{\frac{d}{ms}}
+\frac{R\sqrt{d\log(1/\delta)}}{s\eps}
+\frac{d^{(k-1)/2}}{m^{k-1}R^{k-1}}
+\frac{d^{k/2}\Gamma}{m^{k-1}R^k}
\right]
\]
with probability at least $9/10$.
This proves~\eqref{eq:one-step-CN} for every $k\ge2$.
\end{proof}

Fix a current estimate $u$ with $\|u-\mu\|\le\Gamma$
and a block of $s$ fresh users. We must choose the clipping
radius $R$ for the next estimation step. Our goal is to
minimize the upper bound on the next error $\Gamma^+$
given by~\eqref{eq:one-step-CN}.

The sampling term $\sqrt{d/(ms)}$ does not depend on $R$.
Suppressing the hidden logarithmic factors, choosing the
radius therefore amounts to minimizing
\[
\frac{R\sqrt{d\log(1/\delta)}}{s\eps}
+\frac{d^{(k-1)/2}}{m^{k-1}R^{k-1}}
+\frac{d^{k/2}\Gamma}{m^{k-1}R^k}
\]
subject to $R\ge c\sqrt{d\log(em)/m}$.
The first term is the privacy noise; the other two bound
the clipping bias. In particular, the last term accounts
for the current center's error, so the radius must depend
on $\Gamma$ as well as on $s$.

Set
\[
\Lambda_s:=\log(e s m d/(\eps\delta))
\]
and define the target error
\begin{equation}
A_s=C'_k\Lambda_s^{C'_k}\left[
\sqrt{\frac d{ms}}
+\frac{d\sqrt{\log(1/\delta)}}{s\eps\sqrt m}
+\left(
\frac{d\sqrt{\log(1/\delta)}}{ms\eps}
\right)^{1-1/k}
\right],
\label{eq:def-As}
\end{equation}
where $C'_k$ is sufficiently large and depends only on $k$.
Choosing $s$ so that $A_s$ is a sufficiently small multiple
of $\alpha$ gives the three accuracy-dependent terms
in~\eqref{eq:directional-final}, up to logarithmic factors.

The following lemma specifies the radius and shows that,
with high probability, $\Gamma^+\le\Gamma/2$ whenever
$\Gamma$ exceeds a sufficiently large constant multiple
of $A_s$. This is the improvement needed to refine a
coarse estimate down to error $O_k(A_s)$.

\begin{lemma}[Radius choice and contraction]
\label[lemma]{lem:directional-contraction}
Under the assumptions of Lemma~\ref{lem:directional-one-step},
let $\Lambda_s$ and $A_s$ be as above, and set
$\eta_s:=1/[100\log(es)]$. Choose
\begin{equation}
\begin{aligned}
R=C_k\Lambda_s^{C_k}\max\Biggl\{
&\sqrt{\frac{d\log(em)}m},\quad
\left(
\frac{d^{(k-1)/2}s\eps}
     {m^{k-1}\sqrt{d\log(1/\delta)}}
\right)^{1/k},\\
&\left(
\frac{d^{k/2}\Gamma s\eps}
     {m^{k-1}\sqrt{d\log(1/\delta)}}
\right)^{1/(k+1)}
\Biggr\}.
\end{aligned}
\label{eq:target-aware-rho}
\end{equation}
For sufficiently large $C_k\ge1$ and then sufficiently large $C'_k$
in~\eqref{eq:def-As}, the resulting estimate satisfies
\begin{equation}
\Pr\left(
\|\widehat\mu^+-\mu\|\le C_kA_s
\left[1+\left(\frac{\Gamma}{A_s}\right)^{1/(k+1)}\right]
\right)\ge1-\eta_s.
\label{eq:target-aware-recursion}
\end{equation}
All constants depend only on $k$.
\end{lemma}

\begin{proof}
The second and third radii in the maximum are obtained by equating
each clipping-bias term with the privacy-noise term. Taking their
maximum ensures that neither bias term exceeds the noise term.
We first verify this and compute the resulting error; we then
account for the logarithmic factors in $A_s$.

Write
\[
q:=\frac{\sqrt{d\log(1/\delta)}}{s\eps},
\qquad
a:=\frac{d^{(k-1)/2}}{m^{k-1}},
\qquad
b:=\frac{d^{k/2}}{m^{k-1}},
\]
so the noise and bias terms are $qR$, $a/R^{k-1}$, and
$b\Gamma/R^k$. Set
\[
\rho_0:=\sqrt{\frac{d\log(em)}m},
\qquad H:=C_k\Lambda_s^{C_k}.
\]
Then~\eqref{eq:target-aware-rho} reads
\[
R=H\max\{\rho_0,(a/q)^{1/k},(b\Gamma/q)^{1/(k+1)}\}.
\]
Since $H\ge1$,
\[
R^k\ge a/q,
\qquad R^{k+1}\ge b\Gamma/q,
\]
and hence
\[
\frac{a}{R^{k-1}}\le qR,
\qquad \frac{b\Gamma}{R^k}\le qR.
\]
The resulting noise term satisfies
\[
qR\le H\left[
q\rho_0+(aq^{k-1})^{1/k}+(b\Gamma q^k)^{1/(k+1)}
\right].
\]

Let $E_s$ denote the bracketed expression in~\eqref{eq:def-As},
so that $A_s=C'_k\Lambda_s^{C'_k}E_s$. Direct calculation gives
\begin{align*}
q\rho_0
&=\frac{d\sqrt{\log(1/\delta)}}{s\eps\sqrt m}
  \sqrt{\log(em)}
\le E_s\sqrt{\Lambda_s},\\
(aq^{k-1})^{1/k}
&=\left(
\frac{d\sqrt{\log(1/\delta)}}{ms\eps}
\right)^{1-1/k}
\le E_s,\\
(b\Gamma q^k)^{1/(k+1)}
&=\Gamma^{1/(k+1)}
\left(
\frac{d\sqrt{\log(1/\delta)}}{s\eps\,m^{1-1/k}}
\right)^{k/(k+1)}\\
&\le\Gamma^{1/(k+1)}
\left(
\frac{d\sqrt{\log(1/\delta)}}{s\eps\sqrt m}
\right)^{k/(k+1)}
\le\Gamma^{1/(k+1)}E_s^{k/(k+1)}.
\end{align*}
Here $\log(em)\le\Lambda_s$, and $k\ge2$ implies
$m^{1-1/k}\ge\sqrt m$.

Apply the probability bounds in the proof of
Lemma~\ref{lem:directional-one-step} with failure probability $\eta_s$.
The sampling term increases by a factor of order $\eta_s^{-1/2}$;
the Gaussian norm bound contains
$\sqrt d+\sqrt{2\log(2/\eta_s)}$ in place of $\sqrt d$.
Since $\eta_s^{-1}=100\log(es)\le100\Lambda_s$, both increases
are bounded by fixed powers of $\Lambda_s$. Thus, for constants
$D_k,p_k>0$, with probability at least $1-\eta_s$,
\[
\Gamma^+
\le D_k\Lambda_s^{p_k}
\left[
\sqrt{\frac d{ms}}+qR+\frac{a}{R^{k-1}}+\frac{b\Gamma}{R^k}
\right]
\le D_k\Lambda_s^{p_k}
\left[\sqrt{\frac d{ms}}+3qR\right].
\]
Combining this with the three calculations above yields
\begin{equation}
\Gamma^+
\le M_s\left[E_s+E_s^{k/(k+1)}\Gamma^{1/(k+1)}\right],
\qquad M_s:=8D_kH\Lambda_s^{p_k+1}.
\label{eq:tcn-before-contraction}
\end{equation}
Choose $C'_k$ large enough that
\[
C'_k\Lambda_s^{C'_k}\ge\max\{1,M_s^{(k+1)/k}\}.
\]
Such a choice depends only on $k$, since $M_s$ is a fixed
power of $\Lambda_s$ times a constant depending only on $k$.
It ensures
\[
M_sE_s\le A_s,
\qquad M_sE_s^{k/(k+1)}\le A_s^{k/(k+1)}.
\]
Substituting in~\eqref{eq:tcn-before-contraction} gives
\[
\Gamma^+\le A_s+A_s^{k/(k+1)}\Gamma^{1/(k+1)}
=A_s\left[1+(\Gamma/A_s)^{1/(k+1)}\right],
\]
which implies~\eqref{eq:target-aware-recursion}.
In particular, if $\Gamma/A_s\ge(4C_k)^{(k+1)/k}$, then on
the same event
\[
\frac{\Gamma^+}{\Gamma}
\le C_k\frac{A_s}{\Gamma}
   +C_k\left(\frac{A_s}{\Gamma}\right)^{k/(k+1)}
\le\frac14+\frac14=\frac12.
\]
\end{proof}

The preceding lemma reduces a large error to a quantity proportional
to its $1/(k+1)$ power, after normalizing by $A_s$. Iterating this
bound will therefore reduce the initial error to $O_k(A_s)$ in
doubly logarithmically many rounds. To implement the radius choices
without knowing $\mu$, we maintain deterministic bounds $B_t$ on
$\|u_t-\mu\|$ and use $R(B_t)$ in round $t$. Each round uses fresh
users, as required by the one-step lemma.

\begin{lemma}[Iterative refinement]
\label[lemma]{lem:directional-refinement}
Assume $k\ge2$ and~\eqref{eq:intro-directional-model}. For an integer
$s\ge1$, let $A_s$ be as in~\eqref{eq:def-As}, and set
\[
B_0:=\frac14\sqrt{\frac dm},
\qquad
T:=1+\left\lceil
\log_{k+1}\log\!\left(e\max\{1,B_0/A_s\}\right)
\right\rceil.
\]
There is an $(\eps,\delta)$-user-level DP estimator using
$\widetilde O_k(\sqrt d/\eps)$ initialization users and
$T$ additional blocks of $s$ users such that
\[
\Pr\bigl(\|u_T-\mu\|\le C_{\rm ref,k}A_s\bigr)\ge9/10,
\qquad C_{\rm ref,k}:=e(2C_k)^{(k+1)/k}.
\]
The estimator requires no bound on $\|\mu\|$.
\end{lemma}

\begin{proof}
We first obtain a private estimate $u_0$ with error at most $B_0$.
We then apply the preceding lemma repeatedly, projecting each new
estimate onto $K:=B(u_0,B_0)$. When initialization succeeds,
$\mu\in K$, so this projection cannot increase the error and
ensures it stays below $2B_0$. This maintains the hypothesis
$\Gamma\le\sqrt{d/m}$ needed for each refinement step.

\paragraph{Algorithm.}
Apply Lemma~\ref{lem:location-free-histogram} with failure probability
$\eta_s=1/[100\log(es)]$, using a separate group of
$\widetilde O_k(\sqrt d/\eps)$ users. Let $u_0$ be its output and
set $K:=B(u_0,B_0)$.
Write $R(\Gamma)$ for the right-hand side
of~\eqref{eq:target-aware-rho}.
For $t=0,\ldots,T-1$, use $s$ fresh users and set
\begin{equation}
R_t:=R(B_t).
\label{eq:implementable-rho}
\end{equation}
For their averages $Y_{t,i}:=m^{-1}\sum_{j=1}^m z_{t,i,j}$, compute
\[
v_{t+1}:=\frac1s\sum_{i=1}^s\Pi_{B(u_t,R_t)}Y_{t,i}+\xi_t,
\qquad
\xi_t\sim\mathcal N\!\left(
0,\frac{C^2R_t^2\log(1/\delta)}{s^2\eps^2}I_d
\right),
\]
using independent Gaussian draws. Set $u_{t+1}:=\Pi_K(v_{t+1})$
and update
\begin{equation}
B_{t+1}:=\min\left\{
2B_0,\,
C_kA_s\left[1+\left(\frac{B_t}{A_s}\right)^{1/(k+1)}\right]
\right\}.
\label{eq:envelope-recursion}
\end{equation}
Return $u_T$.

The numbers $B_t$ and $R_t$ depend only on public
parameters; computing them does not require the true errors.

\paragraph{Validity of the error bounds.}
Initialization gives $\|u_0-\mu\|\le B_0$ with probability at least
$1-\eta_s$. Suppose initialization and the first $t$ refinement
rounds have succeeded, and $\|u_t-\mu\|\le B_t$.
Then $B_t\le2B_0\le\sqrt{d/m}$, and the next block is independent
of $u_t$. Lemma~\ref{lem:directional-contraction}, with $\Gamma=B_t$,
gives
\[
\|v_{t+1}-\mu\|
\le C_kA_s\left[1+\left(\frac{B_t}{A_s}\right)^{1/(k+1)}\right]
\]
with conditional probability at least $1-\eta_s$.
On this event, since $\mu\in K$,
\[
\|u_{t+1}-\mu\|
\le\min\{\|v_{t+1}-\mu\|,2B_0\}
\le B_{t+1}.
\]
Induction proves $\|u_t-\mu\|\le B_t$ whenever all preceding
steps succeed.

\paragraph{Number of rounds.}
Let $x_t:=\max\{1,B_t/A_s\}$ and $p:=1/(k+1)$.
Equation~\eqref{eq:envelope-recursion} gives
\begin{equation}
x_{t+1}\le2C_kx_t^p.
\label{eq:normalized-tcn-recursion}
\end{equation}
Iterating this inequality,
\[
x_T
\le(2C_k)^{1+p+\cdots+p^{T-1}}x_0^{p^T}
\le(2C_k)^{(k+1)/k}x_0^{p^T}.
\]
Our choice of $T$ ensures
\[
p^T\log x_0\le1,
\qquad x_0^{p^T}\le e.
\]
Consequently $B_T\le C_{\rm ref,k}A_s$, proving the accuracy
claim whenever every step succeeds.

\paragraph{Success probability and user count.}
Since $A_s\ge\sqrt{d/(ms)}$,
\[
B_0/A_s\le\sqrt s/4,
\qquad T+1\le3+\log_3\log(es)\le3\log(es).
\]
Initialization fails with probability at most $\eta_s$.
Conditional on success of all preceding steps, each refinement
round also fails with probability at most $\eta_s$.
Summing over the possible first failed step gives
\[
\Pr(\text{any failure})
\le(T+1)\eta_s\le3/100<1/10.
\]
The total number of users is the initialization cost plus $Ts$.

\paragraph{Privacy.}
Initialization is $(\eps,\delta)$-user-level DP. For fixed preceding
outputs, each refinement round is $(\eps,\delta)$-user-level DP
by the Gaussian mechanism, and projection onto $K$ is post-processing.
The user blocks are disjoint. If one user is replaced, the preceding
outputs are unchanged in distribution, only that user's block
requires a privacy guarantee, and subsequent computations use its
private output and the other, unchanged blocks. Adaptive parallel
composition therefore gives $(\eps,\delta)$-user-level DP for
the entire procedure.
\end{proof}

\subsection{Large target error: \texorpdfstring{$\alpha\ge\sqrt{d/m}$}{alpha >= sqrt(d/m)}}
For these accuracies, the norm-moment estimator already gives the
required user count. We verify that the directional-moment assumption
implies $\wtG_2=\wtG_k=\sqrt d$, then substitute these bounds into
Theorem~\ref{thm:mean}.

\begin{lemma}[Norm-moment estimation for large target error]
\label[lemma]{lem:few-user-regime-three}
Under~\eqref{eq:intro-directional-model}, let $k\ge2$ and
$\alpha\ge\sqrt{d/m}$. Algorithm~\ref{alg:main-me}, with
$\wtG_2=\wtG_k=\sqrt d$, is $(\eps,\delta)$-user-level DP and
achieves error at most $\alpha$ with probability at least $9/10$ using
\[
n=\widetilde O_k\left[
\frac d{\alpha\eps\sqrt m}+\frac1\eps\right]
\]
users.
\end{lemma}

\begin{proof}
The directional-moment assumption applies to each coordinate vector.
Thus Minkowski's inequality in $L_{k/2}$ gives
\[
\bigl(\E\|z-\mu\|^k\bigr)^{2/k}
=\left\|\sum_{j=1}^d(z_j-\mu_j)^2\right\|_{L_{k/2}}
\le\sum_{j=1}^d\bigl(\E|z_j-\mu_j|^k\bigr)^{2/k}
\le d.
\]
Also $\E\|z-\mu\|^2\le d$ by Lyapunov's inequality.
Using the second-moment bound in Theorem~\ref{thm:mean}, with
$\wtG_2=\wtG_k=\sqrt d$ and $\beta=1/10$, gives
\[
\|\widehat\mu-\mu\|
\le\widetilde O_k\left[
\sqrt{\frac d{nm}}+\frac d{n\eps\sqrt m}
+\frac{d^{3/4}}{\sqrt{nm\eps}}
\right]
\]
with probability at least $9/10$, provided the theorem's
$\widetilde\Omega_k(1/\eps)$ user requirement holds.
Making each error term at most $\alpha/3$ therefore requires
\[
n=\widetilde O_k\left[
\frac d{m\alpha^2}+\frac d{\alpha\eps\sqrt m}
+\frac{d^{3/2}}{m\eps\alpha^2}+\frac1\eps
\right]
\]
users. Since $\alpha\ge\sqrt{d/m}$ and $\eps\le1$,
\[
\frac d{m\alpha^2}\le1\le\frac1\eps,
\qquad
\frac{d^{3/2}}{m\eps\alpha^2}
=\frac d{\alpha\eps\sqrt m}\,
  \frac{\sqrt d}{\alpha\sqrt m}
\le\frac d{\alpha\eps\sqrt m}.
\]
Substituting these inequalities proves the claimed user count.
\end{proof}

\subsection{Proof of Theorem~\ref*{thm:main-directional}}
\begin{proof}
We prove the upper bound and then the lower bound.
\paragraph{Upper bound.} We combine the estimators for the three accuracy ranges.
For $k>2$ and $\alpha\le1/\sqrt m$, the small-error argument
above applies: the additional $\widetilde O_k(d/\eps)$ users
in the estimator of~\citet{agarwal2024person} are covered by
the term $d/(\alpha\eps\sqrt m)$ in~\eqref{eq:directional-final}. We apply their theorem with $\alpha/2$ and $\eps/2$, which lie in $(0,1)$ and change the user count only by a constant factor depending on $k$.

For $k>2$ and $1/\sqrt m<\alpha<\sqrt{d/m}$, or $k=2$ and
$0<\alpha<\sqrt{d/m}$, apply Lemma~\ref{lem:directional-refinement}.
Choose $s$ so that $C_{\rm ref,k}A_s\le\alpha$. Solving this
inequality for each of the three terms in~\eqref{eq:def-As} gives
\begin{equation}
s=\widetilde O_k\left[
\frac d{m\alpha^2}
+\frac{d\sqrt{\log(1/\delta)}}{\alpha\sqrt m\,\eps}
+\frac{d\sqrt{\log(1/\delta)}}{m\alpha^{k/(k-1)}\eps}
\right].
\label{eq:target-aware-s}
\end{equation}
The constants and logarithmic factors are chosen large enough
that all three contributions to $C_{\rm ref,k}A_s$ are at most
$\alpha/3$. Initialization is covered by the same user count because
\[
\alpha<\sqrt{d/m}
\quad\Longrightarrow\quad
\frac d{\alpha\eps\sqrt m}>\frac{\sqrt d}{\eps}.
\]
The $T=O_k(1+\log\log(es))$ rounds multiply $s$ by only a
logarithmic factor. Lemma~\ref{lem:directional-refinement}
therefore gives error at most $\alpha$ with probability at least
$9/10$, user-level $(\eps,\delta)$-DP, and the user count
in~\eqref{eq:directional-final}.

For $\alpha\ge\sqrt{d/m}$, apply
Lemma~\ref{lem:few-user-regime-three}. Its user count is covered
by the second and fourth terms of~\eqref{eq:directional-final},
up to logarithmic factors.

\paragraph{Lower bound.}
We work under the small-$\delta$ conditions in the theorem.
There are three sources of lower bounds: the statistical and
heavy-tail terms from~\citet{agarwal2024person}, the Gaussian
privacy term, and the cost of estimating an unrestricted mean
location. We check that each applies to our moment class.

First, let $M_k$ be the constant moment bound in
\citet[Corollary~6.4]{agarwal2024person}, and set
$b_k:=\max\{1,M_k^{1/k}\}$.
Dividing an observation from their class by $b_k$ gives
directional $k$-th moment at most one. Conversely, an estimator
with error $\alpha$ for our class would give error $b_k\alpha$
for their class by scaling its input and output.
For $\alpha\le1/b_k$, that corollary therefore gives
\[
n\ge\widetilde\Omega_k\left[
\frac d{m\alpha^2}
+\frac d{m\eps\alpha^{k/(k-1)}}
\right].
\]
The polynomial small-$\delta$ condition in our theorem is
sufficient for this application after adjusting its constants.

For the Gaussian privacy term, set
$\sigma_k:=(\E|g|^k)^{-1/k}$ for $g\sim\mathcal N(0,1)$.
Every $P=\mathcal N(\mu,\sigma_k^2I_d)$ satisfies
\[
\sup_{\|v\|=1}\E_{z\sim P}|\langle z-\mu,v\rangle|^k
=\sigma_k^k\E|g|^k=1.
\]
Lemma~\ref{lem:unbounded-gaussian} consequently gives
\[
n\ge\Omega_k\left(\frac d{\alpha\eps\sqrt m}\right)
\]
for every $\alpha>0$. Its privacy condition~\eqref{eq:gaussian-delta}
follows from the theorem's assumption $\delta\le c\eps^2/d$:
\[
\delta\bigl(\sqrt d+\sqrt{\log(e/\delta)}\bigr)
\le\delta\sqrt d+\sqrt\delta
\le(c+\sqrt c)\eps.
\]
Here we used $\delta\log(e/\delta)\le1$ and chose $c$ sufficiently
small. For $\alpha>1/b_k$, the other two accuracy-dependent
terms are bounded by constant multiples of this Gaussian term:
\[
\frac{d/(m\alpha^2)}{d/(\alpha\eps\sqrt m)}
=\frac{\eps}{\alpha\sqrt m}\le b_k,
\qquad
\frac{d/(m\eps\alpha^{k/(k-1)})}{d/(\alpha\eps\sqrt m)}
=\frac1{\sqrt m\,\alpha^{1/(k-1)}}\le b_k^{1/(k-1)}.
\]
Thus the first three lower-bound terms hold for every $\alpha>0$.

Finally, the point-mass construction in
\citet[Lemma~5.1]{narayanan2022few} requires
$n=\Omega(\log(1/\delta)/\eps)$ users for any finite uniform
accuracy guarantee when $\delta\le\eps^2$.
Every centered moment of a point mass is zero, so these
distributions belong to our class. Combining the four necessary
user counts gives
\[
n\ge\widetilde\Omega_k\left[
\frac d{m\alpha^2}
+\frac d{\alpha\eps\sqrt m}
+\frac d{m\eps\alpha^{k/(k-1)}}
+\frac{\log(1/\delta)}\eps
\right],
\]
which matches~\eqref{eq:directional-final} up to logarithmic factors.
\end{proof}

%% file: appendix/06_runtime.tex
\readingstart{runtime} \revisionstart \section{Computational complexity}\revisionend{}
\label{app:runtime}
 We count real-arithmetic operations for fixed $k$, separating calls to a sample-gradient oracle and Euclidean projection onto $\W$.  The bounds include reading the samples and computing user averages.  As in the underlying algorithms, evaluating elementary functions, bin indices, and sampling the stated noise distributions are counted as arithmetic operations; these are not bit-complexity bounds for unbounded real inputs.  Logarithms, including those in failure probabilities and target accuracy for directional estimation, are suppressed by $\widetilde O_k$.

 \paragraph{\revisionstart Norm-moment mean estimation.\revisionend{}}
 Computing all user averages costs $O(nmd)$.  FriendlyCore's concentration and filtering counts require $O(n^2)$ pairwise distances in $\R^d$, hence $O(n^2d)$ operations.  Forming the center, clipping the second group, averaging, and adding noise cost $O(nd)$.   The total is $O(nmd+n^2d)$; if user averages are supplied, omit the $O(nmd)$ preprocessing cost.

 \paragraph{\revisionstart SCO: cost per ERM iteration.\revisionend{}}
 An iteration on $s$ users uses $sm$ sample-gradient evaluations, one projection onto $\W$, and $O(dsm+ds^2)$ additional arithmetic operations.  These operations form the user gradients, compute pairwise distances, filter, average, and clip.  Projections onto Euclidean clipping balls cost $O(d)$ each and are included in this arithmetic count.

 \paragraph{\revisionstart SCO: number of iterations and total cost.\revisionend{}}
 Use $I,J,s,\Delta,\lambda_i$ from Algorithm~\ref{alg:poploc} and put $E_s=\mathcal E_{s,m}(G_2,G_k)$.  At phase $i$, Proposition~\ref{prop:private-erm-solver} uses
 \[
T_i=\left\lceil C_k\frac{G_2^2+\lambda_i^2D^2}{E_s^2}\right\rceil.
\]
 Since $E_s\ge G_2/\sqrt{sm}$ and $\lambda_iD=32^{i-1}\Delta$ with $\Delta=6C_k\Lambda^4E_s$,
 \[
T_i\lesssim_k 1+sm+\Lambda^8 32^{2(i-1)}
=\widetilde O_k(nm),
\]
 where $32^{I-1}\le32\sqrt{nm}$.  There are $IJ=\widetilde O(1)$ ERM calls, so the total number of iterations is $\widetilde O_k(nm)$.  Using $s\le n$ gives
  \[
\begin{array}{ll}
\text{sample-gradient evaluations:}&\widetilde O_k(n^2m^2),\\
\text{projections onto $\W$:}&\widetilde O_k(nm),\\
\text{additional arithmetic:}&\widetilde O_k(dn^2m(n+m)).
\end{array}
\]
 Computing geometric aggregations adds $O(dIJ^2)$ operations, absorbed by this bound.  Early termination can only reduce the cost.

\paragraph{\revisionstart Directional-moment mean estimation.\revisionend{}}
For target error $\alpha\ge\sqrt{d/m}$, we use the norm-moment
estimator, which costs $O(nmd+n^2d)$.
For smaller target errors, computing user averages costs
$O(nmd)$, and coordinatewise histogram initialization costs
$O(nd\log n)$ by sorting occupied bin indices.
Each clipping-and-noise refinement on $s$ user averages
costs $O(sd)$, including the small-error refinement
of~\citet{agarwal2024person} used when $k>2$.
Accounting for all rounds gives a total of
 $\widetilde O_k(nmd+n^2d)$ arithmetic operations.
\readingend

%% file: references.bib
@inproceedings{asi2024user,
  title={User-level Differentially Private Stochastic Convex Optimization: Efficient Algorithms with Optimal Rates},
  author={Asi, Hilal and Liu, Daogao},
  booktitle={Proceedings of the 27th International Conference on Artificial Intelligence and Statistics},
  series={Proceedings of Machine Learning Research}, volume={238},
  pages={4240--4248}, year={2024},
  url={https://proceedings.mlr.press/v238/liu24g.html}
}

@inproceedings{asi2024heavy,
  title={Private Stochastic Convex Optimization with Heavy Tails: Near-Optimality from Simple Reductions},
  author={Asi, Hilal and Liu, Daogao and Tian, Kevin},
  booktitle={Advances in Neural Information Processing Systems},
  volume={37}, year={2024},
  url={https://proceedings.neurips.cc/paper_files/paper/2024/hash/6cc066e48b391a749c2161f271fda80b-Abstract-Conference.html}
}

@inproceedings{lowy2024faster,
  title={Faster Algorithms for User-Level Private Stochastic Convex Optimization},
  author={Lowy, Andrew and Liu, Daogao and Asi, Hilal},
  booktitle={Advances in Neural Information Processing Systems},
  volume={37}, year={2024},
  url={https://proceedings.neurips.cc/paper_files/paper/2024/hash/ae338b3fe6624798efa80b6581924028-Abstract-Conference.html}
}

@article{lowy2026pure,
  title   = {Optimal Rates for Pure $\varepsilon$-Differentially Private Stochastic Convex Optimization with Heavy Tails},
  author  = {Lowy, Andrew},
  journal = {arXiv preprint arXiv:2604.06492},
  year    = {2026}
}

@inproceedings{agarwal2024person,
  title={Private Mean Estimation with Person-Level Differential Privacy},
  author={Agarwal, Sushant and Kamath, Gautam and Majid, Mahbod and Mouzakis, Argyris and Silver, Rose and Ullman, Jonathan},
  booktitle={Proceedings of the 2025 Annual ACM-SIAM Symposium on Discrete Algorithms},
  pages={2819--2880}, year={2025},
  doi={10.1137/1.9781611978322.92},
  url={https://epubs.siam.org/doi/10.1137/1.9781611978322.92}
}

@inproceedings{narayanan2022few,
  title={Tight and Robust Private Mean Estimation with Few Users},
  author={Esfandiari, Hossein and Mirrokni, Vahab and Narayanan, Shyam},
  booktitle={Proceedings of the 39th International Conference on Machine Learning},
  series={Proceedings of Machine Learning Research}, volume={162},
  pages={16383--16412}, year={2022},
  url={https://proceedings.mlr.press/v162/narayanan22a.html}
}

@article{barber2014privacy,
  title={Privacy and Statistical Risk: Formalisms and Minimax Bounds},
  author={Barber, Rina Foygel and Duchi, John C.},
  journal={arXiv preprint arXiv:1412.4451},
  year={2014}
}

@article{narayanan2023better,
  title={Better and Simpler Lower Bounds for Differentially Private Statistical Estimation},
  author={Narayanan, Shyam},
  journal={IEEE Transactions on Information Theory},
  volume={71}, number={2}, pages={1376--1388}, year={2025},
  doi={10.1109/TIT.2024.3511624}
}

@inproceedings{kamath2022improved,
  title={Improved Rates for Differentially Private Stochastic Convex Optimization with Heavy-Tailed Data},
  author={Kamath, Gautam and Liu, Xingtu and Zhang, Huanyu},
  booktitle={Proceedings of the 39th International Conference on Machine Learning},
  series={Proceedings of Machine Learning Research}, volume={162},
  pages={10633--10660}, year={2022}
}

@inproceedings{lowy2023large,
  title={Private Stochastic Optimization with Large Worst-Case Lipschitz Parameter: Optimal Rates for (Non-Smooth) Convex Losses and Extension to Non-Convex Losses},
  author={Lowy, Andrew and Razaviyayn, Meisam},
  booktitle={Proceedings of the 34th International Conference on Algorithmic Learning Theory},
  series={Proceedings of Machine Learning Research}, volume={201},
  pages={986--1054}, year={2023}
}

@inproceedings{levy2021user,
  title={Learning with User-Level Privacy},
  author={Levy, Daniel and Sun, Ziteng and Amin, Kareem and Kale, Satyen and Kulesza, Alex and Mohri, Mehryar and Suresh, Ananda Theertha},
  booktitle={Advances in Neural Information Processing Systems},
  volume={34}, pages={12466--12479}, year={2021}
}

@article{mollenhauer2025fuk,
  title={Fuk--Nagaev Inequality in Smooth Banach Spaces: Optimum Bounds for Distributions of Heavy-Tailed Martingales},
  author={Mollenhauer, Mattes and Fiedler, Christian},
  journal={arXiv preprint arXiv:2512.10012},
  year={2025}
}

@inproceedings{klsu19,
  title={Privately Learning High-Dimensional Distributions},
  author={Kamath, Gautam and Li, Jerry and Singhal, Vikrant and Ullman, Jonathan},
  booktitle={Proceedings of the 32nd Conference on Learning Theory},
  series={Proceedings of Machine Learning Research}, volume={99},
  pages={1853--1902}, year={2019},
  url={https://proceedings.mlr.press/v99/kamath19a.html}
}

@inproceedings{zhao2024huber,
 title={A Huber Loss Minimization Approach to Mean Estimation under User-level Differential Privacy},
 author={Zhao, Puning and Lai, Lifeng and Shen, Li and Li, Qingming and Wu, Jiafei and Liu, Zhe},
 booktitle={Advances in Neural Information Processing Systems}, volume={37}, year={2024},
 url={https://proceedings.neurips.cc/paper_files/paper/2024/hash/ead8e195df586b9e7c10fcd114a6b9d1-Abstract-Conference.html}}

@inproceedings{wang2020heavy,
 title={On Differentially Private Stochastic Convex Optimization with Heavy-tailed Data},
 author={Wang, Di and Xiao, Hanshen and Devadas, Srinivas and Xu, Jinhui},
 booktitle={Proceedings of the 37th International Conference on Machine Learning},
 series={Proceedings of Machine Learning Research},volume={119},pages={10081--10091},year={2020}}

@inproceedings{kamath2020heavy,
 title={Private Mean Estimation of Heavy-Tailed Distributions},
 author={Kamath, Gautam and Singhal, Vikrant and Ullman, Jonathan},
 booktitle={Proceedings of the 33rd Conference on Learning Theory},
 series={Proceedings of Machine Learning Research},volume={125},pages={2204--2235},year={2020}}

@inproceedings{dwork2006calibrating,
 title={Calibrating Noise to Sensitivity in Private Data Analysis},
 author={Dwork, Cynthia and McSherry, Frank and Nissim, Kobbi and Smith, Adam},
 booktitle={Theory of Cryptography Conference},pages={265--284},year={2006}}

@article{nemirovski2009robust,
 title={Robust Stochastic Approximation Approach to Stochastic Programming},
 author={Nemirovski, Arkadi and Juditsky, Anatoli and Lan, Guanghui and Shapiro, Alexander},
 journal={SIAM Journal on Optimization},volume={19},number={4},pages={1574--1609},year={2009}}

@inproceedings{tsfadia2022friendlycore,
 title={FriendlyCore: Practical Differentially Private Aggregation},
 author={Tsfadia, Eliad and Cohen, Edith and Kaplan, Haim and Mansour, Yishay and Stemmer, Uri},
 booktitle={Proceedings of the 39th International Conference on Machine Learning},
 series={Proceedings of Machine Learning Research}, volume={162}, pages={21828--21863},
 year={2022},
 url={https://proceedings.mlr.press/v162/tsfadia22a.html}
}

@inproceedings{bassily2023user,
 title={User-level Private Stochastic Convex Optimization with Optimal Rates},
 author={Bassily, Raef and Sun, Ziteng},
 booktitle={Proceedings of the 40th International Conference on Machine Learning},
 series={Proceedings of Machine Learning Research}, volume={202}, pages={1838--1851},
 year={2023}, url={https://proceedings.mlr.press/v202/bassily23a.html}
}

@misc{duchi2023notes,
  author={Duchi, John},
  title={Lecture Notes on Statistics and Information Theory},
  year={2023},
  url={https://web.stanford.edu/class/stats311/lecture-notes.pdf}
}

@article{dwork2014algorithmic,
  author={Dwork, Cynthia and Roth, Aaron},
  title={The Algorithmic Foundations of Differential Privacy},
  journal={Foundations and Trends in Theoretical Computer Science},
  volume={9}, number={3--4}, pages={211--407}, year={2014},
  doi={10.1561/0400000042},
  url={https://www.cis.upenn.edu/~aaroth/Papers/privacybook.pdf}
}

@article{pinelis1994optimum,
  author={Pinelis, Iosif},
  title={Optimum Bounds for the Distributions of Martingales in Banach Spaces},
  journal={The Annals of Probability},
  volume={22}, number={4}, pages={1679--1706}, year={1994}
}

@inproceedings{kamath2022new,
  title={New Lower Bounds for Private Estimation and a Generalized Fingerprinting Lemma},
  author={Kamath, Gautam and Mouzakis, Argyris and Singhal, Vikrant},
  booktitle={Advances in Neural Information Processing Systems},
  volume={35}, pages={24405--24418}, year={2022},
  url={https://papers.neurips.cc/paper_files/paper/2022/hash/9a6b278218966499194491f55ccf8b75-Abstract-Conference.html}
}

@inproceedings{kothari2022private,
  title={Private Robust Estimation by Stabilizing Convex Relaxations},
  author={Kothari, Pravesh and Manurangsi, Pasin and Velingker, Ameya},
  booktitle={Proceedings of Thirty Fifth Conference on Learning Theory},
  series={Proceedings of Machine Learning Research}, volume={178},
  pages={723--777}, year={2022}, publisher={PMLR},
  url={https://proceedings.mlr.press/v178/kothari22a.html}
}
